\documentclass[10pt,twocolumn,letterpaper]{article}

\usepackage[cameraready]{iccv}
\usepackage{times}
\usepackage{epsfig}
\usepackage{graphicx}
\usepackage{amsmath}
\usepackage{amssymb}
\usepackage{amsthm}
\usepackage{wrapfig}
\usepackage{caption}
\usepackage{subcaption}
\usepackage{tabularx}
\usepackage{makecell}
\usepackage {multirow}
\usepackage{adjustbox}
\usepackage{tcolorbox}
\usepackage{import}

\usepackage{tikz}
\usetikzlibrary{calc, backgrounds}

\usepackage{algorithm,algorithmic}

\definecolor{commentcolor}{RGB}{110,154,155}   
\newcommand{\PyCode}[1]{\STATE \ttfamily\textcolor{black}{#1}} 
\newlength\torchindent
\newcommand\INDB[1]{%
  \begingroup
  \setlength{\itemindent}{#1\torchindent}
  \addtolength{\algorithmicindent}{#1\torchindent}
}
\newcommand\INDE{\endgroup}

\newtheorem{thm}{Theorem}
\newtheorem{coro}{Corollary}

\newtheorem{lemma}{Lemma}
\newtheorem{defn}{Definition}
\newtheorem{assum}{Assumption}

\newcommand{\argmax}{\arg\max}
\newcommand{\argmin}{\arg\min}

\newcommand{\gC}{\mathcal{C}}

\newcommand{\gE}{\mathcal{E}}
\newcommand{\gF}{\mathcal{F}}

\newcommand{\gM}{\mathcal{M}}
\newcommand{\gN}{\mathcal{N}}

\newcommand{\gH}{\mathcal{H}}

\newcommand{\gX}{\mathcal{X}}
\newcommand{\gZ}{\mathcal{Z}}

\newcommand{\gFED}{\gF_{\epsilon,\delta}}
\newcommand{\gDc}{\mathcal{D}_{c}}
\newcommand{\gDcp}{\mathcal{D}_{c'}}

\newcommand{\sumC}{\sum_{c\in\mathcal{C}}}
\newcommand{\sumCp}{\sum_{c'\in\mathcal{C}}}
\newcommand{\exptx}[1]{\mathbb{E}_{x\sim #1}}
\newcommand{\exptxp}[1]{\mathbb{E}_{x'\sim #1}}

\usepackage{mathtools}
\usepackage{empheq}
\usepackage{nicefrac}
\usepackage{bm}

\newcommand*{\KL}{\mathrm{KL}}
\newcommand{\cE}{\mathcal{E}}

\newcommand{\cT}{\mathcal{T}}

\newcommand{\cX}{\mathcal{X}}

\definecolor{iccvblue}{rgb}{0.21,0.49,0.74}

\usepackage[pagebackref,breaklinks,colorlinks,allcolors=iccvblue]{hyperref}

\def\paperID{14121}
\def\confName{ICCV}
\def\confYear{2025}

\author{Won-Seok Choi\thanks{This manuscript was initially submitted to ICCV 2025 and is now made available as a preprint.}, Dong-Sig Han, Suhyung Choi, Hyeonseo Yang, Byoung-Tak Zhang\\
Seoul National University, Seoul, Korea\\
{\tt\small \{wchoi,dshan,shchoi,hsyang,btzhang\}@bi.snu.ac.kr}
}

\begin{document}

\title{OBSER: Object-Based Sub-Environment Recognition\\for Zero-Shot Environmental Inference}

\maketitle

\begin{abstract}
We present the Object-Based Sub-Environment Recognition (OBSER) framework, a novel Bayesian framework that infers three fundamental relationships between sub-environments and their constituent objects. In the OBSER framework, metric and self-supervised learning models estimate the object distributions of sub-environments on the latent space to compute these measures. Both theoretically and empirically, we validate the proposed framework by introducing the ($\epsilon,\delta$) statistically separable (EDS) function which indicates the alignment of the representation. Our framework reliably performs inference in open-world and photorealistic environments and outperforms scene-based methods in chained retrieval tasks. The OBSER framework enables zero-shot recognition of environments to achieve autonomous environment understanding.
\end{abstract}

\section{Introduction}

Deep learning agents are rapidly proliferating in the broader world, steered by advancements in artificial intelligence. In recent studies, large-scale models have opened a new avenue for exploration of a wider world beyond laboratory settings, encompassing diverse regional characteristics~\citep{shah2023vint,sridhar2024nomad}. To understand such complex environments, previous methods of environment recognition \citep{DBLP:journals/corr/abs-2106-10458,DBLP:journals/corr/abs-1908-06387} focus primarily on scene-based representation to specify the agent's location from the visual features of given observations~\citep{dorbala2022clip,agrawal2023clipgraphs,chen20232}. These methods have enabled sophisticated navigation and environmental recognition, yet they are often constrained by the need for supplementary inputs, such as language. By efficiently leveraging the information acquired from the environment, these limitations can be overcome and automated inference can be achievable.

\begin{figure}[t]
    \centering
    \includegraphics[width=\columnwidth]{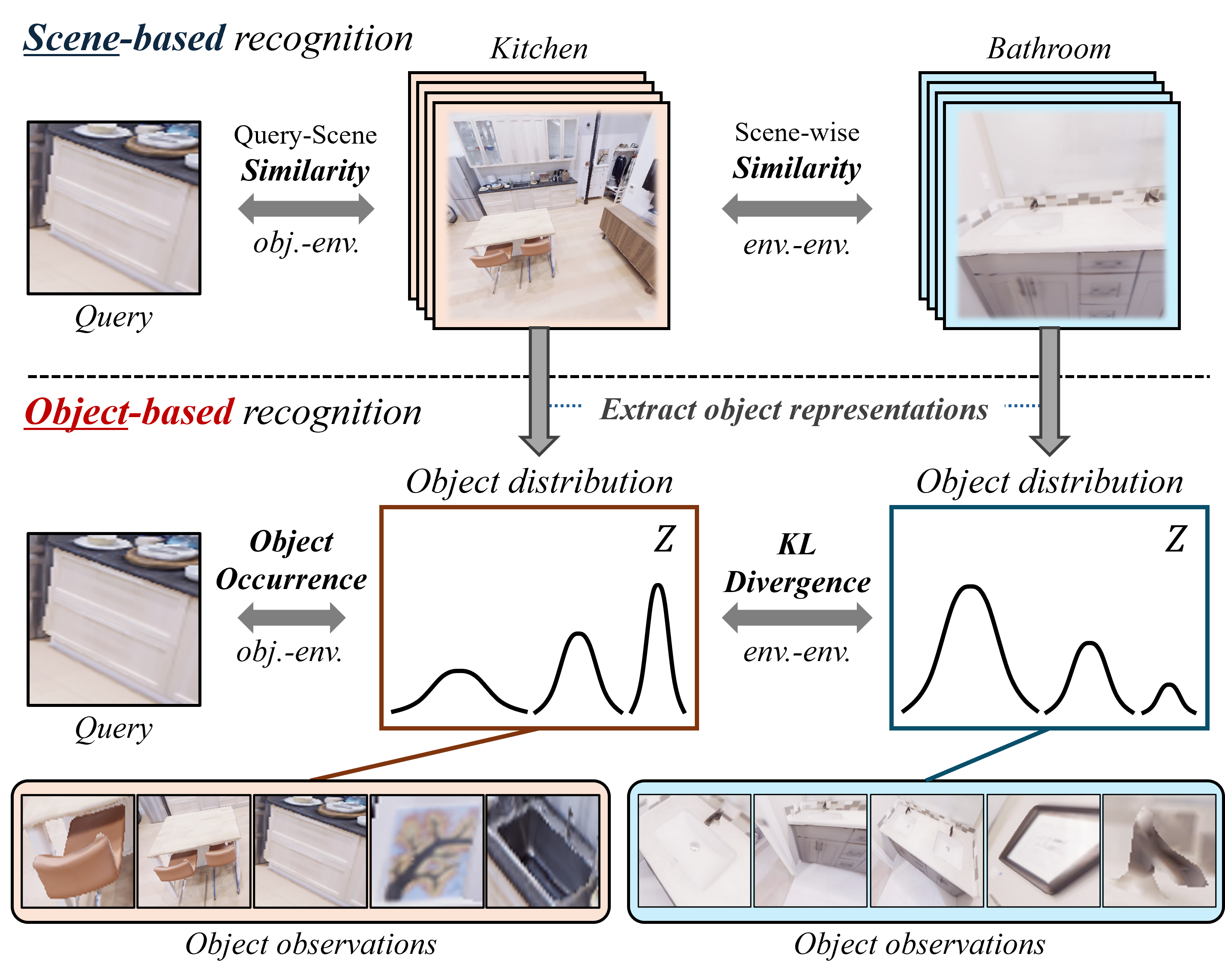}
    \caption{Comparison between scene-based and object-based (OBSER) environmental recognition. Unlike the conventional scene-based method—which treats object and environment recognition at the single level abstraction—the object-based approach establishes environmental recognition at multiple levels, enabling more precise inference.}
    \label{fig-1}
\end{figure}

\textit{A sub-environment}, with unique characteristics, is locally connected to form a complex environment. Humans recognize and differentiate these characteristics by identifying task-related objects in their surroundings. For example, an abundance of fish might indicate a river or ocean, while kitchens and bathrooms are distinguished by the presence of dishware. In this context, we claim that sub-environment recognition is defined as an inductive process, assuming an environment as a probability distribution of occurring objects. Based on these assumptions, we design the object-based recognition, as illustrated in Figure \ref{fig-1}.

\begin{figure*}[t]
    \centering    \includegraphics[width=.9\textwidth]{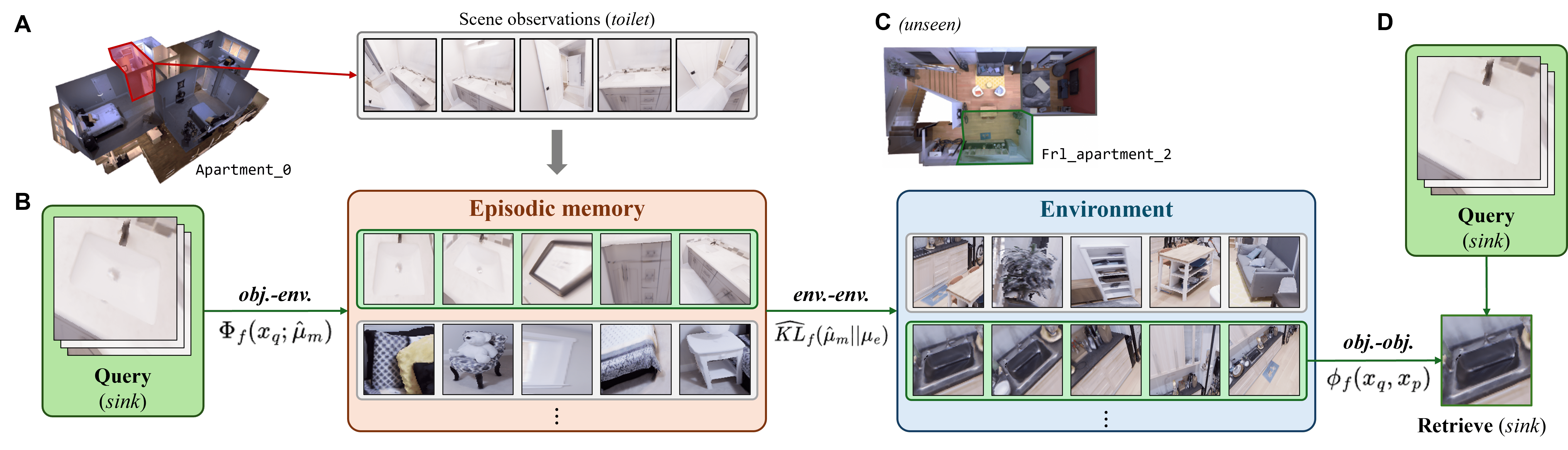}
    \caption{Visualization of chained inference using the OBSER framework in the Replica environment. (A) Scene observations are gathered from each room, and object observations are then extracted. (B) When a query is provided in the form of object observations, the framework first retrieves the corresponding room from its memory. (C) For inference in unseen environments, the rooms similar to the selected room are identified within that environment. (D) Finally, the framework retrieves the target object corresponding to the given query.}
    \label{fig-main}
\end{figure*}

\textit{Then how can environmental inference be achieved using a collection of object observations}? To address the issue, we utilize kernel density estimation to approximate the empirical object distributions on the latent space. For the feature extractor, we choose metric learning and self-supervised learning models~\citep{khosla2020supervised,chen2021empirical,chen2020simple,oquab2023dinov2}, since these methods have advantages over supervised methods in dealing with unstructured data in real-world problems~\citep{cao2019learning,liu2021self,choi2024duel}. The robustness and generalizability of these models enable our framework to achieve unsupervised zero-shot inference through comprehensive environmental understanding.

In this paper, we propose the Object-Based Sub-Environment Recognition (OBSER) framework, a Bayesian approach that leverages the empirical distribution of task-aware object representations. The proposed framework utilizes three fundamental relationships between sub-environments and their constituent objects—\textit{object-object, object-environment, and environment-environment}—and defines measures to establish these relationships. With an episodic memory, an agent can efficiently perform inference at the test time. Figure \ref{fig-main} shows the procedure of the chained inference with OBSER framework. With the given query, the agent retrieves related sub-environment from the memory and performs appropriate inference to solve the task.

To validate the proposed measures, we introduce the ($\epsilon, \delta$) statistically separable (EDS) function, which indicates \textit{separability} ($\epsilon$) and \textit{concentration} ($\delta$) of representations estimated from the feature extractor. We theoretically verify that when the model satisfies high concentration and separability, the proposed measures can sufficiently approximate their exact values. The experiment with the artificially generated environment using ImageNet dataset~\citep{deng2009imagenet} consistently supports our claims.

The OBSER framework is validated in both open-world and photorealistic environments. Experiments in the Minecraft environment, widely used in various tasks~\citep{baker2022video,qin2024mp5,li2024auto,chen2024diffusion}, confirm the robustness of environmental inference and show that self-supervised learning models better capture environmental factors than metric learning models. In the photorealistic Replica environment~\citep{straub2019replica}, the framework employs a pre-trained feature extractor for object images to achieve strong zero-shot inference performance and robustness in realistic observations. Additionally, we confirm that the OBSER framework outperforms methods that utilize vision-language models, such as CLIP~\citep{radford2021learning}.

\section{Related work}

\paragraph{Environment recognition}{Environment recognition is essential for embodied agents to successfully perform tasks such as navigation \citep{s20164532}. Previous research has primarily focused on understanding environments by analyzing the semantic distance between current and target observations \citep{yokoyama2023vlfmvisionlanguagefrontiermaps,shah2023vint,sridhar2024nomad} or by generating scene and environment graphs \citep{wang2023graphbasedenvironmentrepresentation} based on trained models. Also, recent works frequently employ vision-language models, such as CLIP~\citep{radford2021learning}, for environment recognition~\citep{dorbala2022clip,agrawal2023clipgraphs,chen20232}. In contrast, our interest lies in environmental relationships through the empirical distribution of occurring objects, providing a new method for understanding environments.
}

\paragraph{Metric learning}{Metric learning~\citep{sohn2016improved,liu2017sphereface,khosla2020supervised} is a method that optimizes the metric between objects on latent space to reflect the object-object relationships. Self-supervised learning, an advanced form of metric learning, achieves improved generalization performance by obtaining positive samples through data augmentation. We aim to extend the object-object relationship to express environmental relationships and employ these models as feature extractors for this purpose. Specifically, we use SupCon~\citep{khosla2020supervised}, MoCo variants~\citep{chen2020improved,chen2021empirical}, SimCLR~\citep{chen2020simple}, and DINO~\citep{caron2021emerging,oquab2023dinov2} to evaluate the proposed framework.}

\paragraph{Kernel method}{Kernel density estimation is a non-parametric method for estimating measures such as probability density function~\citep{zhang2018kernel,ghosh2006classification} and Kullback-Liebler divergence~\citep{ahuja2019estimating,ghimire2021reliable}. This work formulates the object similarity as a kernel to approximate the distribution of sub-environments. We also validate the precision of the measures with the EDS function, which computes the kernel density accumulated with the class-wise distribution.}

\section{Background}
In this section, we discuss key concepts to implement the proposed framework. OBSER framework uses kernel density estimation (Section 3.1) with metric learning models (Section 3.2) to reflect an empirical distribution of the environment (Section 3.3) with object-based feature extractor.

\subsection{Kernel density estimation}
Suppose that an object $x$ is defined on a data domain $X$. 
A \textit{latent} class $c\in \gC$ indicates abstracted concepts from an object required to solve a certain task $\cT$~\citep{arora2019theoretical,ash2021investigating,awasthi2022more}. With latent classes, $X$ is partitioned into $X_{c}$, which satisfies $X=\bigcup_{c\in\mathcal{C}}X_{c},X_{c}\cap X_{c'}=\emptyset$. For probabilistic inference, we define a kernel $\phi_f(x, x')$ to quantify \textit{belief} of sharing the identical \textit{latent} class:
\begin{align}
\phi_f(x, x') \coloneqq h\bigl(d_Z(f(x),f(x'))\bigr)\quad
x,x'\in X,
\label{eq-kernel}
\end{align}
with a metric function $d_Z:Z\times Z\rightarrow \mathbb{R}_{+}$ and a monotonic decreasing kernel function $h:\mathbb{R}_+\rightarrow [0,1],\enspace h(0)=1$. With the kernel function, we can compute the \textit{kernel density} of $x$ with respect to the given distribution $\mu$,
\begin{equation}
\Phi_f(x;\mu):=\exptxp{\mu}\bigl[\phi_f(x, x^\prime)\bigr].
\end{equation}
Kernel density implies the estimated probabilistic density of queried data $x$ for the distribution $\mu$. 

\subsection{Representation alignment in metric learning}

Analyses on representation alignment for classification tasks have gained increasing attention for \textit{robustness} and \textit{generalizability} in classification tasks. Recent works on Neural Collapse ~\citep{papyan2020prevalence,li2022principled,awasthi2022more} reveal that during optimization with classification or contrastive loss (InfoNCE), representations gradually form a simplex-ETF structure. However, these studies primarily focus on micro-level alignment using data point-wise measurements, making them less applicable to real-world noisy data.

On the other hand, some studies have proposed the uniformity measure, which quantifies the overall information in a data distribution~\citep{wang2020understanding,tian2021understanding,fang2024rethinking,choi2024duel}. While these works provide insights into macro-level alignment, they remain difficult to directly apply to kernel density estimation analysis. In this context, we propose a new statistical way to unify the understanding of representation alignment in terms of \textit{separability} and \textit{concentration}, as discussed in Section \ref{eds-function}.

\subsection{Sub-environment as an object distribution}
\label{sec-3.3}
Real-world data are often unstructured and imbalanced resulting in long-tailed distributions of their occurance~\citep{manning1999foundations,joseph2021towards}. Also, since object occurrence varies across different environments, it can serve as an important criterion in determining the environment~\citep{denis2017space}. Consequently, the following assumptions are introduced to define sub-environments in terms of a mixture of class-wise distributions.
\begin{assum}[Sub-environment]
A marginal data distribution $\mu(x)$ of a sub-environment is defined as:
\vspace*{-0.5em}
\begin{gather}
\mu(x):=\sumC \mu(c,x),\enspace\mu(c,x):= \omega(c)\cdot\rho_c(x),
\end{gather}
where $\omega(c)$ is the \underline{object occurrence} with latent class $c$ and $\rho_c(x) \coloneqq p(x|c)$ denotes the \underline{class-wise object distribution} defined on its corresponding partition $X_{c}\subseteq X$.
\end{assum}
\begin{assum}
The class-wise object distribution $\rho_c$ is \underline{consistent} in all sub-environments.
\end{assum}
In this paper, the environment $\cE \coloneqq \{(\mu_{i},R_{i})\}_{i=1}^N$ consists of multiple sub-environments $\mu_{i}$, each associated with a region $R_i$. In practice, the agent can empirically recognize the sub-environment $\hat\mu_i$ by obtaining object observations $\{x_{ij}\}_j$ at its locations $\{p_{ij}\}_j\in R_i$.

\section{(\texorpdfstring{$\epsilon,\delta$}{ε,δ}) statistically separable function}
\label{eds-function}

First, we introduce the $(\epsilon,\delta)$ statistically separable (EDS) function. For all objects, $x$ with latent class $c\in\gC$, delta ($\delta$) and epsilon ($\epsilon$) are defined with the kernel densities with the distribution $\gDc$ with the same class, and the distributions $\gDcp$ with different classes, respectively.

\begin{defn}[($\epsilon,\delta$) statistically separable function]
A function $f: X\to Z$ is ($\epsilon,\delta$) statistically separable if it satisfies 
\vspace*{-5pt}
\begin{subequations}
\begin{empheq}[left={\empheqlbrace\mspace{3mu}}]{align}
\:x\in X_{c},\forall c\in \gC,&\quad
\delta\le\Phi_f(x;\rho_c)\le 1,\\
\forall c'\neq c,&\quad
\delta\epsilon\le\Phi_f(x;\rho_{c^\prime})\le \epsilon.
\end{empheq}
\end{subequations}
in $\mu$-almost everywhere with $\exists\epsilon,\delta$, $0\le\epsilon\le\delta\le1$.
\label{def-eds-function}
\end{defn}
\noindent By the definition above, small $\epsilon$ and large $\delta$ indicate that the feature extractor is robust in the given task. In other words, with highly \textit{concentrated} ($\delta\simeq1$) and highly \textit{separated} ($\epsilon\simeq0$) representations, the estimated distribution $\hat\mu_f(c,x)$ with a Bayesian classifier $\hat\mu_f(c|x)$,
\begin{equation}
\hat\mu_f(c,x):=\hat\mu_f(c|x)\cdot\mu(x),
\end{equation}
becomes equivalent to the ground-truth distribution $\mu(c,x)$. With this evidence, we show that both $\epsilon$ and $\delta$ are optimized while minimizing the KL divergence between $\mu$ and $\hat\mu_f$, especially with a compact latent space. 
\begin{thm}[Bayesian metric learning] For an EDS function $f$, let $\exists k\ge 1, \delta=k\cdot\epsilon$. An upperbound $\Delta\gH$ of $\:\KL(\mu(c,x)\Vert\hat{\mu}_f(c,x))$ is derived as:
\begin{equation}
\KL(\mu\Vert\hat{\mu}_f)\le\log\bigl(1+({\lvert\gC\rvert-1})/{k}\bigr):=\Delta\gH,
\end{equation}
and if $\Delta\gH\rightarrow+0$, then $k\rightarrow\infty$.
\label{thm1}
\end{thm}

\begin{coro}[Compact space] Let $d_{Z}: Z\times Z\rightarrow [0, d_{\max}]$, monotonic deacreasing $h: \mathbb{R}_+\rightarrow[0,1]$ with $h(0)=1$.
Without any additional restrictions of $Z$,
$\Delta\gH \rightarrow \Delta\gH_{\min}$, and
$\delta\rightarrow 1,\:\epsilon\rightarrow \phi_{\min}$ for some $\Delta\gH_{\min}$ and $\phi_{\min}$.
\label{thm1-compact}
\end{coro}

Proofs of Theorem \ref{thm1} and Corollary \ref{thm1-compact} are provided in Appendix \ref{apdx-eds-opt}. Note that the KL divergence $\KL(\mu\Vert\hat{\mu}_f)$ decomposes into $f$-dependent term and an entropy term:
\begin{align}
&\KL\bigl(\mu\Vert \hat{\mu}_f\bigr)=\exptx{\mu}\biggl[-\log \frac{\Phi_f(x;\rho_{\scriptscriptstyle+})}{\Phi_f(x;\mu)}\biggr]+\mathcal{H}(\bm{\omega}),
\end{align}
with \textit{positive} distribution $\rho_{\scriptscriptstyle+}$ associated with given data $x$. We denote the former term as \textit{generalized} metric learning since it represents diverse loss functions in metric learning.

\section{Object-based sub-environment recognition}

In this section, we introduce object-based sub-environment recognition (OBSER) with three fundamental relationships: object-object, object-environment, and environment-environment relationships.

\begin{defn}[Object-based sub-environment recognition]
With a task-aware feature extractor $f:X\to Z$, the OBSER is defined with following measurements:
\vskip 0.5em

\noindent\quad\:\: i) Similarity measure (obj.$-$obj.):\enspace $\phi_f(x,x^\prime)$,

\noindent\quad\: ii) Object occurrence
(obj.$-$env.):\enspace  $\hat{\omega}_f(c), x\in X_c$,

\noindent\quad iii) KL divergence (env.$-$env.):\enspace  $\widehat{\KL}_f(\mu\Vert\nu)$.
\end{defn}

Each relationship is formulated through the lens of the kernel density estimation, and each measure is estimated using an empirical distribution of observations. Furthermore, we show both theoretically and practically that the optimized EDS function, with its high concentration and separability, guarantees the convergence of the estimated measures to their exact values.


\subsection{Object-object recognition} 
The object-object recognition is used to retrieve the most appropriate object with the given query. Object-object recognition is a fundamental recognition concept for achieving other sub-environment recognition. With the query $x_q$ and candidates $\{x_k\}_k^K$ for comparison, conventional evaluation metrics for the classification task originated from the following formula:
\begin{equation}
\textstyle x^*:=\argmax_{x_k \in\{x_k\}_k^K}\phi_f(x_q,x_k).
\end{equation}
To show that object-object recognition is directly related to the ($\epsilon,\delta$) values, we evaluate both task accuracies and EDS values for every model in Section \ref{sec-val-imagenet}.

\subsection{Object-environment recognition} 

Object-environment recognition is defined to retrieve a suitable sub-environment $\mu_i$ in an environment $\gE$ with a given query object. With object-environment recognition, an agent can infer a sub-environment that contains queried objects with the highest chance. We define the object occurrence of sub-environment given a query $x_q$ as $\hat\omega_f(c)$. With samples $\{x_1^{\mu},\cdots,x_N^{\mu}\}\sim\mu$ and query $x_q\in X_c$, the object occurrence $\hat\omega_f(c)$ can be computed as $\hat\omega_f(c):=\Phi_f(x;\mu)$.

Depending on the \textit{concentration} of the EDS function, the representation of the query can be distant from the other objects with the same latent class. Therefore, we utilize multiple queries $Q=\{x_1,\cdots,x_k\}$ instead of a single query. In this case, the mean representation $\bar{r}$ is used to compute the density instead of individual representations.

\subsection{Environment-environment recognition} 
The environment-environment recognition allows the agent to measure the changes in its circumstances and infer similar sub-environments. To define the difference between two sub-environments, we utilize the Kullback-Leibler (KL) divergence between distributions from each sub-environment $\mu$ and $\nu$. Under the assumptions in Section \ref{sec-3.3}, the KL divergence between $\mu$ and $\nu$ is derived with the object occurrence $\omega^\mu$ and $\omega^\nu$:
\begin{equation}
\KL(\mu(c,x)\Vert\nu(c,x))=\sum_{c\in\gC}\omega^\mu(c)\cdot\log\frac{\omega^\mu(c)}{\omega^\nu(c)}. 
\end{equation}
Since the agent cannot access the class information without supervision, we use kernel density estimation to approximate the KL divergence, denoted as $\widehat{\KL}_f(\mu\Vert\nu)$.
\begin{defn}[KL divergence estimation]
With given samples $\{x_1^\mu,\cdots,x_N^\mu\}\sim\mu$ and $\{x_1^\nu,\cdots,x_M^\nu\}\sim\nu$, approximated KL divergence $\widehat{\KL}_f(\mu \Vert \nu)$ can be computed as:
\begin{equation}
\widehat{\KL}_f(\mu\Vert\nu):=\exptx{\mu}\biggl[\log \frac{\Phi_f(x;\mu)}{\Phi_f(x,\nu)}\biggr].
\end{equation}
\label{KL-div-estimation}
\end{defn}

In terms of kernel density estimation, low \textit{separability} causes the \textit{oversmoothing} effect, which increases ambiguity between different objects. Additionally, low \textit{concentration} leads to a \textit{fragmentation} effect, increasing the misclassification of the same object. Thus, addressing both separability and oversmoothing is essential for precise estimation.

\begin{figure*}[t]
    \centering    \includegraphics[width=\textwidth]{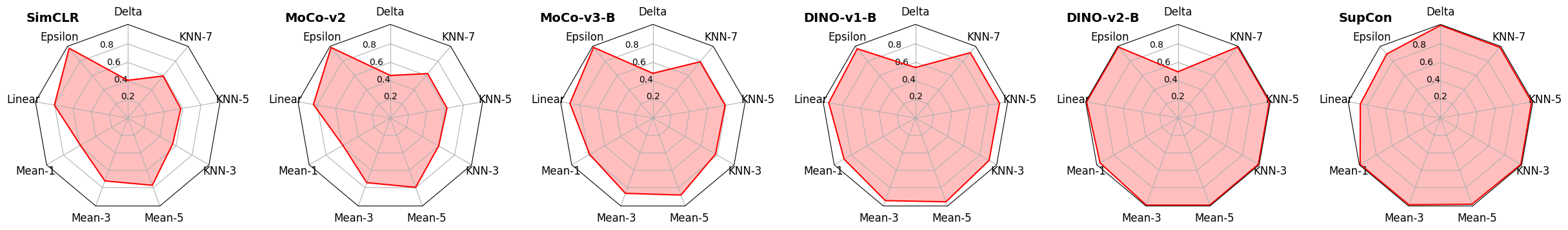}
    \caption{Radar chart of EDS values ($\tau=0.5$) and down-stream task accuracies for metric learning and self-supervised learning models. The separability value ($1-\epsilon$) is reported instead of $\epsilon$ for the ease of comparison. Each score is normalized within the interval $[0,1]$. The table reported in \cref{apdx-eds-table} shows the original, unnormalized results.}
    \label{eds-imagenet}
\end{figure*}

\subsection{OBSER with the optimized EDS function}

To validate the proposed measures, we show that each measure converges to the ground-truth value with the optimized EDS function. Specifically, we prove the following lemma and theorem under the conditions $\delta\rightarrow 1$ and $\epsilon\rightarrow 0$:
\vskip 0.5em

\noindent\textbf{Lemma \ref{object-occurrence-eds}.} $\hat{\omega}_f(c)$ \textit{converges to} $ \omega(c), \enspace x \in \cX_c$,
\vskip 0.5em
\noindent\textbf{Theorem \ref{KL-div-eds}.} $\widehat{\KL}_f(\mu||\nu)$ \textit{converges to} ${\KL}(\mu||\nu)$.
\vskip 0.5em

In other words, by optimizing EDS function $f$ via Theorem \ref{thm1} in a compact space, the error bound becomes tighter, and eventually each measure is squeezed to its ground-truth value. Detailed proofs are provided in Appendix \ref{apdx-ser-eds}.

\subsection{Validation with ImageNet dataset}
\label{sec-val-imagenet}

In this section, we validate the proposed measures with metric learning and SSL models using the ImageNet dataset. To employ a hypersphere space as an embedding space instead of a Euclidean space, we set the kernel as
\begin{equation}
\phi_f(x,x'):=\exp((f(x)^\top f(x')-1)/\tau),
\end{equation}
with temperature $\tau$. With the temperature $\tau$, we can adjust the influence of both $\delta$ and $\epsilon$. We choose SupCon~\citep{khosla2020supervised}, MoCo-variants~\citep{chen2020improved,chen2021empirical}, SimCLR~\citep{chen2020simple}, DINO-variants~\citep{caron2021emerging,oquab2023dinov2} for comparison. For reproducibility, we use the pre-trained models that are publicly reported. More details and results can be found in Appendix \ref{apdx-imagenet-res}.

\paragraph{Object-object}{For the experiments, we choose the mean classifier to show the influence of \textit{separability} ($\epsilon$) and KNN classifier for \textit{concentration} ($\delta$) each. Figure~\ref{eds-imagenet} demonstrates EDS values ($\epsilon,\delta$) and classification accuracies of various metric and self-supervised learning (SSL) models with the ImageNet dataset. SupCon, a metric learning model, shows better mean classifier accuracy and KNN accuracy than other SSL models. DINO-v2 shows the most competitive performance among SSL models and the best linear probing accuracy. Consistently, we can say that a larger gap between $\delta$ and $\epsilon$ is important for the downstream task performance, and models with larger embedding space can perform better when the models have similar EDS values.}

\begin{figure}[t]
\centering
    \begin{subfigure}{0.48\columnwidth}
        \includegraphics[width=\textwidth]{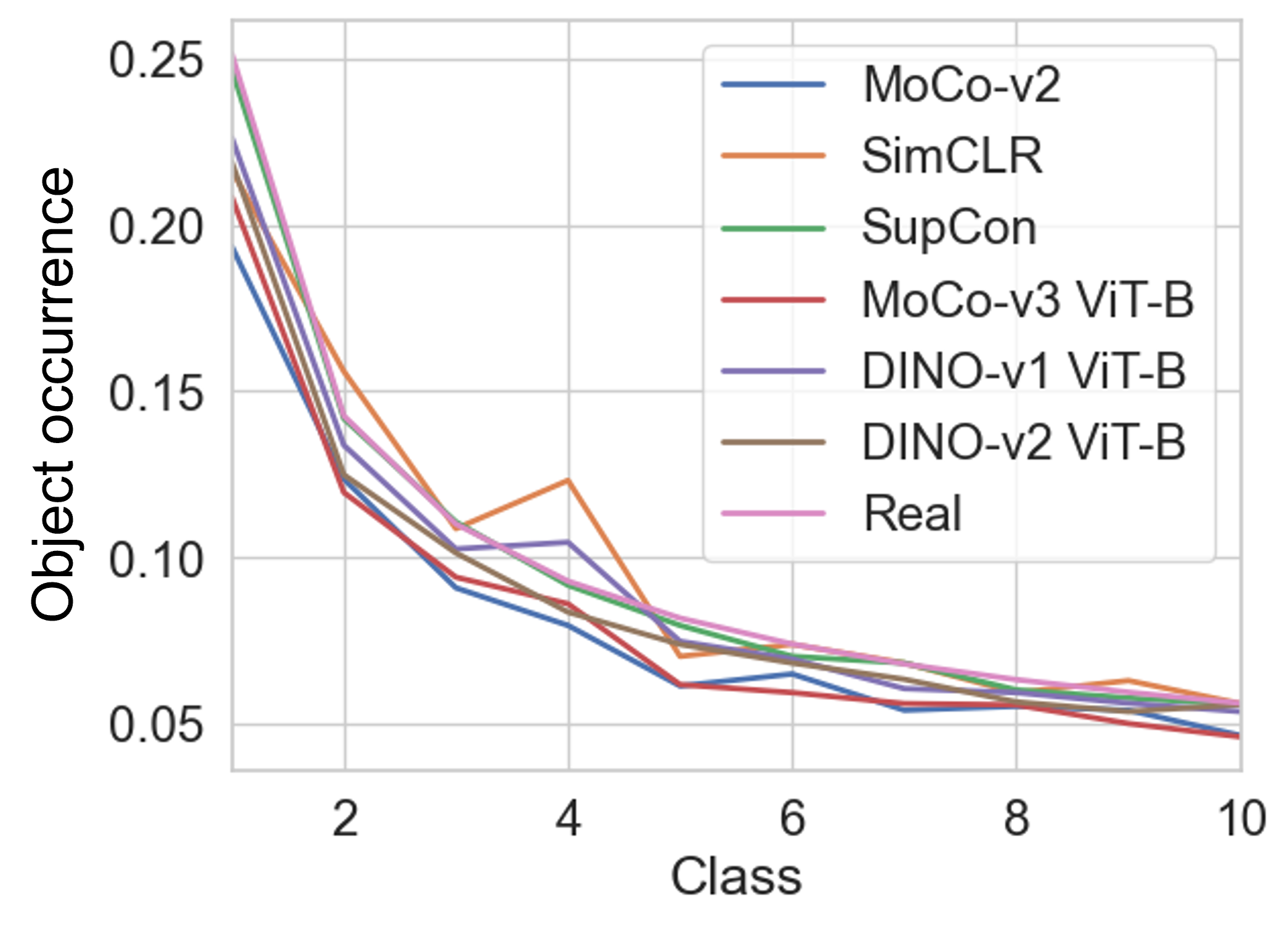}
        \caption{Proposed (Zipf)}
    \end{subfigure}%
    \hfill
    \begin{subfigure}{0.48\columnwidth}
        \includegraphics[width=0.9\textwidth]{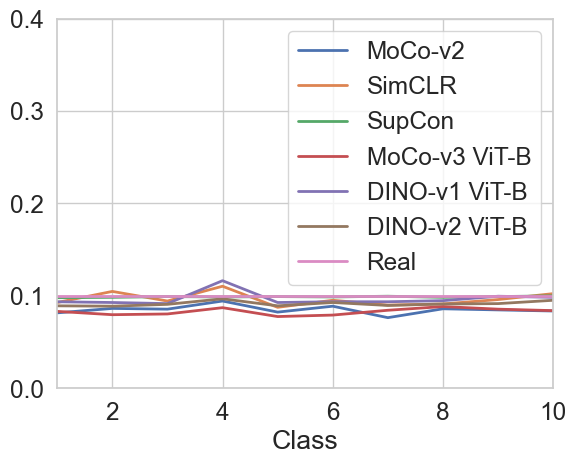}
        \caption{Proposed (Uniform)}
    \end{subfigure}
    \begin{subfigure}{0.48\columnwidth}
        \hspace* {0.5em}
        \includegraphics[width=0.9\textwidth]{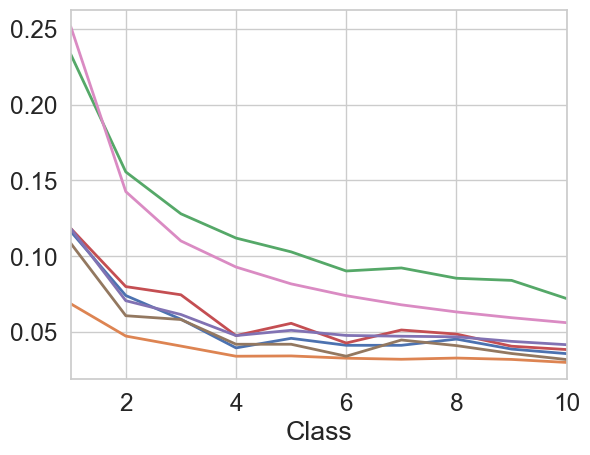}
        \caption{Direct ($\tau=0.25$)}
        \label{occurrence-under}
    \end{subfigure}%
    \hfill
    \begin{subfigure}{0.48\columnwidth}
        \includegraphics[width=0.9\textwidth]{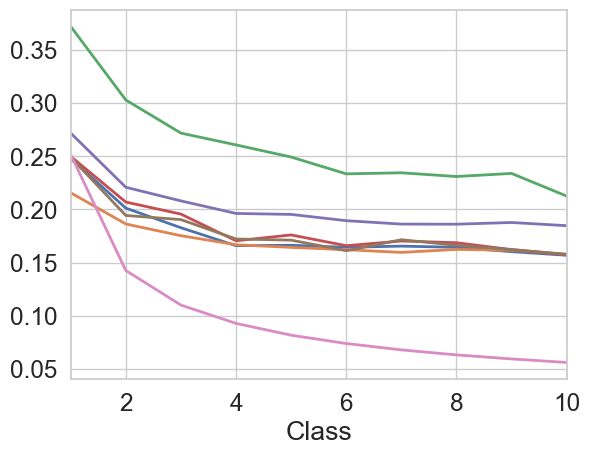}
        \caption{Direct ($\tau=0.5$)}
        \label{occurrence-over}
    \end{subfigure}
    \caption{Estimated object occurrence using distributions generated from ImageNet. Our method approximates both the (a) Zipf and (b) uniform distributions; however, without an adaptive classifier, it under- or overestimates based on $\tau$ (c, d).}
    \label{res-imagenet-occurrence}
\end{figure}

\paragraph{Object-environment}{Figure \ref{res-imagenet-occurrence} shows the estimated object occurrence using artificially generated subsets of ImageNet. We choose a Zipf distribution to mimic the real-world situations, with $\omega(c)\propto1/c^{-\alpha}, \alpha=0.5$ and a uniform distribution for the evaluation. At low temperatures, models with low \textit{concentration} (low $\delta$) tend to underestimate object occurrence (Figure \ref{occurrence-under}), while tending to overestimate at high temperatures (Figure \ref{occurrence-over}), due to low \textit{separability} (high $\epsilon$). To address this issue, we use an adaptive classifier with the mean vector of the query $\bar{r}$. With this quantization, the precision is significantly improved.
}

\begin{figure}[t]
    \centering
    \begin{subfigure}{0.48\columnwidth}
        \includegraphics[width=\textwidth]{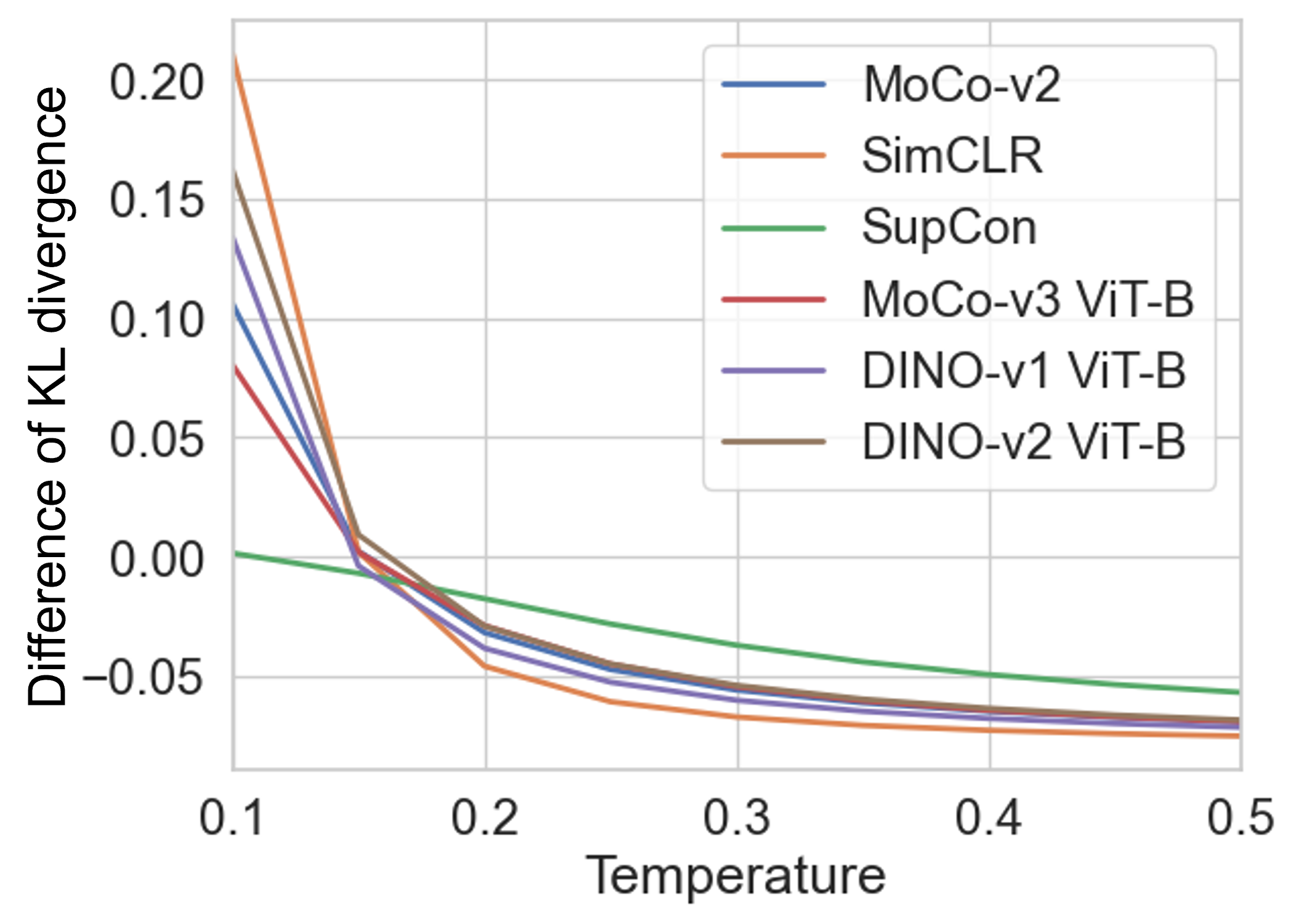}
        \caption{Scenario 1}
    \end{subfigure}%
    \hfill
    \begin{subfigure}{0.48\columnwidth}
        \includegraphics[width=0.95\textwidth]{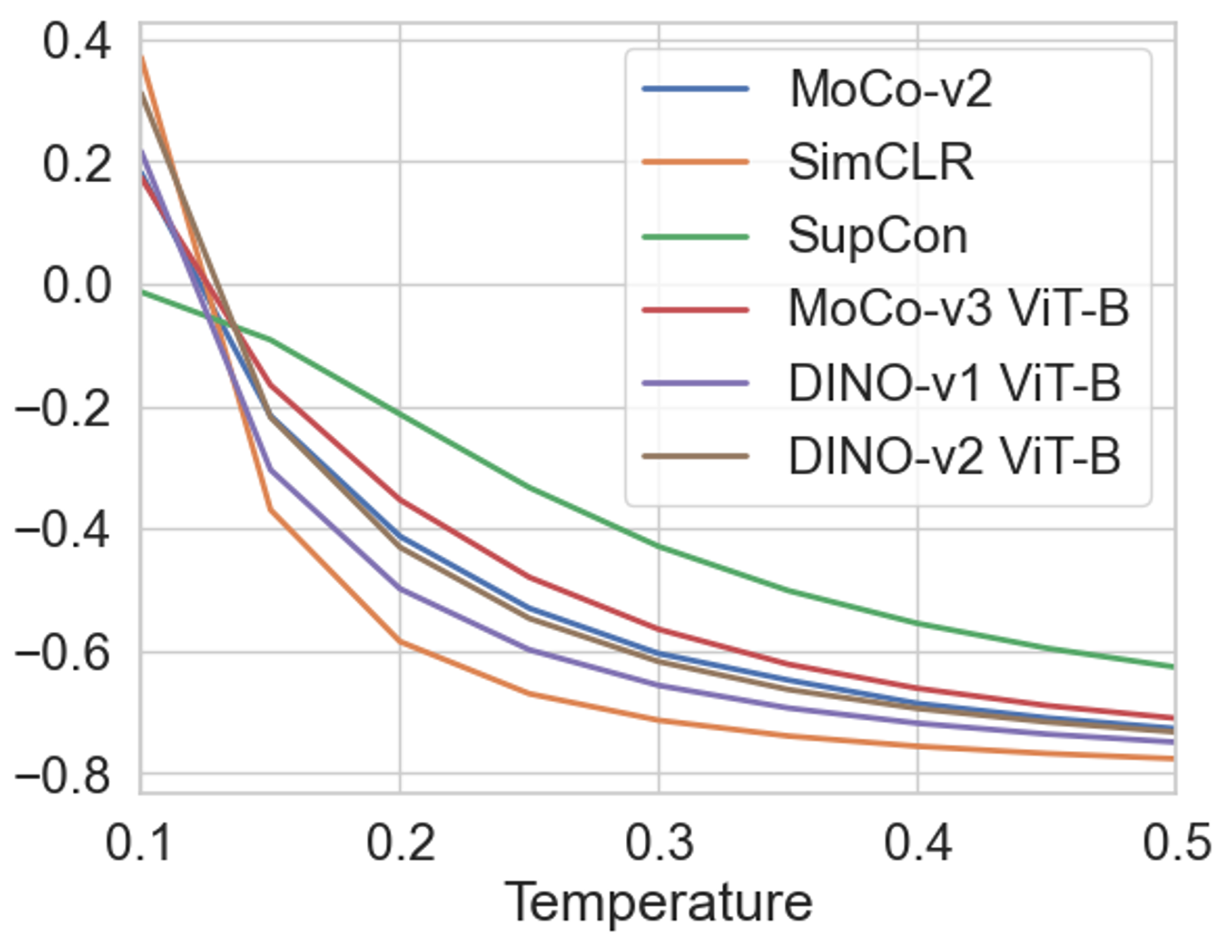}
        \caption{Scenario 2}
    \end{subfigure}
    \vskip -0.5em
    \caption{Difference between the exact KL divergence values and their estimates across scenarios. High temperatures lead to oversmoothing, while low temperatures cause fragmentation, highlighting the need for an appropriate $\tau$.}
    \label{res-imagenet-kld}
\end{figure}

\paragraph{Environment-environment}{Figure \ref{res-imagenet-kld} shows the difference between the exact and estimated KL divergence. We observe both \textit{oversmoothing} and \textit{fragmentation} effects in all models. We have also empirically found that the $\tau$ near $[0.12,0.18]$ performs best in these scenarios. Additional analysis for the experiment is presented in Appendix \ref{apdx-imagenet-kldiv-res}.}

\begin{figure*}[!t]
\centering
    \begin{subfigure}[t]{0.24\textwidth}
        \centering
        \includegraphics[height=10.5em]{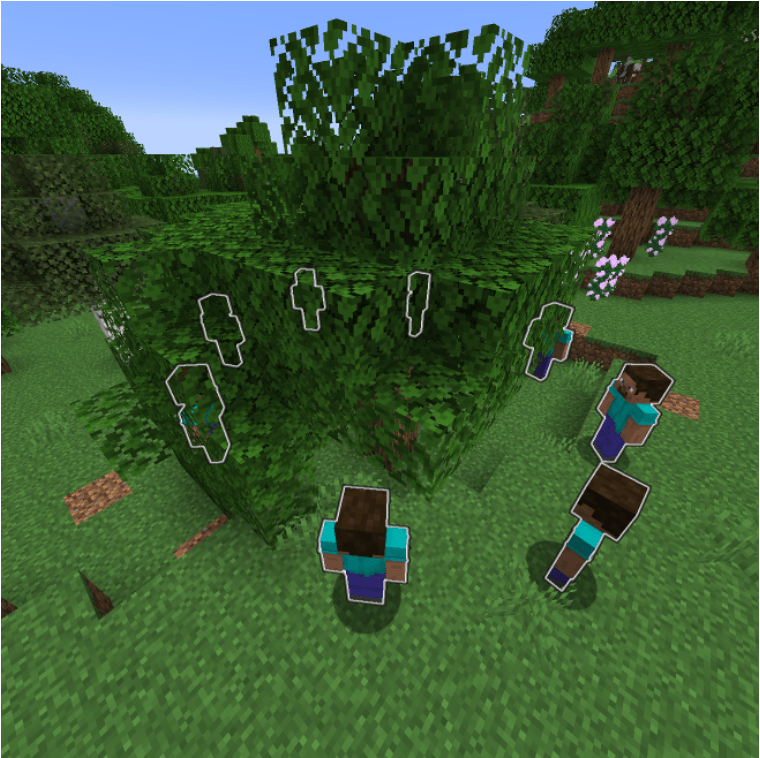}
        \caption{Object observations}
        \label{minecraft-obj-obs}
    \end{subfigure}
    \hfill
    \begin{subfigure}[t]{0.28\textwidth}
        \centering
        \includegraphics[height=10.5em]{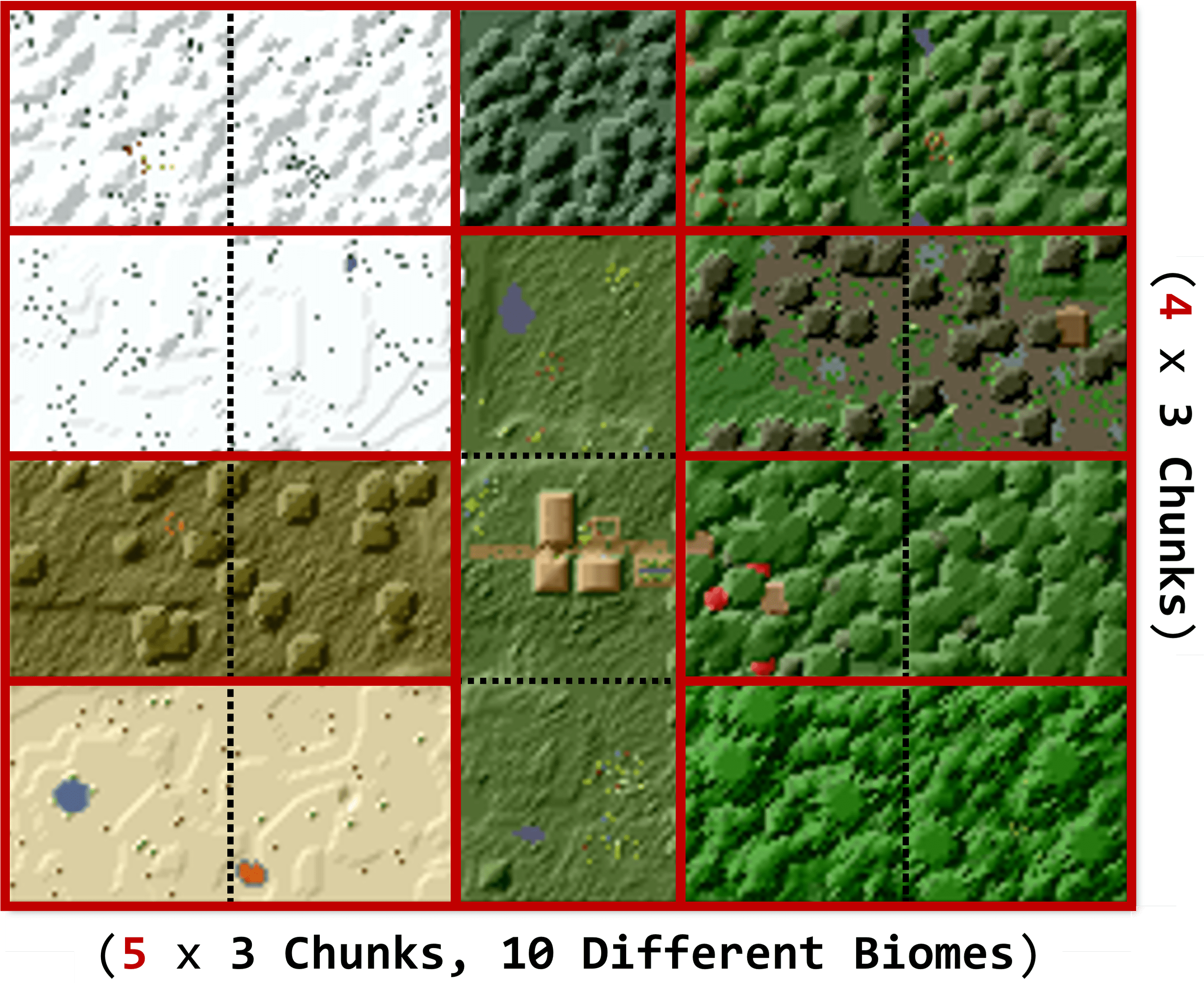}
        \caption{Miniature environment}
        \label{minecraft-miniature}
    \end{subfigure}
    \hfill
    \begin{subfigure}[t]{0.47\textwidth}
        \centering
        \includegraphics[height=10.5em]{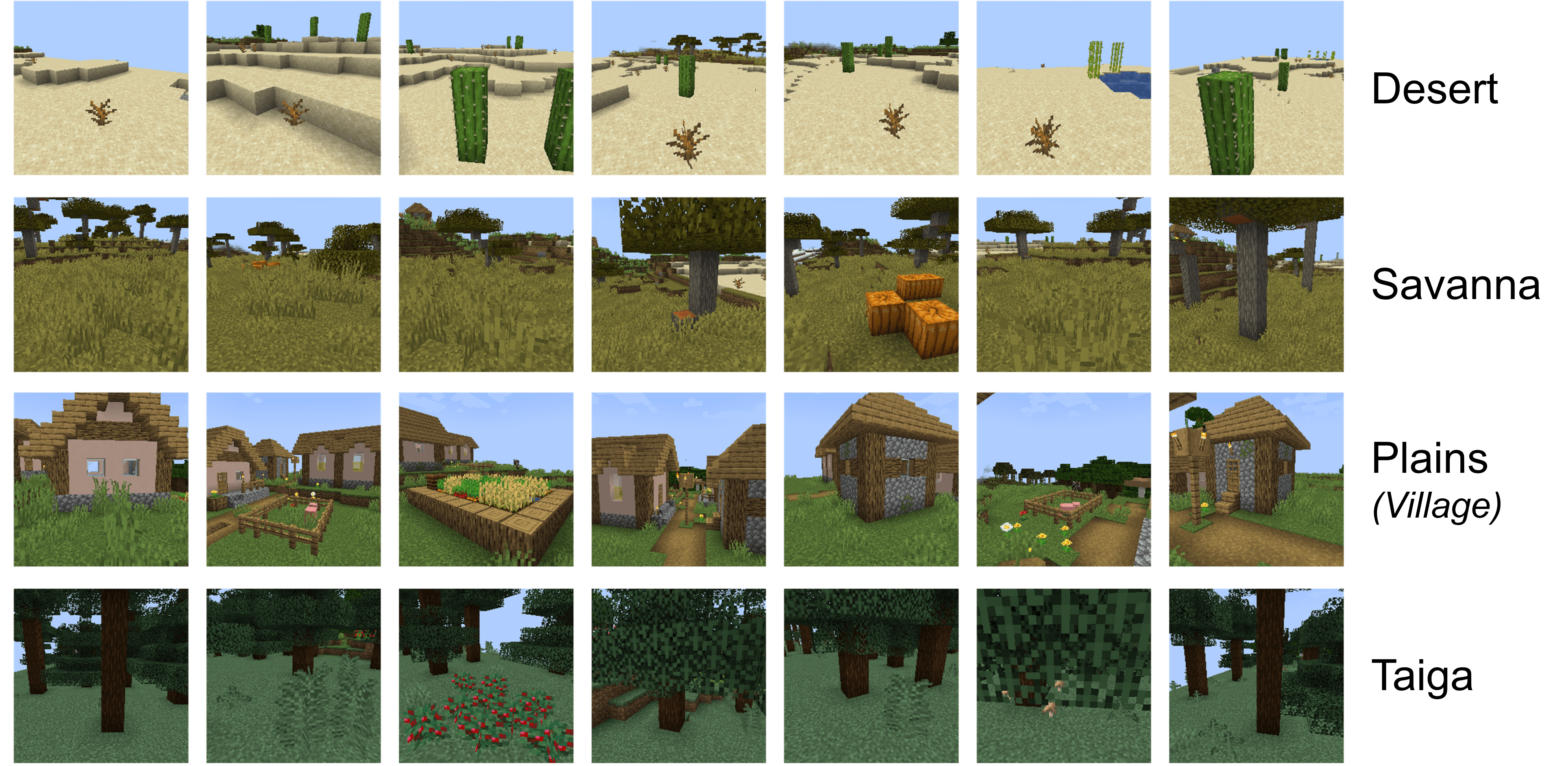}
        \caption{Example scene observations from each biome}
        \label{minecraft-example-obs}
    \end{subfigure}
    \caption{Visualization of Minecraft experiments. (a) We first gather a dataset of object observations from various sub-environments. Observations in different directions are considered as the same object. (b) We validate the OBSER framework in the miniature environment that consists of 10 different biomes: \texttt{snowy\_taiga}, \texttt{taiga}, \texttt{forest}, \texttt{snowy\_plains}, \texttt{plains}, \texttt{swamp}, \texttt{savanna}, \texttt{dark\_forest}, \texttt{desert} and \texttt{jungle}, arranged from top-left to bottom-right. (c) Each biome has distinct environmental characteristics.}
    \label{minecraft-env}
    \vskip -0.5em
\end{figure*}

\begin{table}[t]
\centering
\renewcommand{\arraystretch}{1.1}
    \caption{EDS values of metric learning and SSL models with classification accuracies (Minecraft).}
    \begin{adjustbox}{max width=\columnwidth}
        \begin{tabular}{c||cc|ccc|ccc}
        \Xhline{3\arrayrulewidth}
        \multirow{2}{*}{Model}&\multicolumn{2}{c|}{EDS ($\tau=0.1$)}&\multicolumn{3}{c|}{Mean}&\multicolumn{3}{c}{KNN}\\
        \cline{2-9}
        &del&eps&1&3&5&3&5&7\\
        \hline
        \hline        MoCoV2&0.376&0.005&74.74&82.77&84.10&94.08&94.04&94.04\\
        \hline
        SimCLR&0.267&0.003&76.95&83.61&84.36&95.57&95.30&95.46\\
        \hline
        SupCon&\bf0.782&0.005&\bf86.49&\bf86.56&\bf86.67&\bf96.83&\bf96.83&\bf96.83\\
        \Xhline{3\arrayrulewidth}
        \end{tabular}
    \end{adjustbox}
    \vskip -0.5em
    \label{res-minecraft}
\end{table}

\section{OBSER framework}

In this section, we propose a Bayesian framework through OBSER, which recognizes the environment with episodic memory. In the OBSER framework, the episodic memory $\gM$ is defined as previous observations from distinct spaces. Suppose that an agent has gathered observations $\hat{\mu}_m:=\{x_{mo}\}_{o=1}^{O}$ from its locations $\hat{R}_{m}:=\{p_{mo}\}_{o=1}^{O}$ from a sub-environment $(\mu_m,R_m)\in\gE$. Then the tuple $(\hat{\mu}_m,\hat{R}_{m})$ becomes a new element of the memory $\gM$. Consequently, episodic memory is defined as a set of empirical sub-environments: $\gM:=\{(\hat{\mu}_m,\hat{R}_m)\}_{m=1}^{M}$. The episodic memory enables the agent to i) recall the most probable sub-environment in episodic memory, ii) locate the most similar sub-environment with a given memory, and iii) retrieve the object in such sub-environment.

To validate the OBSER framework, we design two separate experiments: an open-world environment (Minecraft) to validate the completeness and a photo-realistic environment (Replica) to assess the efficacy of the proposed framework. Please refer to Appendix \ref{apdx-minecraft} and \ref{apdx-replica} for details and additional results of each experiment.

\subsection{Open-world environment (Minecraft)}

\paragraph{Minecraft environment}{Minecraft has been widely used as a benchmark testbed for open-world environments~\citep{baker2022video,qin2024mp5,li2024auto,chen2024diffusion}. In Minecraft, various sub-environments called \textit{biomes} mimic natural landscapes, each featuring unique objects. We choose 10 different biomes and gather ego-centric object observations, illustrated in Figure \ref{minecraft-obj-obs}, in each biome to build a dataset. The gathered dataset includes about 26k observations from 25 object classes. Additional information regarding the dataset is presented in Appendix \ref{apdx-minecraft-dataset}.}

\begin{figure}[t]
    \centering
    \begin{subfigure}{\columnwidth}
        \includegraphics[width=\textwidth]{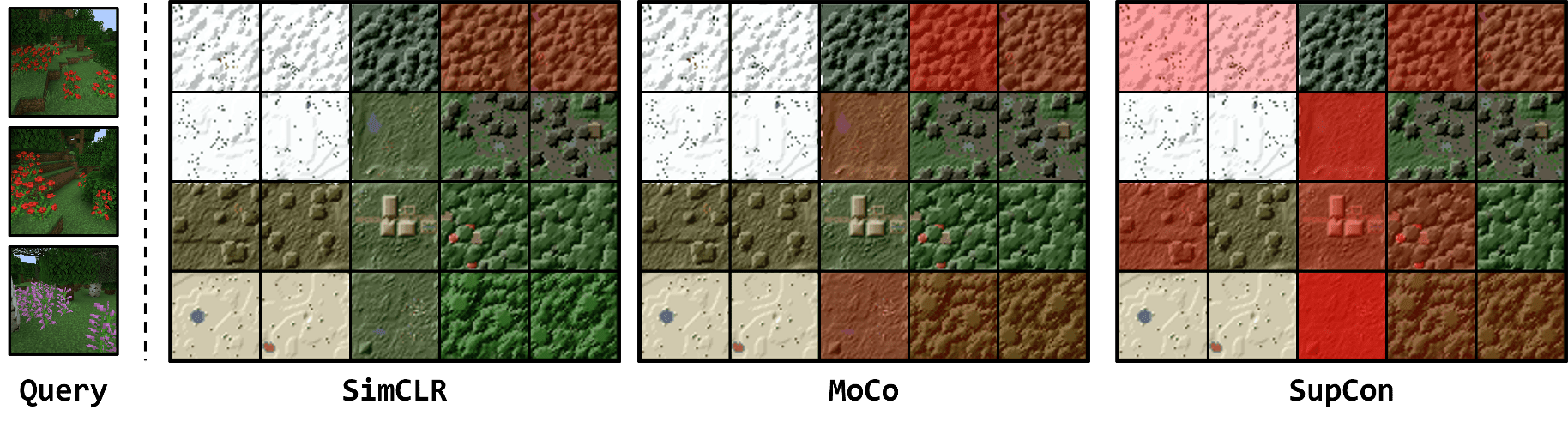}
        \caption{Query (\texttt{Flower} in \texttt{Forest})}
        \label{minecraft-occurrence-a}
    \end{subfigure}
    \vskip 0.2em
    \begin{subfigure}{\columnwidth}
        \includegraphics[width=\textwidth]{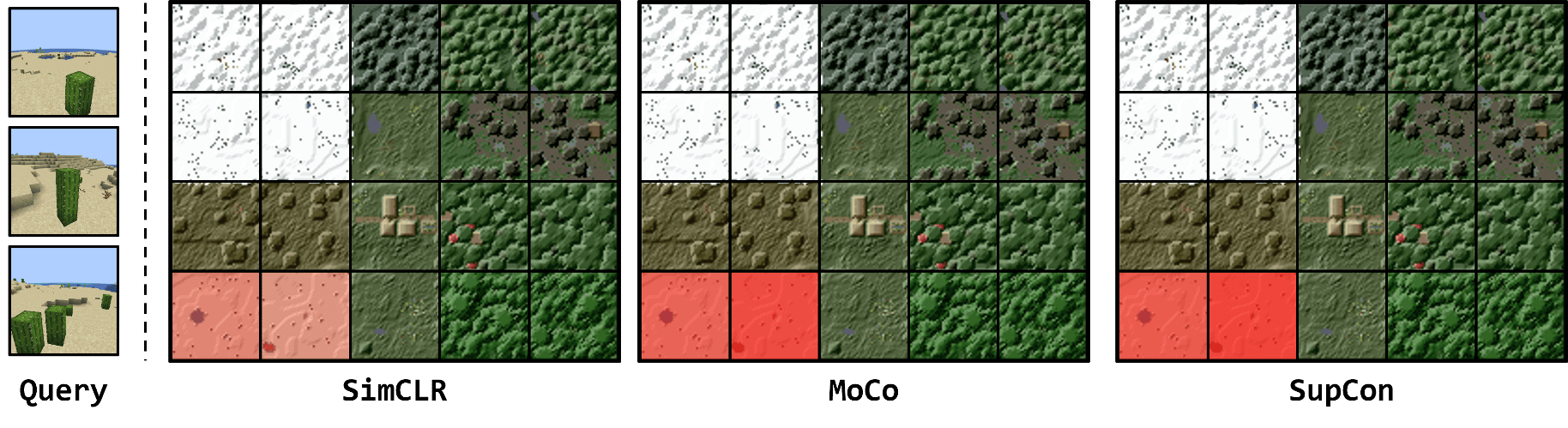}
        \caption{Query (\texttt{Cactus} in \texttt{Desert})}
    \end{subfigure}
    \caption{Heatmaps of object occurrence with the queries in a Miniature environment. Every models can approximate object occurrence well; however, (a, rightmost) SupCon fails to capture the environment-specific information.}
    \label{minecraft-res-obj-retrieval}
\end{figure}

\paragraph{Model description}{SimCLR, MoCo-v2, and SupCon models are used for this experiment. We choose ResNet-50~\citep{he2016deep} as the backbone. The dimension of the embedding space is set to 128, and the temperature $\tau$ is set to 0.2. Each model is trained for 10 epochs as a warm-up. We then train SupCon and SimCLR for 20 epochs, and MoCo for 100 epochs due to its slow learning at the beginning of training. Observations of the same object from different angles offer extra inductive bias~\citep{pantazis2022matching,scherr2022self}. These views are used as positive samples in both metric learning and SSL training.
}

\begin{figure}[t]
    \centering
    \includegraphics[width=\columnwidth]{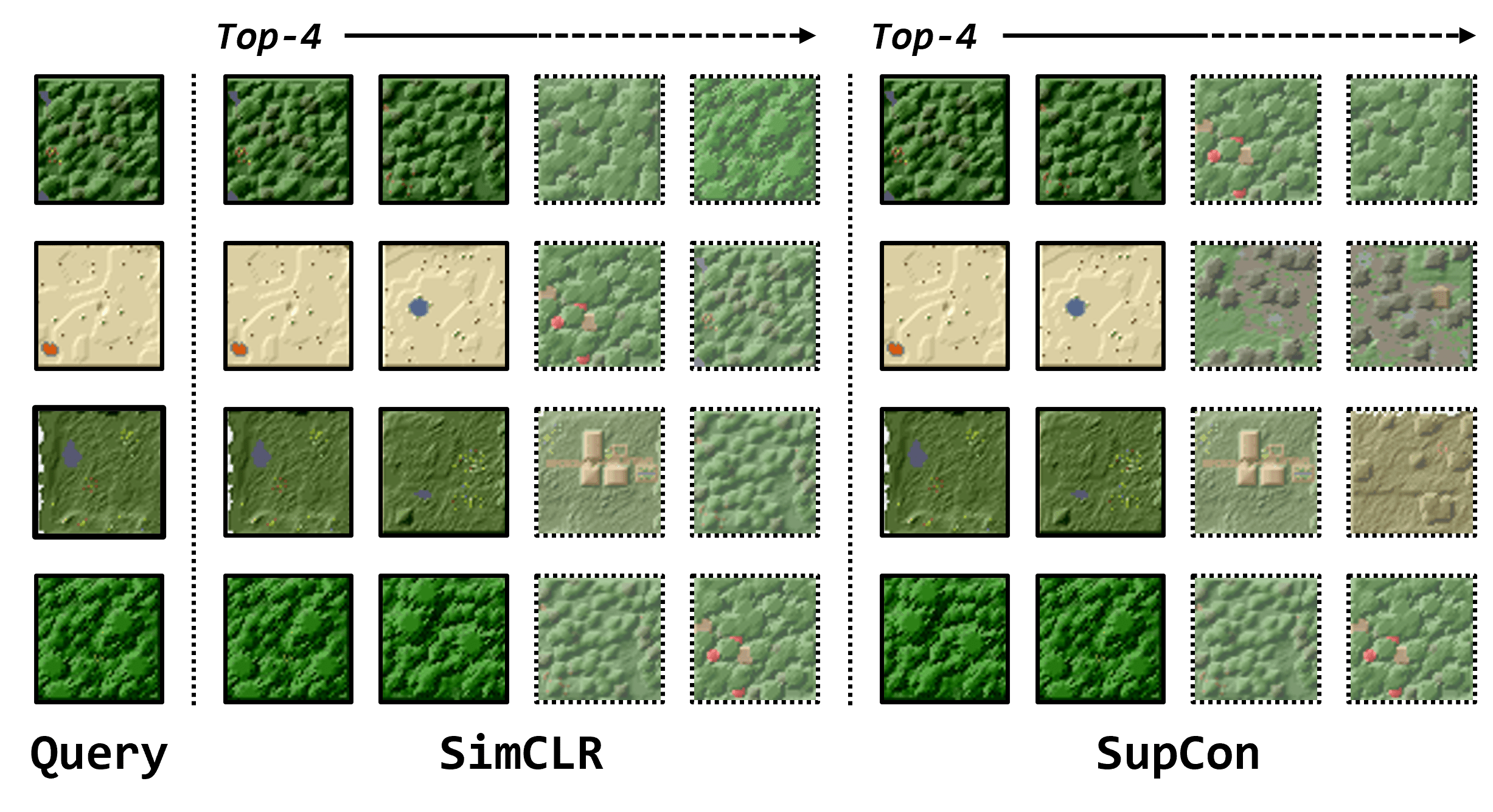}
    \vskip -0.5em
    \caption{Environment-environment retrieval task with Miniature environment. With the queries of observations from 4 different biomes, \texttt{forest}, \texttt{desert}, \texttt{plains} and \texttt{jungle}, the OBSER framework successfully retrieves the similar sub-environments.}
    \vskip -0.5em
    \label{minecraft-res-env-retrieval}
\end{figure}

\begin{table*}[t]
\centering
\renewcommand{\arraystretch}{1.1}
    \caption{Accuracy of chained object retrieval task in Replica environment. We have queried 10 different objects and have computed accuracies with top-5 retrived objects.}
    \vskip -0.5em
    \begin{adjustbox}{max width=.9\textwidth}
        \begin{tabular}{c||cc|cccccc}
        \Xhline{3\arrayrulewidth}
        &CLIP (V)&CLIP (V$+$L)&SupCon&SimCLR&MoCo-v2&MoCo-v3&DINO-v1&DINO-v2\\
        \hline
        Seen (obj-obj)&0.98&\bf1.00&0.98&0.94&0.94&0.82&0.98&\bf1.00\\
        Seen (Top-1 room)&0.24&0.34&0.66&0.42&0.68&0.68&\bf1.00&0.90\\
        Seen (Top-3 rooms)&0.38&0.42&0.76&0.58&0.80&0.74&\bf1.00&0.94\\
        \hline
        Unseen (obj-obj)&0.82&\bf0.96&0.80&0.88&0.80&0.70&0.90&0.88\\
        Unseen (Top-1 room)&0.30&0.32&0.20&0.20&0.62&0.50&0.44&\bf0.78\\
        Unseen (Top-3 rooms)&0.54&0.60&0.46&0.50&0.54&0.64&0.52&\bf0.78\\
        \Xhline{3\arrayrulewidth}
        \end{tabular}
    \end{adjustbox}
    \vskip -1em
    \label{sucess-replica}
\end{table*}

\paragraph{Miniature environment and episodic memory}{
For evaluation, we design a compact version of the generated world, a Miniature environment, which contains all biomes used for training. Figure \ref{minecraft-miniature} and \ref{minecraft-example-obs} show the landscape of the Miniature environment and scene observations from each biome. We gather random observations from each biome in the map to form an episodic memory for evaluating the proposed framework.}

\paragraph{Object-object recognition}{We first measure the classification accuracies of each trained model. In Table \ref{res-minecraft}, SupCon performs better than SSL models in the classification task. Furthermore, it consistently supports our claim regarding the EDS values and task performance.}

\paragraph{Object-environment recognition}{We validate that each model can estimate the object occurrence with the query objects. Figure \ref{minecraft-res-obj-retrieval} shows a heatmap of the occurrence of \texttt{flower} for each grid. When the query provides observations of \texttt{flower} in a \texttt{forest} biome, SSL models leverage ambient information about the environment to inform their inference, whereas metric learning models focus solely on estimating the occurrence of \texttt{flower} without incorporating environmental context.}

\paragraph{Environment-environment recognition}{To validate environment-environment recognition, we design an evaluation task to retrieve similar grids by measuring KL divergence using the observations from each grid. Figure \ref{minecraft-res-env-retrieval} presents the Top-4 grid retrieval results for queries across four biomes—\texttt{forest}, \texttt{desert}, \texttt{plains}, and \texttt{jungle}—demonstrating that all models consistently prioritize grids from the queried biome.}

\subsection{Photo-realistic environment (Replica)}

\paragraph{Replica environment}{Replica environment~\citep{straub2019replica} is a 3D indoor environment with high-resolutional meshes. We additionally separate the environment into 48 distinct rooms. To build an episodic memory, we respectively collect 20 scenes from each room and extract object images from each scene with ground-truth segmentations. Details of gathering observations are provided in Appendix \ref{apdx-replica-env}.
}

\paragraph{Model description}{We employ the same metric and self-supervised learning models as feature extractors as described in Section \ref{sec-val-imagenet}. Specifically, ResNet-50 serves as the backbone for both SimCLR and MoCo-v2, whereas ViT-B is utilized for the remaining models. Additionally, we compare the performance of a trained CLIP model in both scene-based and object-based recognition settings. For inference with CLIP, average cosine similarity is used as the evaluation criterion.}

\paragraph{Chained inference}{Using past observations as episodic memory, we apply the OBSER framework to the object retrieval task with a three-step inference process.}

\vskip 0.5em
\begin{enumerate}[label=(\roman*)]
\item \textbf{Recall relative memory} with a given query object
\vspace*{-0.5em}
\begin{equation}
m^*=\argmax_{m\in\{1,\cdots,M\}}\Phi_f(x_q;\hat\mu_m)
\end{equation}
\item \textbf{Retrieve similar sub-environment} with memory
\vspace*{-0.5em}
\begin{equation}
e^*=\argmin_{e\in\gN_{R}(\hat R_{m^*};\gE)}\widehat{\KL}_f(\hat{\mu}_{m^*}||\mu_{e})
\end{equation}
\item \textbf{Find similar objects} in the sub-environment
\vspace*{-0.5em}
\begin{equation}
p^*=\argmax_{p\in R_{e^*}} \phi_f(x_q,x_p)
\end{equation}
\end{enumerate}

We design the experiments in two settings: in the \textit{Seen} setting, the environment, and the episodic memory are composed of the same set of sub-environments, whereas in the \textit{Unseen} setting, they are mutually exclusive. During sub-environment retrieval in step ii), the framework retrieves top-1 and top-3 rooms for comparison. Additionally, we perform an ablation experiment, denoted as \textit{obj-obj}, which retrieves the object directly from all rooms without environmental inference. Through this experiment, we aim to demonstrate that exploring only a small number of relevant rooms is sufficient to achieve competitive performance through the OBSER framework.

Table \ref{sucess-replica} shows the accuracies of the object retrieval task. For the metric learning model SupCon, competitive performance was achieved in the \textit{Seen} setting, but its inference performance dropped in the \textit{Unseen} setting. This is likely because relying on explicit class information results in a loss of environmental context, thereby reducing its generalizability. Among SSL models, DINO-v2 consistently outperforms the others, especially in \textit{Unseen} setting.

\begin{figure}[t]
    \centering
    \includegraphics[width=0.9\columnwidth]{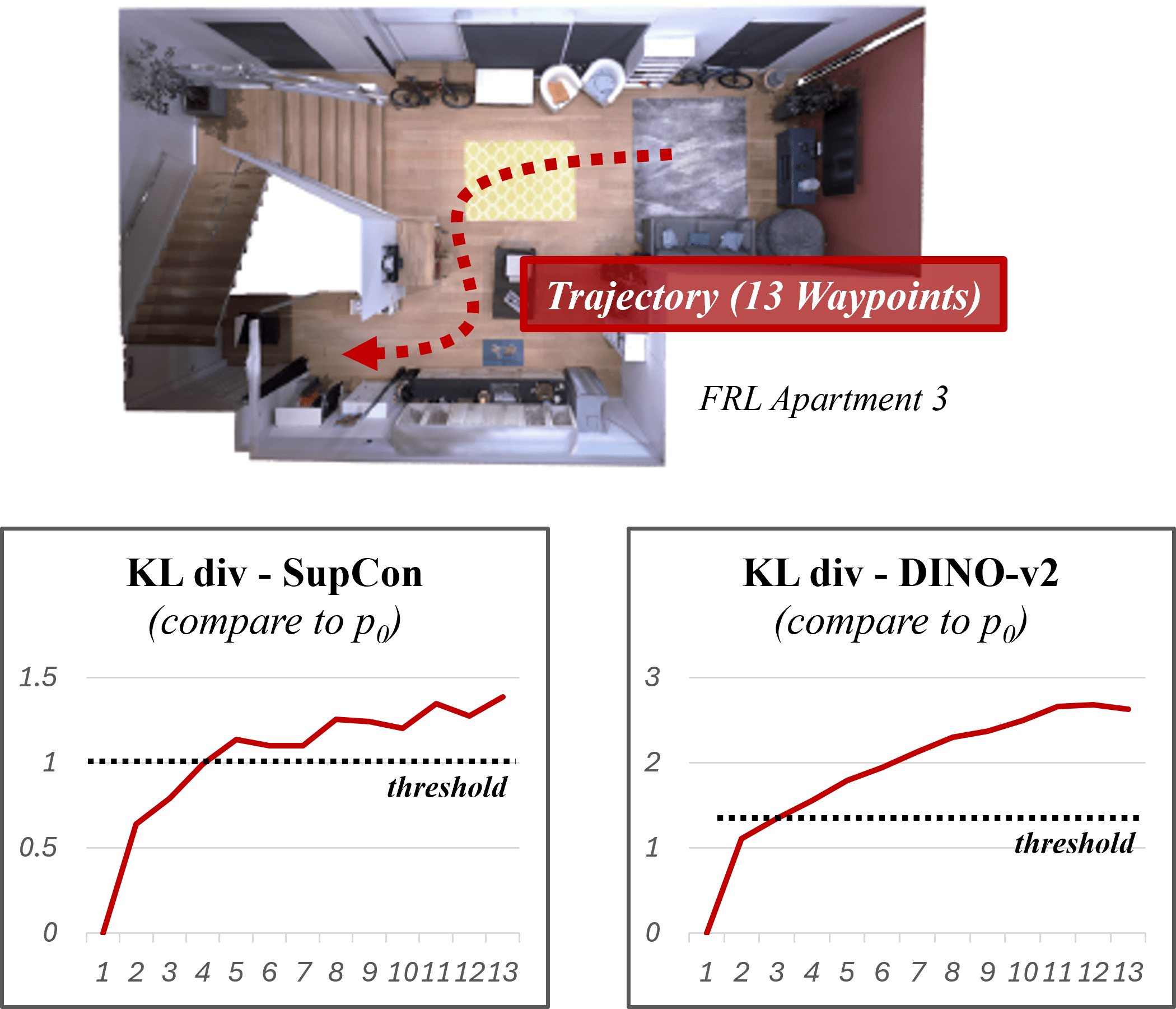}
    \caption{Example trajectory (top) with 13 waypoints and a plot (bottom) showing the KL divergence at each waypoint relative to the first. As the agent's circumstances change along its trajectory, the KL divergence increases, enabling the environment to be segmented into distinct sub-environments at a specific threshold.}
    \label{replica-trajectory}
    \vskip -0.5em
\end{figure}

Another noticeable result is that scene-based recognition using the CLIP model performs worse than the proposed framework in both \textit{Seen} and \textit{Unseen} settings. Augmenting visual observations with language inferences for objects and the environment (CLIP V$+$L) yields higher performance than using only visual observations (CLIP V); however, its performance still falls short of OBSER. For further details on this experiment, we refer the readers to Appendix \ref{apdx-sec-replica-clip}.

\section{Discussion}

\paragraph{Building episodic memory}{To maintain episodic memory within the OBSER framework, we compute the KL divergence estimates across the agent's trajectory. Figure \ref{replica-trajectory} illustrates the KL divergence of the circumstances at successive waypoints along the agent's trajectory, measured relative to the initial point. By selecting an appropriate threshold, the agent can detect meaningful changes in circumstances that delineate distinct spaces, which then form its episodic memory. Further discussion is provided in Appendix \ref{apdx-sec-replica-disc-a}.
}

\begin{figure}[t]
    \centering
    \includegraphics[width=\columnwidth]{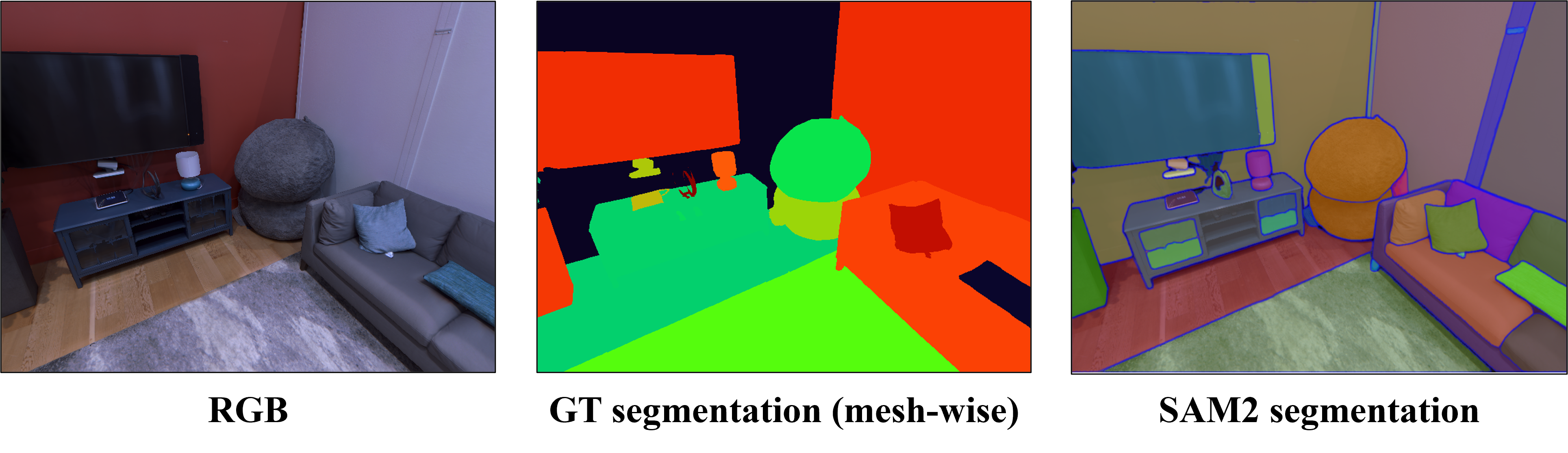}
    \vskip -0.5em
    \caption{Comparison of GT and SAM2 segmentation.}
    \label{SAM-seg}
    \vskip -0.5em
\end{figure}

\paragraph{Recognizing objects}{From an object-based recognition perspective, accurately identifying the objects that make up the environment is essential. We claim that models such as Segment Anything Model~\citep{kirillov2023segment} can enable the OBSER framework to achieve fully unsupervised inference. Figure \ref{SAM-seg} illustrates a comparison between the ground-truth segmentation and the segmentation generated by SAM2.

To validate our claim, we conduct experiments on equivalent chained inference tasks. Table \ref{sam-replica} compares the performance between using ground-truth segmentation and SAM2 segmentation. The results indicate that the OBSER framework with SAM2 segmentation maintains sufficient inference performance, suggesting their potential applicability in real-world scenarios. Qualitative results of this experiment are provided in Appendix \ref{apdx-sec-replica-disc-b}.
}

\begin{table}[t]
\centering
\renewcommand{\arraystretch}{1.1}
    \caption{The average and standard deviation of cosine similarity among retrieved observations for each query object. Object observations are extracted using both GT and SAM2 segmentation.}
    \begin{adjustbox}{max width=\columnwidth}
        \begin{tabular}{c|cc|cc}
        \Xhline{3\arrayrulewidth}
        Backbone&\multicolumn{2}{c|}{DINO-v1-B}&\multicolumn{2}{c}{DINO-v2-B}\\
        \hline
        Segmentation&GT&SAM2&GT&SAM2\\
        \hline
        \hline
        Seen (obj-obj)&0.883$\scriptstyle\pm0.043$&0.836$\scriptstyle\pm0.092$&0.867$\scriptstyle\pm0.053$&0.828$\scriptstyle\pm0.095$\\
        Seen (Top-1)&0.879$\scriptstyle\pm0.046$&0.813$\scriptstyle\pm0.099$&0.835$\scriptstyle\pm0.077$&0.787$\scriptstyle\pm0.106$\\
        Seen (Top-3)&0.882$\scriptstyle\pm0.046$&0.831$\scriptstyle\pm0.097$&0.851$\scriptstyle\pm0.074$&0.803$\scriptstyle\pm0.104$\\
        \hline
        Unseen (obj-obj)&0.824$\scriptstyle\pm0.098$&0.781$\scriptstyle\pm0.118$&0.826$\scriptstyle\pm0.072$&0.795$\scriptstyle\pm0.101$\\
        Unseen (Top-1)&0.637$\scriptstyle\pm0.133$&0.543$\scriptstyle\pm0.073$&0.650$\scriptstyle\pm0.148$&0.563$\scriptstyle\pm0.153$\\
        Unseen (Top-3)&0.676$\scriptstyle\pm0.117$&0.634$\scriptstyle\pm0.100$&0.698$\scriptstyle\pm0.143$&0.621$\scriptstyle\pm0.149$\\
        \Xhline{3\arrayrulewidth}
        \end{tabular}
    \end{adjustbox}
    \label{sam-replica}
\end{table}

\section{Conclusion}
In this paper, we introduced a novel probabilistic framework for object-based sub-environment recognition. By leveraging kernel density estimation and metric learning models, our approach quantifies essential relationships between objects and their environments. The OBSER framework effectively enables agents to handle retrieval tasks in both open-world and photorealistic settings, as supported by qualitative results. Ultimately, this work suggests that agents can perform fully unsupervised, zero-shot inferences by chaining these relationships. This represents a promising step forward in autonomous environment understanding.


{\small
\bibliographystyle{ieee_fullname}
\bibliography{egbib}
}

\clearpage

\appendix
\onecolumn
\maketitlesupplementary

\setcounter{thm}{0}
\setcounter{coro}{0}

\section{Theoretical Details and Proofs of EDS function}

\subsection{Additional Details and Proofs}
\subsubsection{Additional Definitions}
\paragraph{Membership function}{
A membership function $\hat\mu_f(c|x)$ with a mapping function $f:X\rightarrow Z$ is defined as follows:}
\begin{equation*}
\hat\mu_f(c|x),\enspace \sumC \hat\mu_f(c|x)=1.
\end{equation*}
By this definition, when the function $f$ is optimal, the membership function works as a partition indicator.
\begin{equation}
I(c|x)=\mu(c|x)=\nu(c|x)=
    \begin{cases}1\quad x\in\gX_{c}\\0\quad O.W.\end{cases}
\label{eq-indicator}
\end{equation}
With \cref{eq-indicator} and the definition of the sub-environment, the following equation is satisfied with every function $f$.
\begin{equation}
\exptxp{\mu}\left[f(\cdot)\cdot I(c|x)\right]=\omega(c)\cdot\exptxp{\rho_c}\left[f(\cdot)\right]
\end{equation}

\paragraph{Bayesian classifier}{
With a kernel $\phi_f: X\times X\rightarrow [0,1]$, a Bayesian classifier $\hat\mu_f(c|x)$ with a data $x$ can be derived via kernel density estimation:}
\begin{gather}
\hat\mu_f(c|x)=\frac{1}{Z}\cdot\exptxp{\mu}\left[\phi_f(x,x')I(c|x)\right],\\
Z=\sumC\exptxp{\mu}\left[\phi_f(x,x')I(c|x)\right].\\
    \begin{aligned}
    \therefore \hat\mu_f(c|x)&=\frac{\exptxp{\mu}\left[\phi_f(x,x')I(c|x)\right]}{\sumC\exptxp{\mu}\left[\phi_f(x,x')I(c|x)\right]}\\
    &=\frac{\omega(c)\cdot\exptxp{\rho_c}\left[ \phi_f(x,x')\right]}{\sumCp\omega(c)\cdot\exptxp{\rho_cp}\left[ \phi_f(x,x')\right]}.\\
    &=\frac{\omega(c)\Phi_f(x;\rho_c)}{\sumC \omega(c)\Phi_f(x;\rho_c)}=\frac{\omega(c)\Phi_f(x;\rho_c)}{\Phi_f(x;\mu)}.
    \label{eq-message-passing}
    \end{aligned}
\end{gather}

\subsubsection{KL divergence between \texorpdfstring{$\mu(c,x)$}{p(c,x;μ)} and \texorpdfstring{$\hat\mu_f(c,x)$}{p(c,x;μ,F)}}

\begin{lemma}[KL divergence between $\mu(c,x)$ and $\hat\mu_f(c,x)$]{}
\begin{equation}
\KL(\mu(c,x)\Vert\hat{\mu}_f(c,x))=\\\exptx{\mu}\left[-\log \frac{\Phi_f(x;\rho_+)}{\Phi_f(x;\mu)}\right]+\mathcal{H(\omega)}.
\end{equation}
\label{lemma1}
\end{lemma}
\begin{proof}
\begin{align}
\KL(\mu(c,x)\Vert&\hat{\mu}_f(c,x))\\
&=\sumC\int-\mu(c,x)\log\frac{\hat\mu_f(c,x)}{\mu(c,x)}dx\\
&=\sumC\omega(c)\cdot\int-\log\frac{\hat\mu_f(c,x)}{\mu(c,x)}d\rho_c(x)\\
&=\sumC\omega(c)\cdot\int-\log\frac{\hat\mu_f(c|x)\mu(x)}{\mu(x)}d\rho_c(x)\\
&=\sumC\omega(c)\cdot\exptx{\rho_c}\left[-\log \hat\mu_f(c|x)\right] \quad\quad\cdots\:\text{(Cross Entropy)}
\end{align}
By using Bayesian classifier in \cref{eq-message-passing},
\begin{align}
\sumC\omega(c)&\cdot\exptx{\rho_c}\left[-\log \hat\mu_f(c|x)\right]\\
&=\sumC\omega(c)\cdot\exptx{\rho_c}\left[-\log \frac{\omega(c)\Phi_f(x;\rho_c)}{\Phi_f(x;\mu)}\right]\\
&=\sumC\omega(c)\cdot\exptx{\rho_c}\left[-\log \frac{\Phi_f(x;\rho_c)}{\Phi_f(x;\mu)}\right]+\gH(\omega)\\
&=\exptx{\mu}\left[-\log \frac{\Phi_f(x;\rho_+)}{\Phi_f(x;\mu)}\right]+\gH(\omega)
\end{align}
\begin{equation}
\therefore \KL(\mu(c,x)\Vert\hat{\mu}_f(c,x))=\exptx{\mu}\left[-\log \frac{\Phi_f(x;\rho_+)}{\Phi_f(x;\mu)}\right]+\gH(\omega)
\end{equation}
$\because \KL(\cdot\Vert\cdot)\ge0,$
\begin{equation}
\exptx{\mu}\left[-\log \frac{\Phi_f(x;\rho_+)}{\Phi_f(x;\mu)}\right]\ge-\gH(\omega).
\end{equation}
The equality is satisfied with $\mu(c,x)=\hat\mu_f(c,x).$
\end{proof}

\subsection{Optimization of (\texorpdfstring{$\epsilon,\delta$}{ε,δ}) Statistically Separable (EDS) function \texorpdfstring{$f$}{}}
\label{apdx-eds-opt}
We define the optimization of the EDS function as fitting the estimated joint distribution $\hat\mu_f(c,x)$ to the ground-truth joint distribution $\mu(c,x)$. Thus, we formulate the optimization problem of the EDS function as minimizing the KL divergence between $\hat\mu_f(c,x)$ and $\mu(c,x)$. By tightening the bound of KL divergence, the EDS function is optimized in terms of both separability and concentration properties.

\begin{thm}[Bayesian metric learning] For an EDS function $f$, let $\exists k\ge 1, \delta=k\cdot\epsilon$. An upperbound $\Delta\gH$ of $\:\KL(\mu(c,x)\Vert\hat{\mu}_f(c,x))$ is derived as:
\begin{equation}
\KL(\mu(c,x)\Vert\hat{\mu}_f(c,x))\le\log\bigl(1+({\lvert\gC\rvert-1})/{k}\bigr):=\Delta\gH,
\end{equation}
and if $\Delta\gH\rightarrow+0$, then $k\rightarrow\infty$.
\end{thm}
\begin{proof}
With \cref{lemma1},
\begin{align}
\KL(\mu(c,x)\Vert&\hat{\mu}_f(c,x))\\
&=\exptx{\mu}\left[-\log \frac{\Phi_f(x;\rho_+)}{\Phi_f(x;\mu)}\right]+\gH(\omega)\\
&=\sumC\omega(c)\cdot(\exptx{\mathcal{D}_c}\left[-\log\Phi_f(x,\rho_c)\right]+\exptx{\mathcal{D}_c}\left[\log\Phi_f(x,\mu)\right])+\mathcal{H}(\mathcal{C};\mu)\\
&\le\sumC\omega(c)\cdot(-\log\delta+\log(\omega(c)\cdot\delta+\epsilon\cdot(1-\omega(c)))+\mathcal{H}(\omega)\\
&=\sumC\omega(c)\cdot\log(1+\frac{\epsilon\cdot(1-\omega(c))}{\delta\cdot\omega(c)})\\
&\le \log(1+\frac{\epsilon\cdot(|\mathcal{C}|-1)}{\delta})= \log(1+\frac{(|\gC|-1)}{k}):=\Delta\gH.
\end{align}
By rewriting $k$ in terms of $\Delta\gH$, we can get $k = \frac{|\gC|-1}{\exp(\Delta\gH)-1}$. Thus, if $\Delta\gH\rightarrow +0$ then $k\rightarrow\infty$.
\end{proof}
Note that the optimization with $KL(\mu\Vert\hat\mu_f)$ becomes equivalent to optimization of InfoNCE~\citep{oord2018representation} by setting the kernel $\phi_f(x,x')$ as follows:
\begin{gather}
\phi_f(x,x')=\exp\{(f(x)^{\top}f(x')-1)/\tau\},\nonumber\\ \phi_f(x,x')\in[\exp(-2/\tau),1],
\label{kernel-hypersphere}
\end{gather}
with a temperature parameter $\tau$. Note that optimizing the EDS function in Euclidean space, which has no upper bound on the distance, does not guarantee the improvement of the concentration property. Therefore, we claim that the embedding space should be compact to guarantee the convergence of both $\epsilon$ and $\delta$.

\begin{coro}[Compact space] Let $d_{Z}: Z\times Z\rightarrow [0, d_{\max}]$, monotonic deacreasing $h: \mathbb{R}_+\rightarrow[0,1]$ with $h(0)=1$.
Without any additional restrictions of $Z$,
$\Delta\gH \rightarrow \Delta\gH_{\min}$, and
$\delta\rightarrow 1,\:\epsilon\rightarrow \phi_{\min}$ for some $\Delta\gH_{\min}$ and $\phi_{\min}$.
\end{coro}
\begin{proof}
Since the metric is bounded with $[0,d_{\max}]$, $\delta$ and $\epsilon$ also bounded to $[\phi_{\min},1]$ with $\phi_{\min}=h(d_{\max})$. Thus, $\exists k_{\max}=1/\phi_{\min},\enspace k=(\delta/\epsilon)\in[1,k_{\max}]$. By Theorem \ref{thm1}, we can get:
\begin{gather}
\KL(\mu(c,x)\Vert\hat{\mu}_f(c,x))\le\log\left(1+\frac{|\gC|-1}{k}\right)=\Delta\gH.\\
\exists(\Delta\gH_{\min})=\log(1+(|\gC|-1)/k_{\max})=\log(1+(|\gC|-1)\cdot \phi_{\min})\le\Delta\gH,\\
\therefore \Delta\gH\rightarrow \Delta\gH_{\min},k\rightarrow k_{\max},\quad \delta\rightarrow 1,\:\epsilon \rightarrow \phi_{\min}.
\end{gather}
\end{proof}
Note that we can set $\phi_{\min} \simeq 0$ by choosing appropriate kernel function $h$.

\subsubsection{Validation with toy example}
\paragraph{Toy problem}{
We first validate the EDS function and its behavior during the optimization with several elementary environments: Moons and XOR datasets. We choose the hypersphere as the embedding space. A shallow MLP structure is chosen as the feature extractor for both experiments: the number of nodes for each layer is selected as (2,8,4,2), respectively. The feature extractor is trained with a balanced set in 30 epochs. With the trained representations, we measure $\epsilon$ and $\delta$, and show that the proposed approximated measures are accurate with an optimized EDS function.}

\begin{figure}[t]
\centering
    \begin{subfigure}{0.20\textwidth}
        \includegraphics[width=\textwidth]{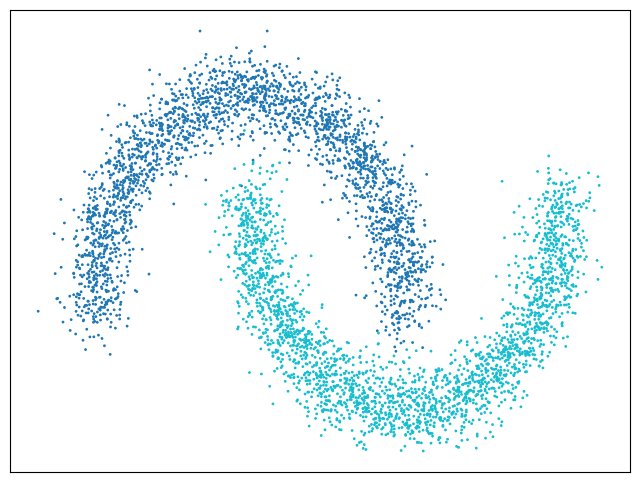}
        \caption{Data distribution}
    \end{subfigure}%
    \hfill
    \begin{subfigure}{0.16\textwidth}
        \includegraphics[width=\textwidth]{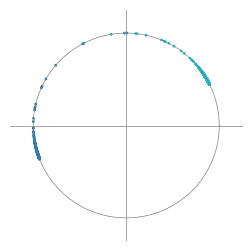}
        \caption{Representation}
    \end{subfigure}%
    \hfill
    \begin{subfigure}{0.20\textwidth}
        \includegraphics[width=\textwidth]{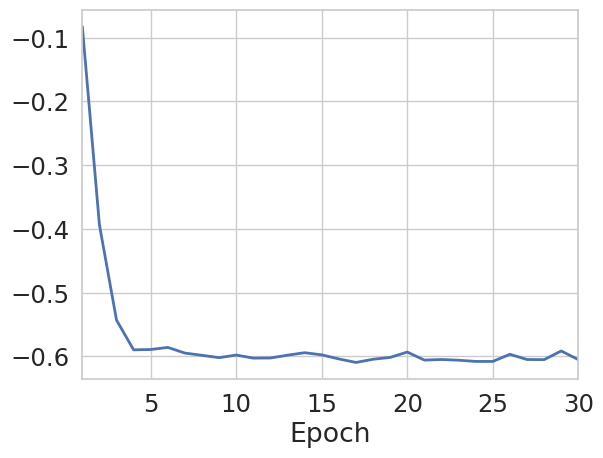}
        \caption{loss function}
    \end{subfigure}%
    \hfill
    \begin{subfigure}{0.20\textwidth}
        \includegraphics[width=\textwidth]{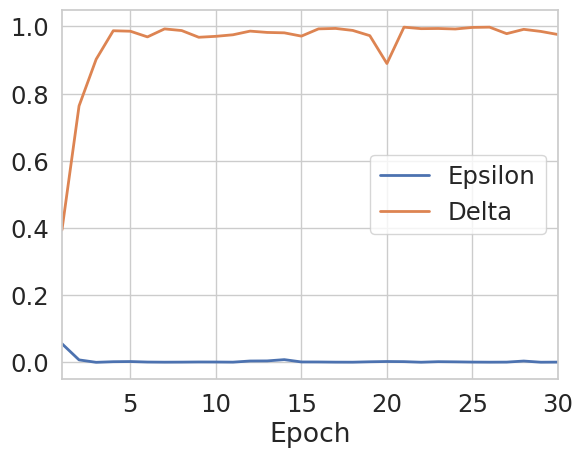}
        \caption{($\epsilon,\delta$) values (Hypersphere)}
    \end{subfigure}%
    \hfill
    \begin{subfigure}{0.20\textwidth}
        \includegraphics[width=\textwidth]{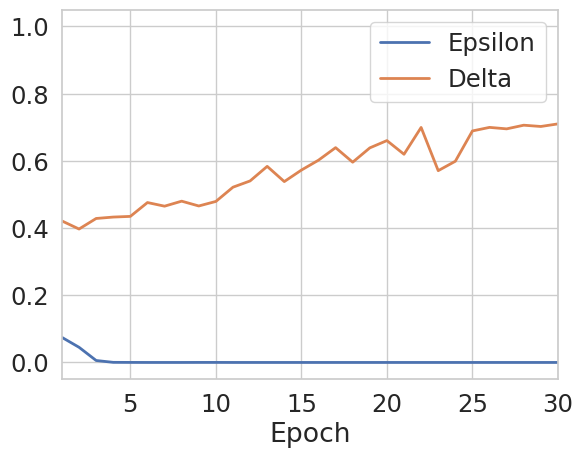}
        \caption{($\epsilon,\delta$) values (Euclidean)}
    \end{subfigure}
    \caption{Visualization of the optimization of the EDS function with (a) Moons dataset. (b-d) The EDS function defined on a hypersphere is optimized succesfully with high concentration ($\delta\simeq 1$) and high separability ($\epsilon\simeq 0$) . (e) On the other hand, with the EDS functions with Euclidean space, $\delta$ cannot reach to 1 because Euclidean space is not a compact space.}
    \vskip -0.5em
    \label{vis-moons}
\end{figure}

\cref{vis-moons} shows the visualization of the optimization of $f$ with the Moons dataset. After the optimization, the trained feature extractor successfully maps the data onto a 2-dimensional sphere. During the optimization, $\epsilon$ converges to 0 while $\delta$ converges to 1 as the training loss is minimized. This implies that the $\epsilon$ and $\delta$ values accurately represent the robustness of arbitrary feature extractors.

\begin{table}[ht]
\centering
\renewcommand{\arraystretch}{1.1}
    \caption{Validation with toy datasets. (3 times, mean values are reported) $S_1$: $[0.5,0.5]\rightarrow[0.1,0.9]$, $S_2$: $[0.2,0.8]\rightarrow[0.8,0.2]$}
    \begin{adjustbox}{max width=0.8\textwidth}
        \begin{tabular}{c|c|ccc|cc|cc}
        \Xhline{3\arrayrulewidth}
        Dataset&Model&$\epsilon\:(\downarrow)$&$\delta\:(\uparrow)$&$\log(\delta/\epsilon)$&$\widehat{\KL}_f(S_{1})$&$|\Delta \KL(S_{1})|$&$\widehat{\KL}_f(S_{2})$&$|\Delta \KL(S_{2})|$\\
        \hline
        \hline
        \multirow{3}{*}{Moons}&Norm&0.0005&0.0975&5.3098&0.4840&0.0268&0.8152&0.0166\\
        &EDS-e&0.0000&0.6174&\bf8.8720&0.5080&0.0030&0.8330&0.0046\\
        &EDS-s&0.0011&\bf0.9922&6.8669&0.5034&0.0075&0.8267&0.0051\\
        \hline
        \multirow{3}{*}{XOR}&Norm&$\sim0$&0.2136&27.0365&0.5163&0.0030&0.8295&0.0029\\
        &EDS-e&$\sim0$&0.6810&\bf53.7363&0.5111&0.0039&0.8271&0.0046\\
        &EDS-s&$\sim0$&\bf0.9991&28.5699&\bf0.5108&\bf0.0000&\bf0.8308&\bf0.0000\\
        \Xhline{3\arrayrulewidth}
        \end{tabular}
    \end{adjustbox}
    \label{apdx-toy-result}
\end{table}

\paragraph{Ablation. the role of the compact space}{To illustrate the difference between Euclidean space and hypersphere, we conduct toy experiments using two small datasets: Moons and XOR. For each dataset, we train models in both Euclidean and hypersphere embedding spaces, then compute the $\epsilon$ and $\delta$ values with a temperature of $\tau = 0.07$. The results are shown in \cref{apdx-toy-result}. In Euclidean space, the metric between data points is unbounded, hindering the convergence of $\delta$ to 1. In contrast, on the hypersphere, $\delta$ converges to nearly 1, while $\log(k)$ remains smaller than 1 in Euclidean space. This convergence of $\delta$ also enhances the precision of KL divergence. These results justify the use of the hypersphere as the embedding space for our experiments.}

\section{Object-Based Sub-Environment Recognition (OBSER) with the EDS Function}
\label{apdx-ser-eds}
\subsection{Object Occurrence estimation with EDS function}
\begin{lemma}[Object Occurrence with EDS function] For an EDS function $f$, the proposed measure has the bound of:
\begin{equation}
\hat\omega(c):=\Phi_f(x;\mu),\quad\delta\le\frac{\hat\omega(c)}{\omega^{\mu}(c)+\epsilon\cdot(1-\omega^{\mu}(c))}
\le1.
\end{equation}
\label{object-occurrence-eds}
\end{lemma}
\begin{proof}
By the definition of the EDS function,
\begin{gather}
\delta\cdot(\omega^{\mu}(c)+\epsilon\cdot(1-\omega^{\mu}(c)))\le\hat\omega(c)\le\omega^{\mu}(c)+\epsilon\cdot(1-\omega^{\mu}(c))\\
\therefore \delta\le\frac{\hat\omega(c)}{\omega^{\mu}(c)+\epsilon\cdot(1-\omega^{\mu}(c))}
\le1.
\end{gather}
In the case of $\delta\rightarrow1$, $\epsilon\rightarrow0$, $\hat\omega(c)\rightarrow\omega^{\mu}(c)$.
\end{proof}

\subsection{KL divergence btw two distributions}
By the definition of sub-environment, KL divergence between the distributions $\mu(c,x)$ and $\nu(c,x)$ of sub-environments is computed as:
\begin{align*}
\KL(\mu(c,x)\Vert\nu(c,x))&=\exptx{\mu}\left[\log\frac{\mu(c,x)}{\nu(c,x)}\right]\\
&=\sumC\omega^{\mu}(c)\cdot\exptx{\rho_c}\left[\log\frac{\sum_{c'}\omega^{\mu}(c')\rho_{c'}(x)}{\sum_{c'}\omega^{\nu}(c')\rho_{c'}(x)}\right]\\
&=\sumC\omega^{\mu}(c)\cdot\exptx{\rho_c}\left[\log\frac{\sum_{c'}\omega^{\mu}(c')\rho_{c'}(x)\rho_{c'}(x)}{\sum_{c'}\omega^{\nu}(c')\rho_{c'}(x)\rho_{c'}(x)}\right].
\end{align*}
Because $\forall c'\neq c,\rho_{c'}(x)\cdot\rho_{c'}(x)=0$,
\begin{equation}
=\sumC\omega^{\mu}(c)\cdot\log\frac{\omega^{\mu}(c)}{\omega^{\nu}(c)}.
\end{equation}

However, since we assume that the latent class $c$ is not directly accessible, we need an nonparametric method to approximate $\KL(\hat\mu_f(c,x)\Vert \hat\nu_f(c,x))$. First we define the pseudo-divergence $D_{\mu}(\hat\mu_f(c,x)\Vert \hat\nu_f(c,x))$:
\begin{equation}
D_{\mu}(\hat\mu_f(c,x)\Vert \hat\nu_f(c,x)):=\sumC\left(\omega^{\mu}(c)\cdot\exptx{\rho_c}\left[\log\frac{\hat\mu_f(c,x)}{\hat\nu_f(c,x)}\right]\right).\\
\end{equation}
By rewriting the pseudo-divergence, we obtain the kernel density based KL divergence $\widehat{\KL}_f(\mu||\nu)$ without using the latent class.

\begin{lemma}[Property of pseudo KL divergence]
\begin{equation}
D_{\mu}(\hat\mu_f(c,x)\Vert \hat\nu_f(c,x))=-\underbrace{\exptx{\mu}\left[\log \frac{\Phi_f(x;\mu)}{\Phi_f(x;\nu)}\right]}_{\widehat{\KL}_f(\mu||\nu)}+ 2\cdot \KL(\mu\Vert\nu)
\end{equation}
\end{lemma}
\begin{proof}
\begin{gather}
D_{\mu}(\hat\mu_f(c,x)\Vert \hat\nu_f(c,x)):=\sumC\left(\omega^{\mu}(c)\cdot\exptx{\rho_c}\left[\log\frac{\hat\mu_f(c,x)}{\hat\nu_f(c,x)}\right]\right)\\
=\KL(\mu(c,x)\Vert \hat\nu_f(c,x))-\KL(\mu(c,x)\Vert \hat\mu_f(c,x))
\end{gather}
i) $\KL(\mu(c,x)\Vert \hat\nu_f(c,x))$

By utilizing the same method as in \cref{lemma1},
\begin{align}
\KL(\mu(c,x)&\Vert \hat\nu_f(c,x)) \nonumber\\
&=\sumC\omega^{\mu}(c)\cdot\exptx{\rho_c}\left[-\log \hat\nu_f(c|x)\right]+\KL(\mu\Vert\nu) \nonumber\\
&=\sumC\omega^{\mu}(c)\cdot\exptx{\rho_c}\left[-\log \omega^{\nu}(c)-\log \frac{\Phi_f(x;\rho_c)}{\Phi_f(x;\nu)}\right]+\KL(\mu\Vert\nu) \nonumber\\
&=\exptx{\mu}\left[-\log \frac{\Phi_f(x;\rho_{+})}{\Phi_f(x;\nu)}\right]+\KL(\mu\Vert\nu)+\text{CE}(\mu\Vert\nu).
\end{align}
ii) $\KL(\mu(c,x)\Vert \hat\mu_f(c,x))$ (\cref{lemma1})
\begin{equation}
\KL(\mu(c,x)\Vert \hat\mu_f(c,x))=\exptx{\mu}\left[-\log\frac{\Phi_f(x;\rho_{+})}{\Phi_f(x;\mu)}\right]+\mathcal{H}(\omega^\mu).
\end{equation}
Putting together,
\begin{gather}
\therefore D_{\mu}(\hat\mu_f(c,x)\Vert \hat\nu_f(c,x))\\
\begin{aligned}
=\exptx{\mu}&\left[-\log\frac{\Phi_f(x;\rho_{+})}{\Phi_f(x;\nu)}\right]+\KL(\mu\Vert\nu)+\text{CE}(\mu\Vert\nu)\\
&-\exptx{\mu}\left[-\log\frac{\Phi_f(x;\rho_{+})}{\Phi_f(x;\mu)}\right]-\mathcal{H}(\omega^\mu)
\end{aligned}
\\=2\cdot \KL(\mu\Vert\nu)-\widehat{\KL}_f(\mu||\nu).
\end{gather}
\end{proof}

Intuitively, when $f\rightarrow f^*$ with $\mu(c,x)=p(c,x;\mu,f^*)$, then $D_{\mu}(\hat\mu_f(c,x)\Vert \hat\nu_f(c,x))\rightarrow \KL(\mu\Vert\nu)$ is satisfied. Therefore, we can say $\widehat{\KL}_f(\mu||\nu)\rightarrow \KL(\mu\Vert\nu)$ when $f\rightarrow f^*$.

\subsection{Approximation on KL divergence with EDS function}
\begin{thm}[KL divergence with the EDS function] For an EDS function $f$, the proposed measure in Definition \ref{KL-div-estimation} has the bound of:
\begin{equation}
\left|\widehat{\KL}_f(\mu||\nu)-\sumC\omega^{\mu}(c)\cdot\log\left(\frac{\omega^{\mu}(c)+(1-\omega^{\mu}(c))\cdot\epsilon}{\omega^{\nu}(c)+(1-\omega^{\nu}(c))\cdot\epsilon}\right)\right|\le-\log\delta,
\end{equation}
in $(\mu,\nu)$-almost everywhere. With optimized $f$, such that $\delta\rightarrow 1, \epsilon\rightarrow 0$, $\widehat{\KL}_f(\mu||\nu)$ converges to $\KL(\mu\Vert\nu)$.
\label{KL-div-eds}
\end{thm}
\begin{proof}
\begin{gather}
\exptx{\mu}\left[\log \frac{\Phi_f(x;\mu)}{\Phi_f(x;\nu)}\right]\\
=\sumC\omega^{\mu}(c)\cdot\left(\exptx{\rho_c}\left[\log \frac{\sumCp\omega^{\mu}(c')\cdot\Phi_f(x;\rho_{c'})}{\sumCp\omega^{\nu}(c')\cdot\Phi_f(x;\rho_{c'})}\right]\right)\\
\le\sumC\omega^{\mu}(c)\cdot\log \left(\exptx{\rho_c}\left[\frac{\sumCp\omega^{\mu}(c')\cdot\Phi_f(x;\rho_{c'})}{\sumCp\omega^{\nu}(c')\cdot\Phi_f(x;\rho_{c'})}\right]\right)\\
\le\sumC\omega^{\mu}(c)\cdot\log\left(\frac{\omega^{\mu}(c)+(1-\omega^{\mu}(c))\cdot\epsilon}{\omega^{\nu}(c)+(1-\omega^{\nu}(c))\cdot\epsilon}\right)-\log\delta.\\
\end{gather}
In a same way,
\begin{gather}
\exptx{\mu}\left[\log \frac{\Phi_f(x;\mu)}{\Phi_f(x;\nu)}\right]\\
=-\sumC\omega^{\mu}(c)\cdot\left(\exptx{\rho_c}\left[\log \frac{\sumCp\omega^{\nu}(c')\cdot\Phi_f(x;\rho_{c'})}{\sumCp\omega^{\mu}(c')\cdot\Phi_f(x;\rho_{c'})}\right]\right)\\
\ge-\sumC\omega^{\mu}(c)\cdot\log \left(\exptx{\rho_c}\left[\frac{\sumCp\omega^{\nu}(c')\cdot\Phi_f(x;\rho_{c'})}{\sumCp\omega^{\mu}(c')\cdot\Phi_f(x;\rho_{c'})}\right]\right)\\
\ge-\sumC\omega^{\mu}(c)\cdot\log\left(\frac{\omega^{\nu}(c)+(1-\omega^{\nu}(c))\cdot\epsilon}{\omega^{\mu}(c)+(1-\omega^{\mu}(c))\cdot\epsilon}\right)+\log\delta\\
\ge\sumC\omega^{\mu}(c)\cdot\log\left(\frac{\omega^{\mu}(c)+(1-\omega^{\mu}(c))\cdot\epsilon}{\omega^{\nu}(c)+(1-\omega^{\nu}(c))\cdot\epsilon}\right)+\log\delta.
\end{gather}
\end{proof}

Note that when $\delta\rightarrow1$ and $\epsilon\rightarrow0$, the measure $\widehat{\KL}_f(\mu||\nu)$ converges to $\KL(\mu\Vert\nu)$.
\subsection{Algorithms of the proposed measures}

\begin{algorithm}[!ht]
    \small
    \begin{algorithmic}
        \PyCode{def occurrence\_estimation(query,mu,tau,type\_='cosine',multiplier=0.25):}
        \INDB{1}
            \PyCode{mean = query.mean(0,keepdim=True)}
            \PyCode{if metric == 'cosine':}
            \INDB{2}
                \PyCode{mean = mean / mean.norm(-1,keepdim=True)}
            \INDE
            \PyCode{mean\_mu\_matrix = dist\_matrix(mu,mean,metric)}
            \PyCode{mean\_mu\_density = kernel(mean\_mu\_matrix,tau,metric).mean(-1)}
            \PyCode{mean\_query\_matrix = dist\_matrix(query,mean,metric)}
            \PyCode{tol = kernel(mean\_query\_matrix,tau,metric).mean() * multiplier}
            \PyCode{mean\_mu\_density = (mean\_mu\_kernel > tol).double()}
            \PyCode{return mean\_mu\_density.mean(0)}
        \INDE
    \end{algorithmic}
    \caption{PyTorch-style pseudocode of object occurrence estimation.}
    \label{apdx-existence-algorithm}
\end{algorithm}

\begin{algorithm}[!ht]
    \small
    \begin{algorithmic}
        \PyCode{def kldiv\_estimation(mu,nu,tau,metric='cosine'):}
        \INDB{1}
            \PyCode{mu\_mu\_matrix = dist\_matrix(mu,mu,metric)}
            \PyCode{mu\_nu\_matrix = dist\_matrix(mu,nu,metric)}
            \PyCode{mu\_mu\_log\_density = kernel(mu\_mu\_matrix,tau,metric).mean(-1).log()}
            \PyCode{mu\_nu\_log\_density = kernel(mu\_nu\_matrix,tau,metric).mean(-1).log()}
            \PyCode{return (mu\_mu\_log\_density-mu\_nu\_log\_density).mean()}
        \INDE
    \end{algorithmic}
    \caption{PyTorch-style pseudocode of KL divergence estimation.}
    \label{apdx-kldiv-algorithm}
\end{algorithm}

\newpage

\section{Validation with ImageNet}
\label{apdx-imagenet-res}

\subsection{ImageNet dataset}
\label{apdx-imagenet-dataset}

For the ImageNet dataset, we evaluate several self-supervised learning models using various metrics. We use MoCo-v2 \cite{chen2020improved}, MoCo-v3 \cite{chen2021empirical}, SimCLR-v1 \cite{chen2020simple}, and SupCon \cite{khosla2020supervised} by utilizing L2-normalized projection head features. However, for DINO (v1 \cite{caron2021emerging}, v2 \cite{oquab2023dinov2}), we use L2-normalized backbone features rather than projection features because the origin papers employed normalized backbone features to measure KNN accuracy. We utilize pre-trained weights which are trained with ImageNet dataset for both the backbone and projection head of each model. The reported Top-1 linear probing accuracy is sourced directly from the respective papers or GitHub repos.

For augmentation set used for evaluation, we apply the most basic augmentations: (1) resizing to 256 $\times$ 256, (2) center cropping to 224 $\times$ 224, and (3) normalizing with mean and standard deviation of ImageNet dataset. We omit normalization for experiments with SimCLR since the original implementation did not use normalization.

\subsection{EDS values and downstream task accuracy}

\subsubsection{Table of EDS values and task accuracies}
\label{apdx-eds-table}

\begin{table}[ht]
\centering
\renewcommand{\arraystretch}{1.25}
    \caption{EDS values of pretrained metric learning and SSL models with ImageNet classification accuracies. To verify the reported performance of the pretrained models, the linear probing accuracy is additionally presented. (*: normalized backbone features are utilized instead of embedding vector.)}
    \vskip -0.5em
    \begin{adjustbox}{max width=\textwidth}
        \begin{tabular}{c|cc||cc|cc|cc||c|ccc|ccc}
        \Xhline{3\arrayrulewidth}
        \multirow{2}{*}{Model}&\multicolumn{2}{c||}{Architecture}&\multicolumn{2}{c|}{EDS ($\tau=1.0$)}&\multicolumn{2}{c|}{EDS ($\tau=0.5$)}&\multicolumn{2}{c||}{EDS ($\tau=0.1$)}&Linear&\multicolumn{3}{c|}{Mean}&\multicolumn{3}{c}{KNN}\\
        \cline{2-16}
        &backbone&dim&del&eps&del&eps&del&eps&1&1&3&5&3&5&7\\
        \hline
        \hline
        MoCo-v2&ResNet-50&128&0.450&0.390&0.219&0.154&0.025&0.001&71.1&46.68&67.62&74.92&47.03&48.90&49.55\\
        \hline
        \multirow{2}{*}{MoCo-v3}&ViT-S&256&0.442&0.391&0.208&0.155&0.023&0.000&73.2&58.35&74.31&79.07&57.00&58.47&59.05\\
        &ViT-B&256&0.462&0.389&0.230&0.153&0.026&0.000&76.7&62.15&78.79&83.27&60.90&62.35&62.77\\
        \hline
        SimCLR&ResNet-50&128&0.427&0.401&0.194&0.163&0.021&0.000&67.8&46.55&65.76&72.69&43.63&45.73&46.86\\
        \hline
        \multirow{2}{*}{DINO-v1 (*)}&ViT-S&384&0.502&0.408&0.264&0.168&0.024&0.000&79.7&71.13&86.43&90.36&73.08&74.05&74.41\\
        &ViT-B&768&0.500&0.410&0.260&0.169&0.023&0.000&80.1&70.15&86.38&90.56&71.62&72.66&72.87\\
        \hline
        \multirow{2}{*}{DINO-v2 (*)}&ViT-S&384&0.473&0.391&0.235&0.154&0.022&0.000&81.1&71.23&87.33&91.44&73.86&74.85&75.13\\
        &ViT-B&768&0.475&0.384&0.238&0.148&0.023&0.000&\bf84.5&76.63&\bf91.08&\bf94.16&\bf78.15&\bf78.95&\bf79.20\\
        \hline
        SupCon&ResNet-50&128&\bf0.681&0.478&\bf0.477&0.232&\bf0.074&0.002&74.1&\bf79.08&\bf90.69&93.18&\bf78.19&\bf78.65&\bf78.81\\
        \Xhline{3\arrayrulewidth}
        \end{tabular}
    \end{adjustbox}
    \vskip -0.5em
    \label{res-imagenet}
\end{table}

\subsubsection{EDS values of pretrained models}
\begin{figure}[ht]
\centering
    \begin{subfigure}{0.29\textwidth}
        \includegraphics[width=\textwidth]{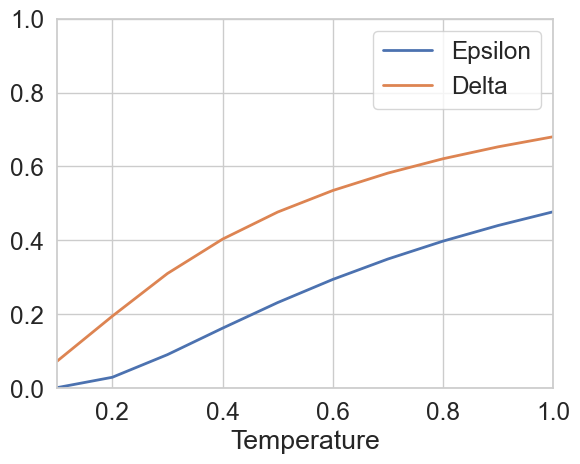}
        \caption{SupCon}
    \end{subfigure}%
    \hspace{0.5em}
    \begin{subfigure}{0.29\textwidth}
        \includegraphics[width=\textwidth]{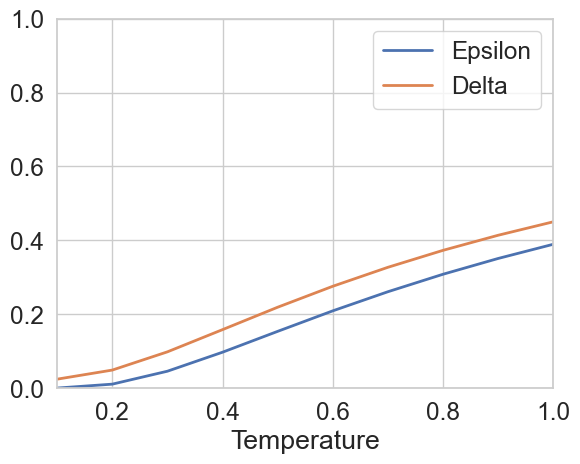}
        \caption{MoCo-v2}
    \end{subfigure}%
    \hspace{0.5em}
    \begin{subfigure}{0.29\textwidth}
        \includegraphics[width=\textwidth]{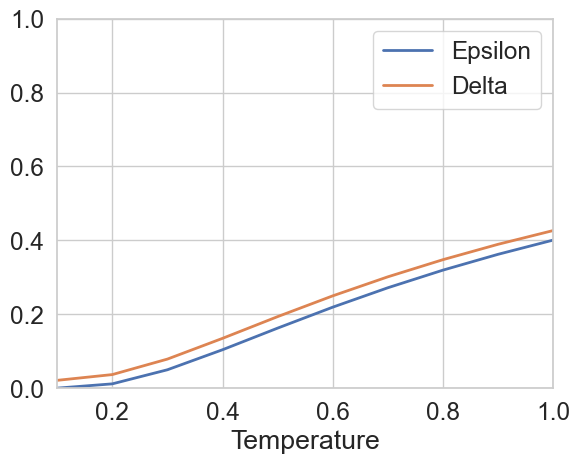}
        \caption{SimCLR}
    \end{subfigure}
     \begin{subfigure}{0.29\textwidth}
        \includegraphics[width=\textwidth]{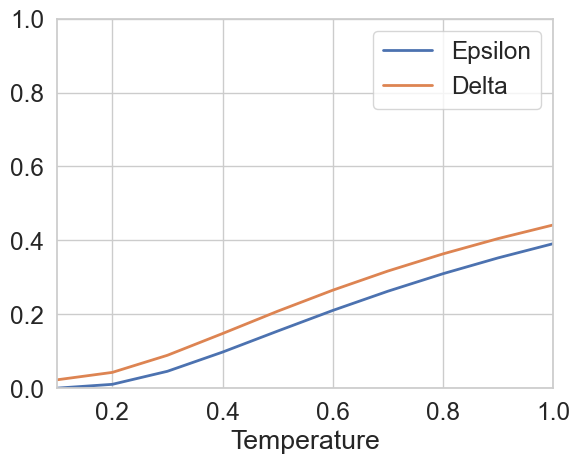}
        \caption{MoCo-v3 (ViT-S)}
    \end{subfigure}%
    \hspace{0.5em}
    \begin{subfigure}{0.29\textwidth}
        \includegraphics[width=\textwidth]{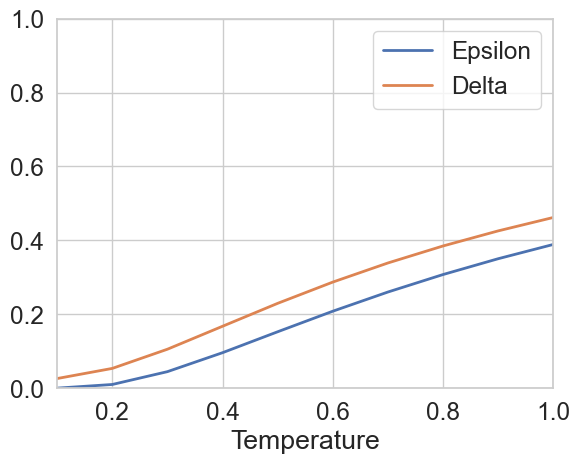}
        \caption{MoCo-v3 (ViT-B)}
    \end{subfigure}%
    \hspace{0.5em}
    \begin{subfigure}{0.29\textwidth}
        \includegraphics[width=\textwidth]{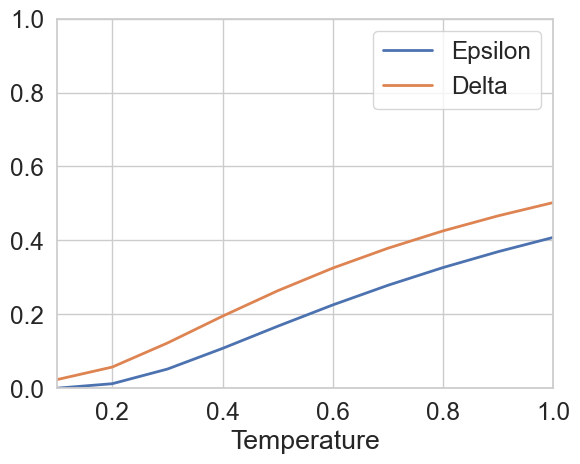}
        \caption{DINO-v1 (ViT-S)}
    \end{subfigure}
     \begin{subfigure}{0.29\textwidth}
        \includegraphics[width=\textwidth]{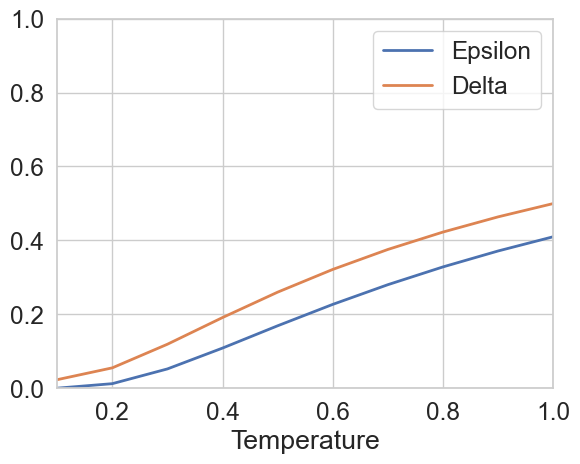}
        \caption{DINO-v1 (ViT-B)}
    \end{subfigure}%
    \hspace{0.5em}
    \begin{subfigure}{0.29\textwidth}
        \includegraphics[width=\textwidth]{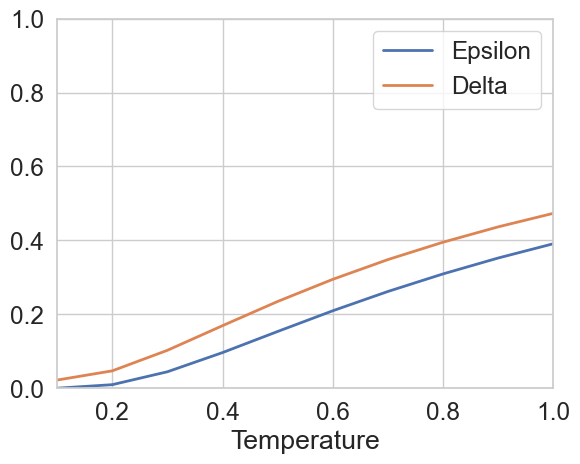}
        \caption{DINO-v2 (ViT-S)}
    \end{subfigure}%
    \hspace{0.5em}
    \begin{subfigure}{0.29\textwidth}
        \includegraphics[width=\textwidth]{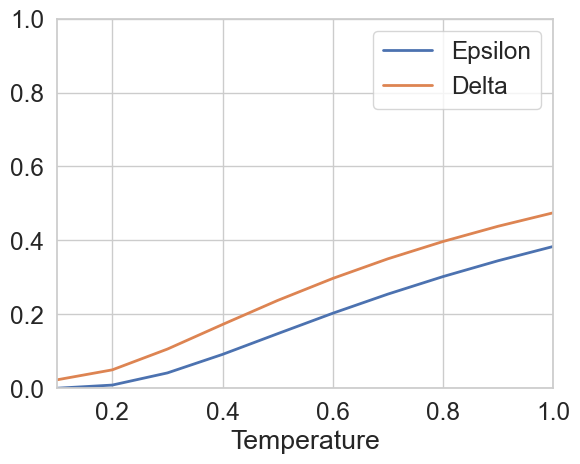}
        \caption{DINO-v2 (ViT-B)}
    \end{subfigure}
    \caption{Visualization of EDS values with different temperatures with ImageNet Dataset.}
    \label{apdx-res-imagenet-eds}
\end{figure}

We first compute the $\epsilon$ and $\delta$ values of $f$ for each trained model. To reduce the influence of outliers, we removed approximately 5\% of the data before aggregating the kernel density. Additionally, we report the mean values of $\epsilon$ and $\delta$ across all classes, rather than the minimum or maximum values, as the size of each class cluster can vary. \cref{apdx-res-imagenet-eds} visualizes the EDS values for each model with different temperature settings of $\tau$.

\subsection{object occurrence estimation}

\begin{figure}[ht]
\centering
    \begin{subfigure}{0.28\textwidth}
        \includegraphics[width=\textwidth]{imgs/imagenet_existence/mean_uniform.png}
        \caption{Uniform distribution}
    \end{subfigure}
    \hskip 1em
    \begin{subfigure}{0.28\textwidth}
        \includegraphics[width=\textwidth]{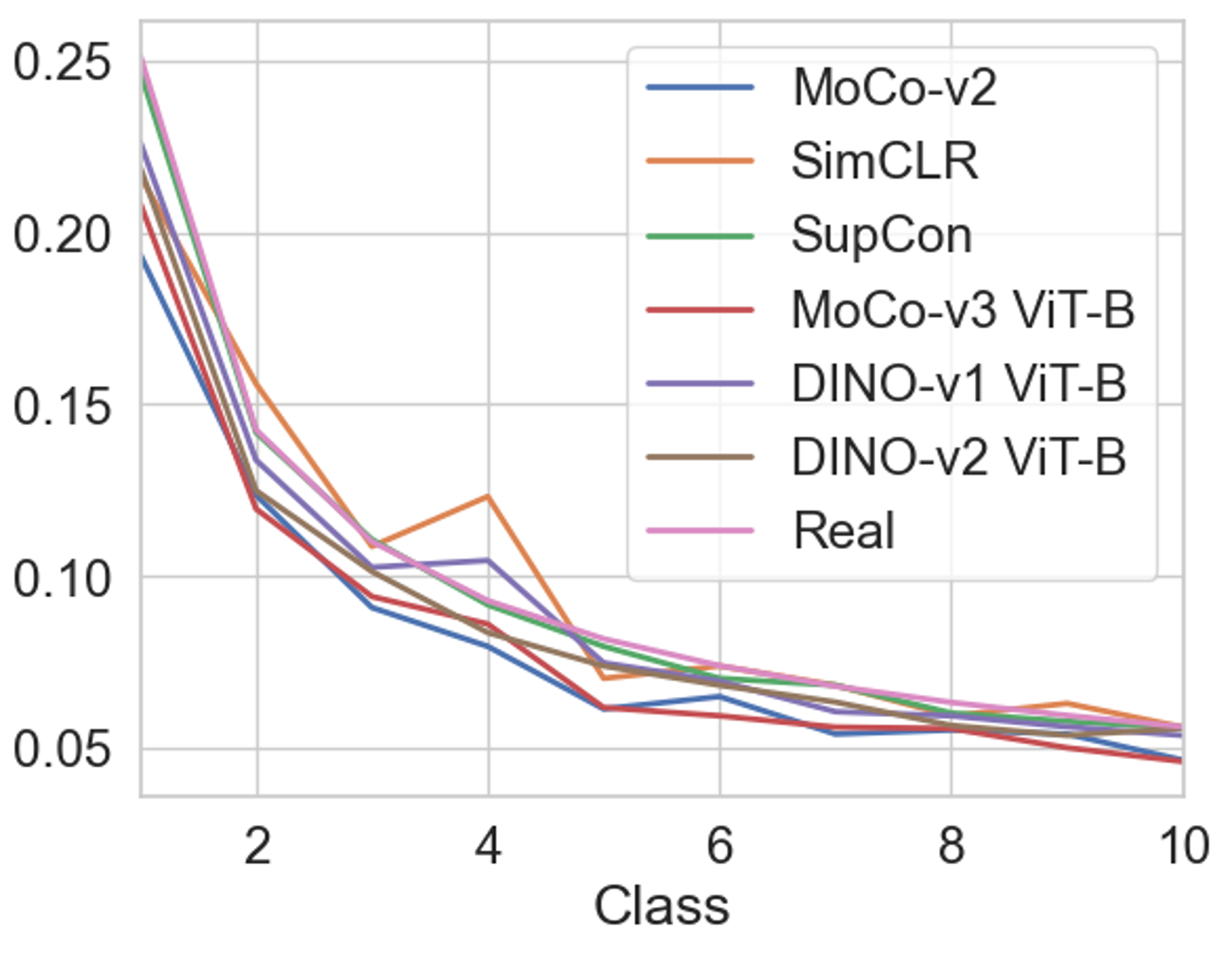}
        \caption{A Zipf distribution ($\alpha=0.5$)}
    \end{subfigure}%
    \hskip 1em
    \begin{subfigure}{0.28\textwidth}
        \includegraphics[width=\textwidth]{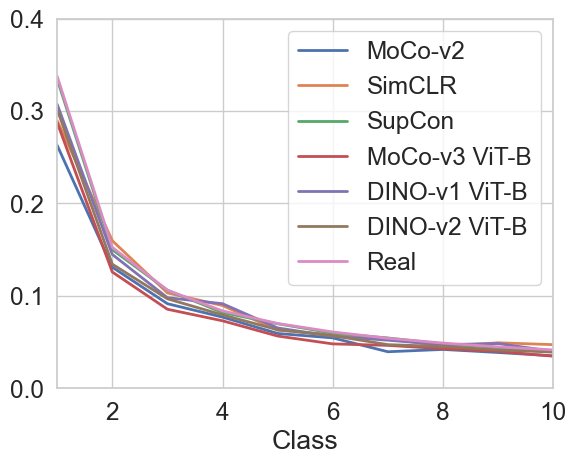}
        \caption{A Zipf distribution ($\alpha=0.7$)}
    \end{subfigure}
    \caption{Visualization of object existence probability estimation with various scenarios. Mean classifier used to enhance the estimation accuracy with temperature $\tau=0.2$ (Mean values are reported with 5 different seeds.)}
    \label{apdx-existence-scenario}
\end{figure}

We conduct additional experiments with different scenarios. \cref{apdx-existence-scenario} shows the results with (a) a uniform distribution and (b,c) Zipf distributions with $\alpha=(0.5,0.7)$, respectively. In every cases, our method successfully estimates the original distribution.

\newpage

\subsection{KL divergence estimation}
\label{apdx-imagenet-kldiv-res}

For the KL divergence estimation experiment, we take the following steps to obtain the artificial data distribution. First, we randomly select a certain number of classes and divide them into two groups. Then, we sample the data according to the proportion of each group. At this point, the optimal KL divergence is the KL divergence value corresponding to the group ratios. The KL divergence estimate is then be computed using \cref{apdx-kldiv-algorithm}. \cref{apdx-kld-scenario} describes the three scenarios in which the experiment was conducted.

\begin{table}[ht]
\centering
\renewcommand{\arraystretch}{1.1}
    \caption{Descriptions of scenarios in KL divergence experiments.}
    \begin{adjustbox}{max width=0.9\textwidth}
        \begin{tabular}{c|ccccc}
        \Xhline{3\arrayrulewidth}
        &Number of classes&$\mu$&$\nu$&Number of images&Optimal KL div.\\
        \hline
        Scenario 1 &10 (5/5)&$[0.4,0.6]$&$[0.6,0.4]$&1000&0.0811\\
        Scenario 2 &10 (5/5)&$[0.2,0.8]$&$[0.8,0.2]$&1000&0.8318\\
        Scenario 3 &40 (20/20)&$[0.2,0.8]$&$[0.8,0.2]$&4000&0.8318\\
        \Xhline{3\arrayrulewidth}
        \end{tabular}
    \end{adjustbox}
    \label{apdx-kld-scenario}
\end{table}

\begin{figure}[ht]
\centering
    \begin{subfigure}{0.28\textwidth}
        \includegraphics[width=\textwidth]{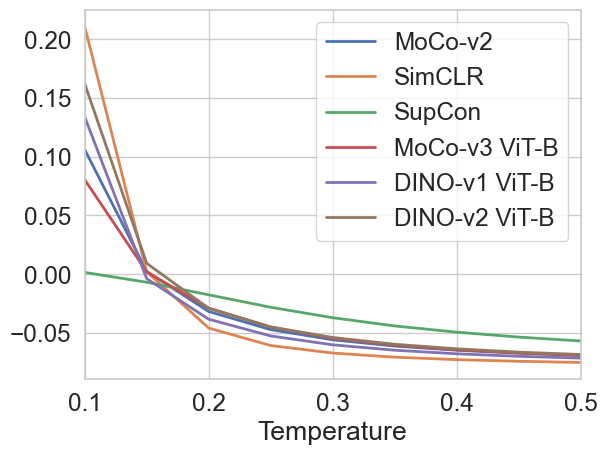}
        \caption{KLD difference (exact)}
    \end{subfigure}%
    \hskip 1em
    \begin{subfigure}{0.28\textwidth}
        \includegraphics[width=\textwidth]{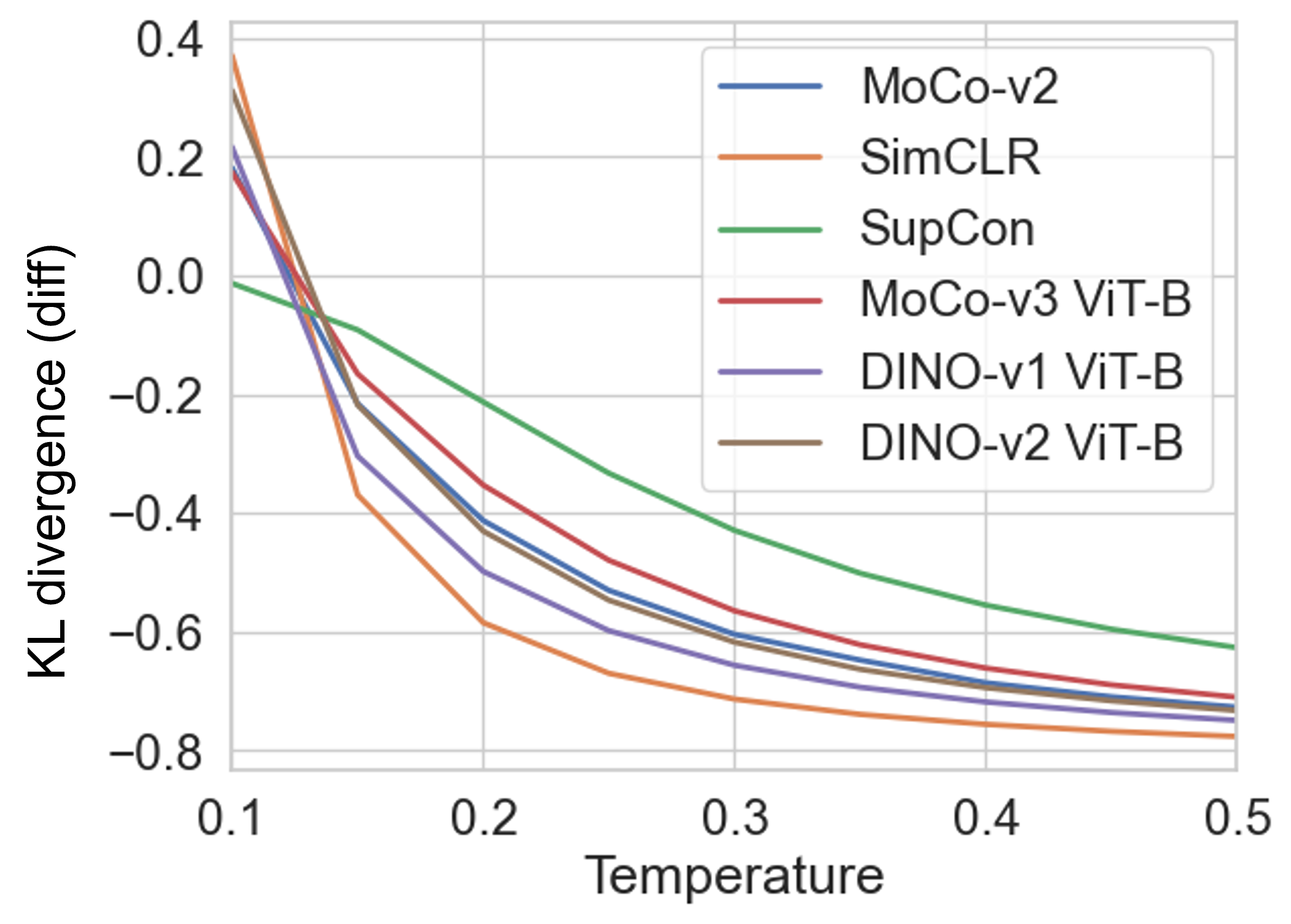}
        \caption{KLD difference (exact)}
    \end{subfigure}%
    \hskip 1em
    \begin{subfigure}{0.28\textwidth}
        \includegraphics[width=\textwidth]{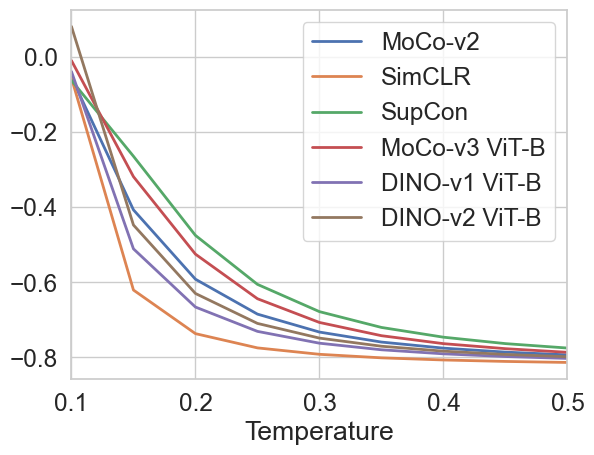}
        \caption{KLD difference (exact)}
    \end{subfigure}
    \begin{subfigure}{0.28\textwidth}
        \includegraphics[width=\textwidth]{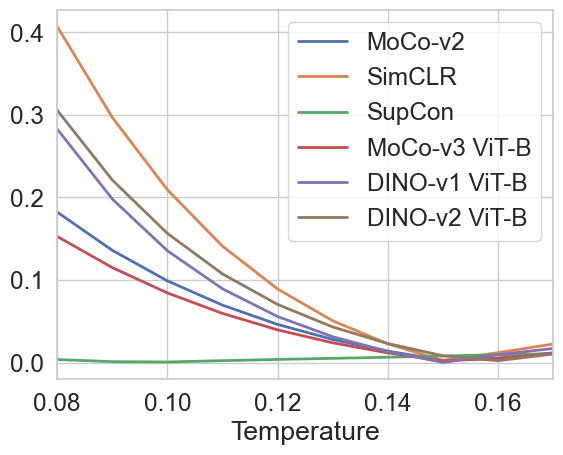}
        \caption{KLD difference (simplex-ETF)}
    \end{subfigure}%
    \hskip 1em
    \begin{subfigure}{0.28\textwidth}
        \includegraphics[width=\textwidth]{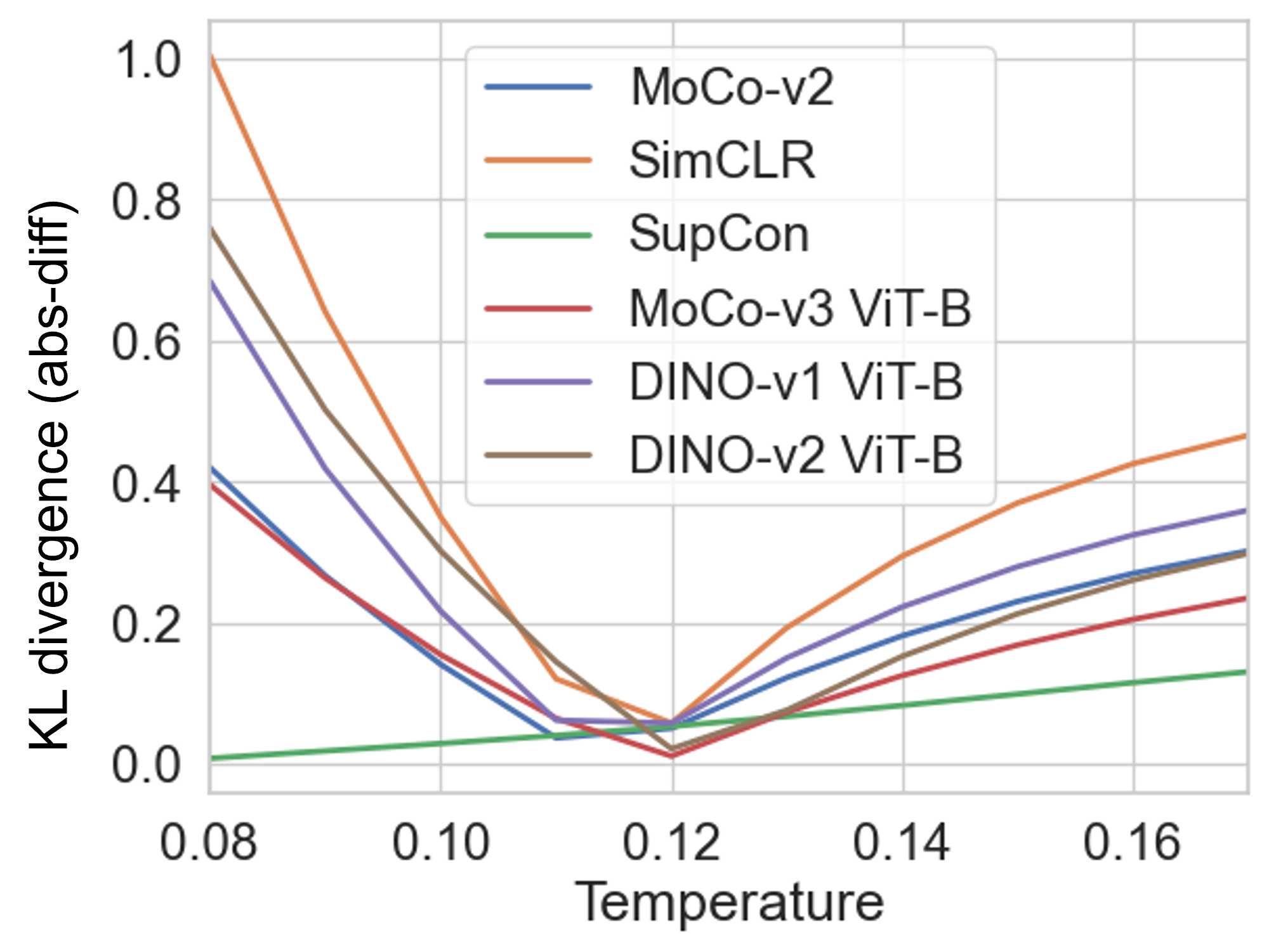}
        \caption{KLD difference (simplex-ETF)}
    \end{subfigure}%
    \hskip 1em
    \begin{subfigure}{0.28\textwidth}
        \includegraphics[width=\textwidth]{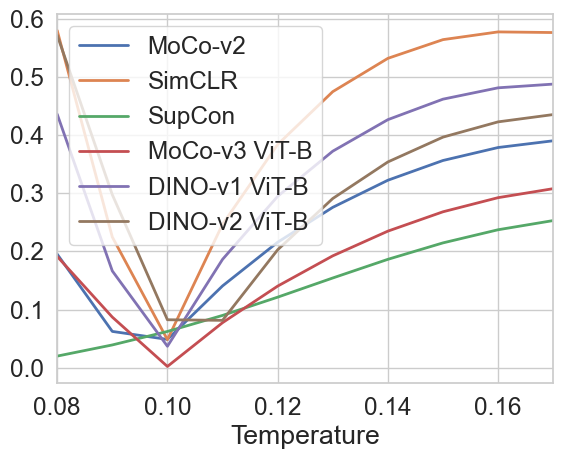}
        \caption{KLD difference (simplex-ETF)}
    \end{subfigure}
    
    \caption{Visualization of KL divergence estimation with three different scenarios. (a)-(d) correspond to Scenario 1, (b)-(e) to Scenario 2, and (c)-(f) to Scenario 3. The scenario setting is shown in Table \cref{apdx-kld-scenario}. (Mean values are reported with 5 different seeds.)}
    \label{apdx-res-imagenet-kld}
    \vskip -0.5em
\end{figure}

\newpage

\section{Applying OBSER framework to open-world environment (Minecraft)}
\label{apdx-minecraft}

\subsection{Minecraft dataset}
\label{apdx-minecraft-dataset}

We construct a Minecraft dataset containing 26,000 images and the corresponding labels. We gather ego-centric observations of objects to train or fine-tune models. To build the dataset, we first choose typical biomes that can represent all objects in the \textit{overworld} in Minecraft. The dataset is derived from two environments: an open-world environment and a miniature environment.

\subsubsection{Open-world multi-angled data collection}
\label{minecraft-dataset-openworld}
\begin{figure}[t]
    \centering
    \includegraphics[width=0.8\textwidth]{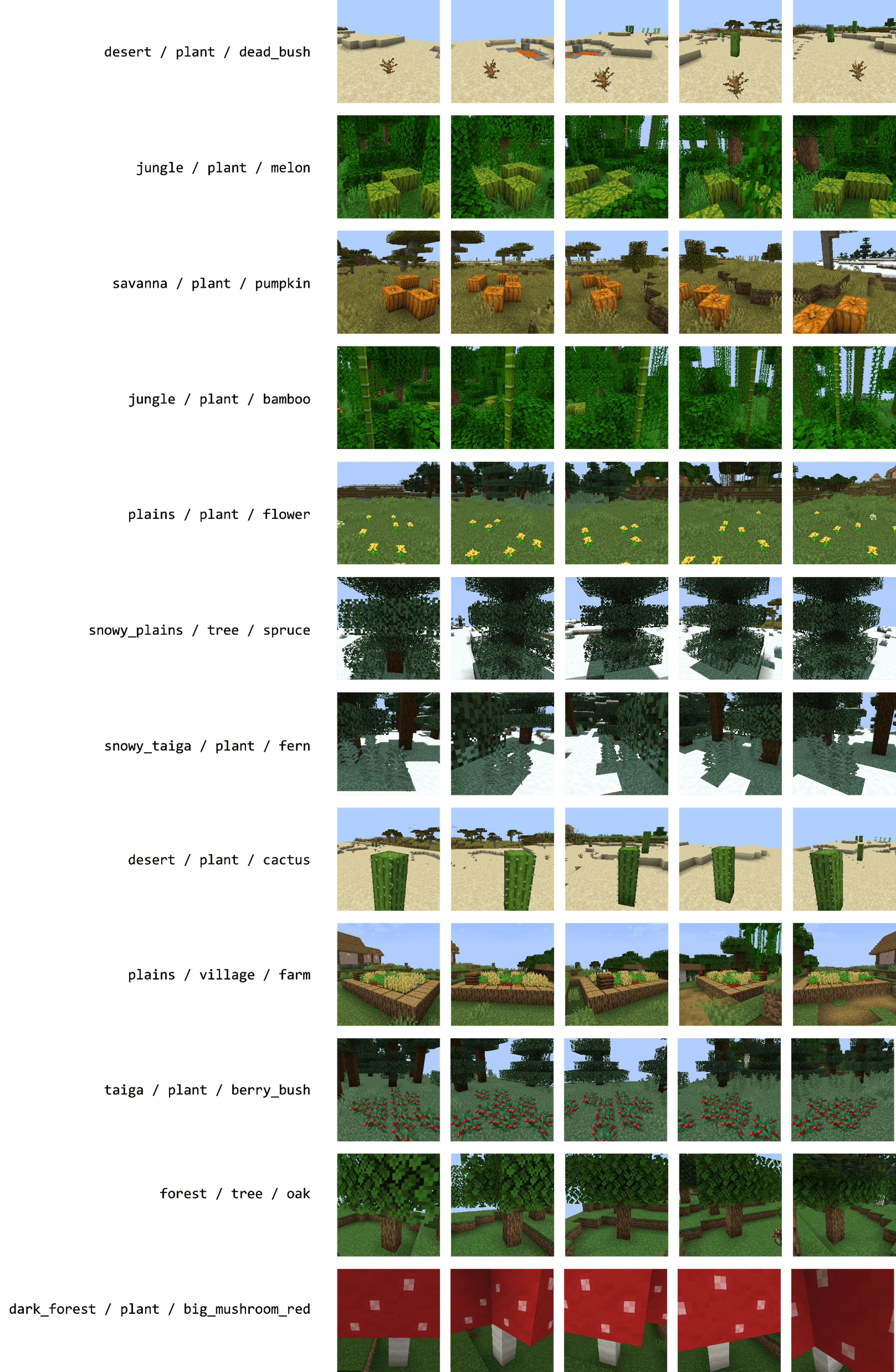}
    \caption{Example observations with various objects from different biomes.}
    \label{fig-apdx-asdf}
\end{figure}

For the open-world multi-angled data collection, we use the Minecraft's default world generation settings to generate a world environment where the agent collects the dataset. In such an environment, the agents can easily get stuck due to composite terrains and the arrangement of objects. This reduces the efficiency of data acquisition. Therefore, we manually locate 100 objects per biome in the world using a fixed seed, rather than relying on fully automatic methods such as the random walk algorithm. For each object, we gather 30 observations by rotating around the object's position. Figure \cref{fig-apdx-asdf} shows examples of the observations collected. We split 10\% of each observation to build the test set. Note that we do not split the dataset by object because object frequencies follow a long-tailed distribution, and object-based splitting could introduce further distortions in the data distribution. We choose two levels of hierarchical concepts to conduct experiments with different levels of abstraction. Table \cref{apdx-table-biome-dist} shows the proportion of hierarchical concepts of the proposed dataset.

\subsubsection{Miniature environment multi-angled data collection}
\label{minecraft-dataset-miniature}
We also collect data from our miniature environment, which consists of 4 $\times$ 5 grids of 48 $\times$ 48 blocks with the same biome. The kind of 10 biomes in the dataset is the same as the ones in the open-world multi-angled data collection and in the table \cref{apdx-table-biome-dist}. To collect the data in the miniature environment, we used the same algorithm as used in open-world environment.

\begin{figure}[ht]
    \includegraphics[width=0.9\textwidth]{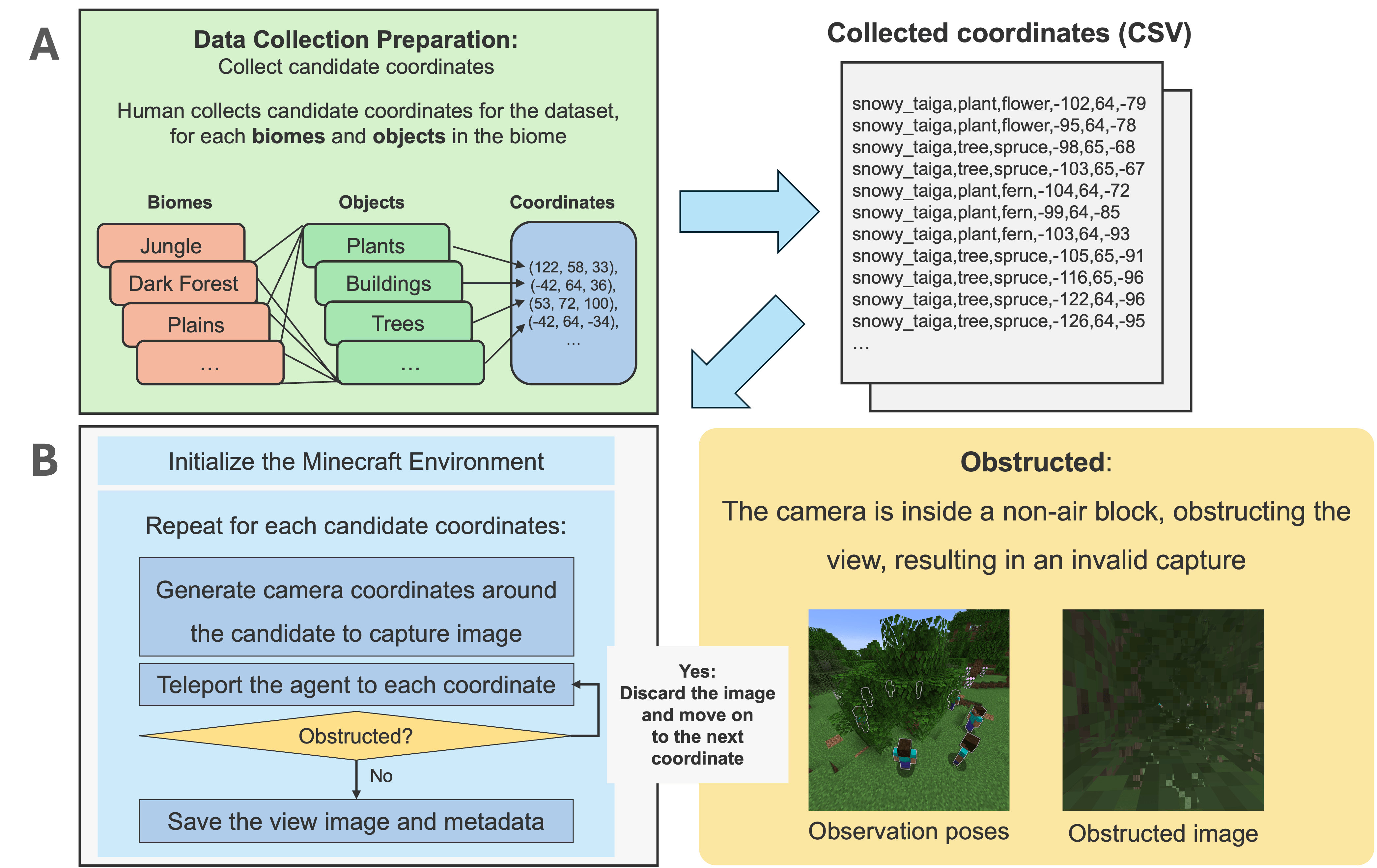}
    \caption{Flowchart illustrating the Minecraft data collection process.
A) A human manually collects candidate coordinates for the dataset, with separate CSV files generated for each biome. With 10 different biomes, we gather coordinates of each biome with 10 distinct CSV files. Each CSV file contains entries for the biome name, object type, object name, and the corresponding x, y, z coordinates for each object within that biome.
B) Using a customized Minecraft automation setup, the agent collects the data with each CSV files. The agent is teleported to specified coordinates and performs a 360-degree rotation around the point of interest, capturing images at 30 intervals of 12 degrees each. If the view of the camera is obstructed by being inside a non-air block, the corresponding image is automatically discarded. This process is repeated across all biomes. Successfully captured images are saved with the associated coordinates in the filename.}
    \label{fig-apdx-obs-vis}
\end{figure}

\begin{table}[t]
    \centering
    \renewcommand{\arraystretch}{1.1}
    \caption{Summary of statistical information for the Minecraft dataset. We focus on gathering unique objects from each biome, and it draws some difference between gathered data and real distribution, especially with villages. For evaluation, we use observations from the Miniature environment instead of gathered dataset.}
    \begin{subtable}[t]{0.44\textwidth}
        \begin{adjustbox}{width=\textwidth}
            \begin{tabular}{c|c|c|c}
                \Xhline{3\arrayrulewidth}
                Biome&Category&Sub-category&Frequency\\
                \hline
                \hline
                \multirow{3}{*}{forest}&\multirow{1}{*}{plant}&flower&0.09\\
                \cline{2-4}
                &\multirow{2}{*}{tree}&oak&0.52\\
                &&birch&0.39\\
                \Xhline{2\arrayrulewidth}   
                \multirowcell{7}{dark\\forest}&\multirow{4}{*}{plant}&flower&0.02\\
                &&big mushroom (brown)&0.12\\
                &&big mushroom (red)&0.17\\
                &&pumpkin&0.01\\
                \cline{2-4}
                &\multirow{3}{*}{tree}&oak&0.16\\
                &&birch&0.12\\
                &&dark oak&0.40\\
                \Xhline{2\arrayrulewidth}   
                \multirow{7}{*}{desert}&\multirow{3}{*}{plant}&cactus&0.35\\
                &&sugarcane&0.19\\
                &&dead bush&0.20\\
                \cline{2-4}
                &tree&azalea&0.05\\
                \cline{2-4}
                &\multirow{3}{*}{village}&building&0.08\\
                &&decorative&0.08\\
                &&farm&0.05\\
                \Xhline{2\arrayrulewidth}   
                \multirow{9}{*}{savanna}&\multirow{4}{*}{plant}&flower&0.07\\
                &&grass&0.20\\
                &&pumpkin&0.05\\
                &&melon&0.05\\
                \cline{2-4}
                &\multirow{2}{*}{tree}&oak&0.09\\
                &&acacia&0.27\\
                \cline{2-4}
                &\multirow{3}{*}{village}&building&0.20\\
                &&decorative&0.06\\
                &&farm&0.05\\
                \Xhline{2\arrayrulewidth}   
                \multirow{10}{*}{swamp}&\multirow{8}{*}{plant}&flower&0.09\\
                &&sugarcane&0.14\\
                &&lily pad&0.11\\
                &&grass&0.06\\
                &&small mushroom (brown)&0.16\\
                &&small mushroom (red)&0.01\\
                &&pumpkin&0.02\\
                &&dead bush&0.08\\
                \cline{2-4}
                &\multirow{1}{*}{tree}&oak&0.31\\
                \cline{2-4}
                &\multirow{1}{*}{structure}&building&0.02\\
                \Xhline{3\arrayrulewidth}
            \end{tabular}
        \end{adjustbox}
    \end{subtable}%
    \hskip 0.5em
    \begin{subtable}[!t]{0.45\textwidth}
        \begin{adjustbox}{width=\textwidth}
            \begin{tabular}{c|c|c|c}
                \Xhline{3\arrayrulewidth}
                Biome&Category&Sub-category&Frequency\\
                \hline
                \hline
                \multirow{7}{*}{plains}&\multirow{3}{*}{plant}&flower&0.26\\
                &&grass&0.33\\
                &&pumpkin&0.03\\
                \cline{2-4}
                &tree&oak&0.21\\
                \cline{2-4}
                &\multirow{3}{*}{village}&building&0.12\\
                &&decorative&0.02\\
                &&farm&0.03\\
                \Xhline{2\arrayrulewidth}   
                \multirowcell{7}{snowy\\plains}&\multirow{3}{*}{plant}&flower&0.16\\
                &&grass&0.28\\
                &&pumpkin&0.01\\
                \cline{2-4}
                &\multirow{1}{*}{tree}&spruce&0.25\\
                \cline{2-4}
                &\multirow{3}{*}{village}&building&0.21\\
                &&decorative&0.07\\
                &&farm&0.02\\
                \Xhline{2\arrayrulewidth}   
                \multirow{9}{*}{taiga}&\multirow{5}{*}{plant}&flower&0.04\\
                &&fern&0.12\\
                &&berry bush&0.12\\
                &&pumpkin&0.03\\
                &&small mushroom (brown)&0.04\\
                \cline{2-4}
                &\multirow{1}{*}{tree}&spruce&0.42\\
                \cline{2-4}
                &\multirow{3}{*}{village}&building&0.16\\
                &&decorative&0.04\\
                &&farm&0.03\\
                \Xhline{2\arrayrulewidth}   
                \multirowcell{6}{snowy\\taiga}&\multirow{5}{*}{plant}&flower&0.05\\
                &&fern&0.23\\
                &&berry bush&0.02\\
                &&pumpkin&0.01\\
                &&small mushroom (brown)&0.05\\
                \cline{2-4}
                &\multirow{1}{*}{tree}&spruce&0.64\\
                \Xhline{2\arrayrulewidth}   
                \multirow{5}{*}{jungle}&\multirow{3}{*}{plant}&flower&0.02\\
                &&bamboo&0.21\\
                &&melon&0.11\\
                \cline{2-4}
                &\multirow{2}{*}{tree}&oak&0.11\\
                &&jungle&0.55\\
                \Xhline{3\arrayrulewidth}
            \end{tabular}
        \end{adjustbox}
    \end{subtable}
    \label{apdx-table-biome-dist}
\end{table}

\subsubsection{Miniature scenario collection}
In the miniature scenario, the agent performs random exploration within the grid to collect visual observations programmatically. For each grid, we teleport the agent to its center. The agent wanders randomly through the grid to collect visual observations and save them as images. The agent is prevented from going outside the grid by checking its distance from the grid's center. If the agent goes farther than 20 blocks from the center of the grid, we force the agent to look at the center of the grid using the teleport command. To reduce the likelihood of the agent’s view capturing parts of neighboring grids with different biomes, we have the agent look slightly downward following a normal distribution, $\mathcal{N}(30, 5^2)$. In this setting, we also use Pareto distribution to determine the distance it moves forward before turning around by degrees randomly following $\mathcal{N}(30, 1^2)$.

\begin{figure}[t]
    \centering
    \includegraphics[width=0.8\textwidth]{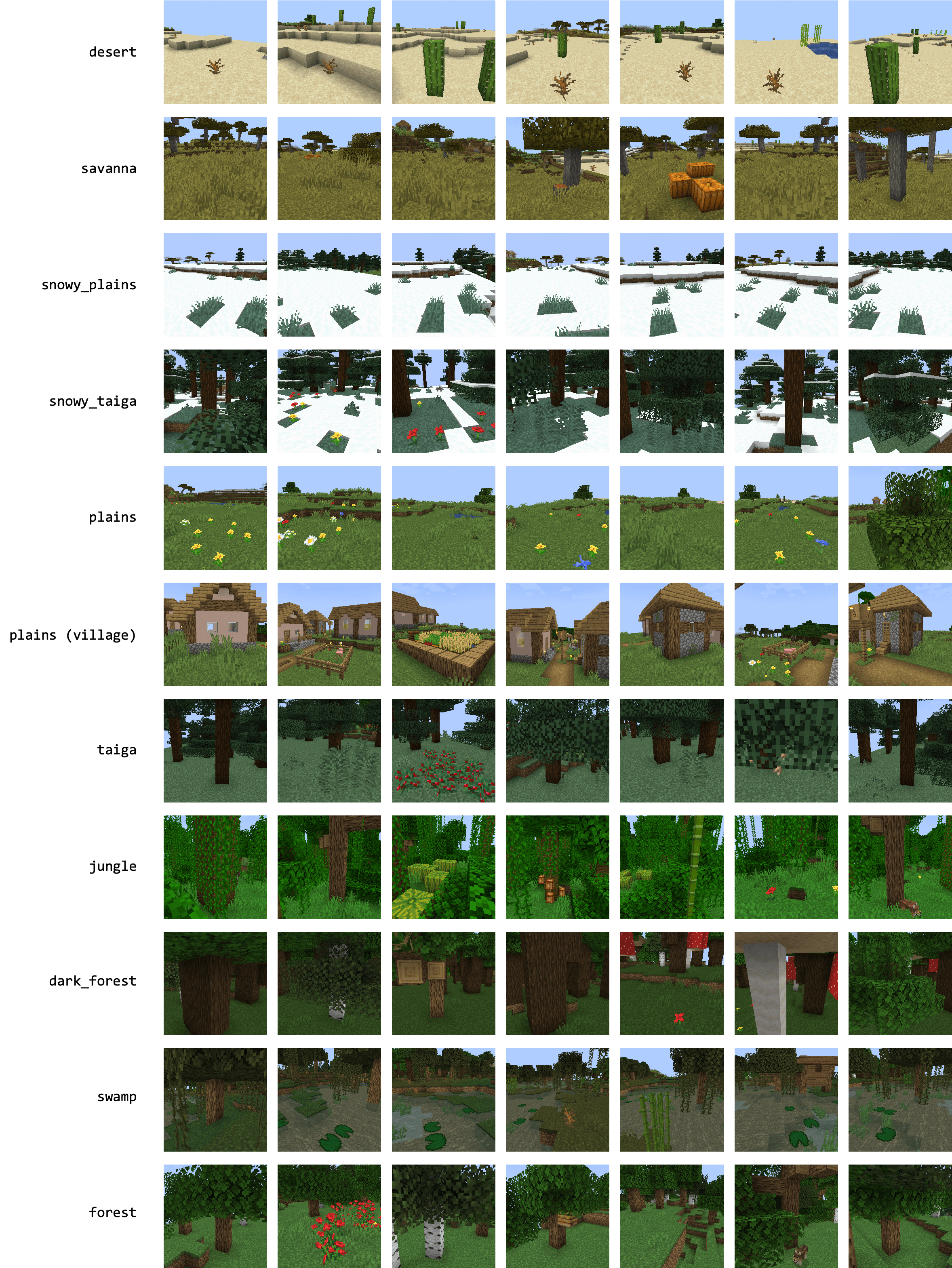}
    \caption{Example observations from different biomes.}
    \label{fig-apdx-biomes}
\end{figure}

\clearpage

\subsection{Minecraft experiment}
\label{apdx-minecraft-experiment}

For experiments in Minecraft environment, we choose SimCLR, MoCo, and SupCon with ResNet-50 as a backbone. In this experiment, we utilize the projection layer of each model in the same way as it was applied to the ImageNet data in the original works. We have used 1 NVIDIA RTX 3090 Ti to train each model. Table \cref{apdx-minecraft-hyperparameter} shows the hyperparameters and details used to train each model. We add \texttt{RandomResizedCrop} to the augmentation set to reflect the various distances between the agent and the object. PyTorch-style pseudocode of the augmentation set is shown in Algorithm \cref{apdx-minecraft-augmentation}.

\begin{table}[ht]
\centering
\renewcommand{\arraystretch}{1.2}
\caption{Details of the hyperparameters and settings used to train each model.}
\begin{adjustbox}{max width=\textwidth}
\begin{tabular}{c|ccccccccc}
\Xhline{3\arrayrulewidth}
Model&Backbone&Emb. Dim&Emb. Type&Batch size & Epochs (Warmup / Train) & Optimizer & Learning rate  & Scheduler & Temperature\\
\hline
\hline
SupCon &ResNet-50&128&Projection&128&10 / 20&LARS&0.15&Cosine&0.2\\
\hline
SimCLR &ResNet-50&128&Projection&128&10 / 20&LARS&0.15&Cosine&0.2\\
\hline
MoCo-v2&ResNet-50&128&Projection&128&10 / 100&SGD&0.03&Cosine&0.2\\
\Xhline{3\arrayrulewidth}
\end{tabular}
\end{adjustbox}
\label{apdx-minecraft-hyperparameter}
\end{table}

\begin{algorithm}[ht]
    \small
    \begin{algorithmic}
        \PyCode{transform = Compose([}
        \INDB{1}
            \PyCode{Resize(256),}
            \PyCode{RandomResizedCrop(size=224,scale=[0.5,1.0]),}
            \PyCode{RandomApply(}
            \INDB{2}
                \PyCode{[ColorJitter(0.2,0.2,0.2,0.1)], p=0.8}
            \INDE
            \PyCode{),}
            \PyCode{RandomHorizontalFlip(),}
            \PyCode{ToTensor(),}
            \PyCode{Normalize(}
            \INDB{2}
                \PyCode{mean=[0.3232, 0.3674, 0.2973],}
                \PyCode{std=[0.2615, 0.2647, 0.3390]}
            \INDE
            \PyCode{),}
        \INDE
        \PyCode{])}
    \end{algorithmic}
    \caption{PyTorch-style code of the augmentation set for Minecraft experiments.}
    \label{apdx-minecraft-augmentation}
\end{algorithm}

\subsubsection{EDS values of trained models}
\begin{figure}[ht]
\centering
    \begin{subfigure}{0.29\textwidth}
        \includegraphics[width=\textwidth]{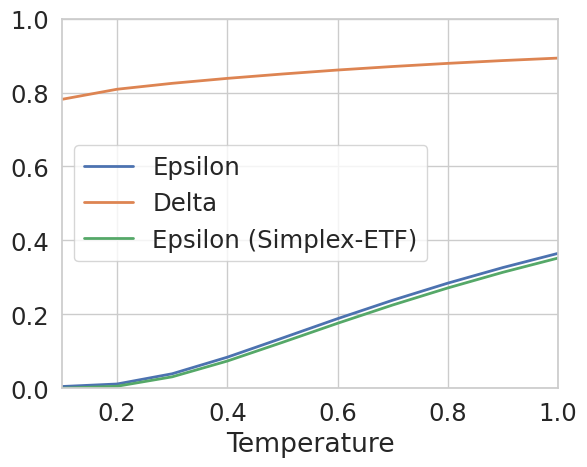}
        \caption{EDS (SupCon)}
    \end{subfigure}%
    \hspace{0.5em}
    \begin{subfigure}{0.29\textwidth}
        \includegraphics[width=\textwidth]{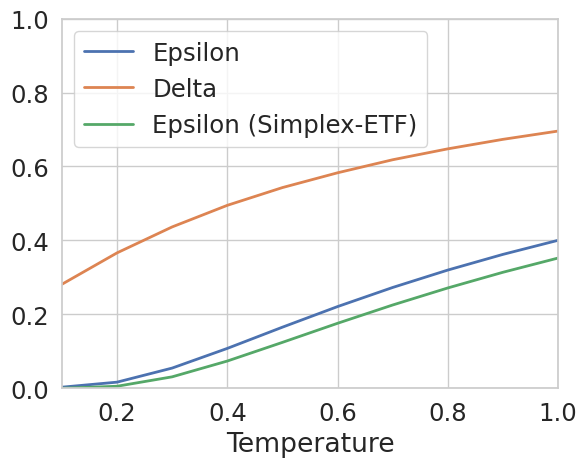}
        \caption{EDS (MoCo)}
    \end{subfigure}%
    \hspace{0.5em}
    \begin{subfigure}{0.29\textwidth}
        \includegraphics[width=\textwidth]{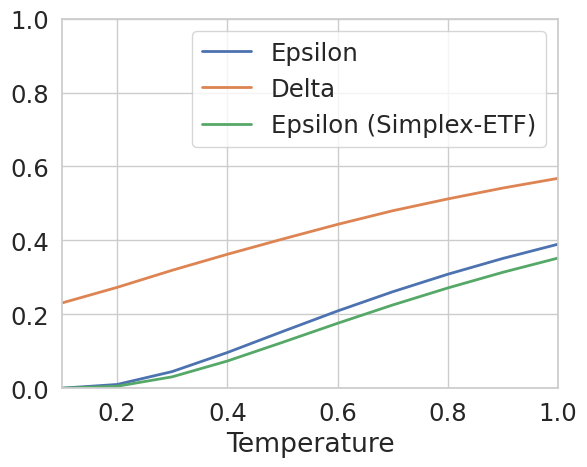}
        \caption{EDS (SimCLR)}
    \end{subfigure}%
    \caption{Visualization of EDS values with different temperatures with Minecraft Dataset.}
    \label{res-minecraft-eds}
\end{figure}

As the same method in \cref{apdx-imagenet-res}, we also visualize the $\epsilon$ and $\delta$ values of each trained model in \cref{res-minecraft-eds}. Due to the simpler task than ImageNet, each model has higher $\delta$ and lower $\epsilon$ than those in ImageNet.

\subsubsection{Object-environment retrieval task}
\begin{figure}[ht]
\centering
    \begin{subfigure}{0.9\textwidth}
        \includegraphics[width=\textwidth]{imgs/res-minecraft-obj-retrieval.png}
        \vskip -0.5em
        \caption{\texttt{Flower} in \texttt{Forest}}
    \end{subfigure}
    \vskip 0.5em
    \begin{subfigure}{0.9\textwidth}
        \includegraphics[width=\textwidth]{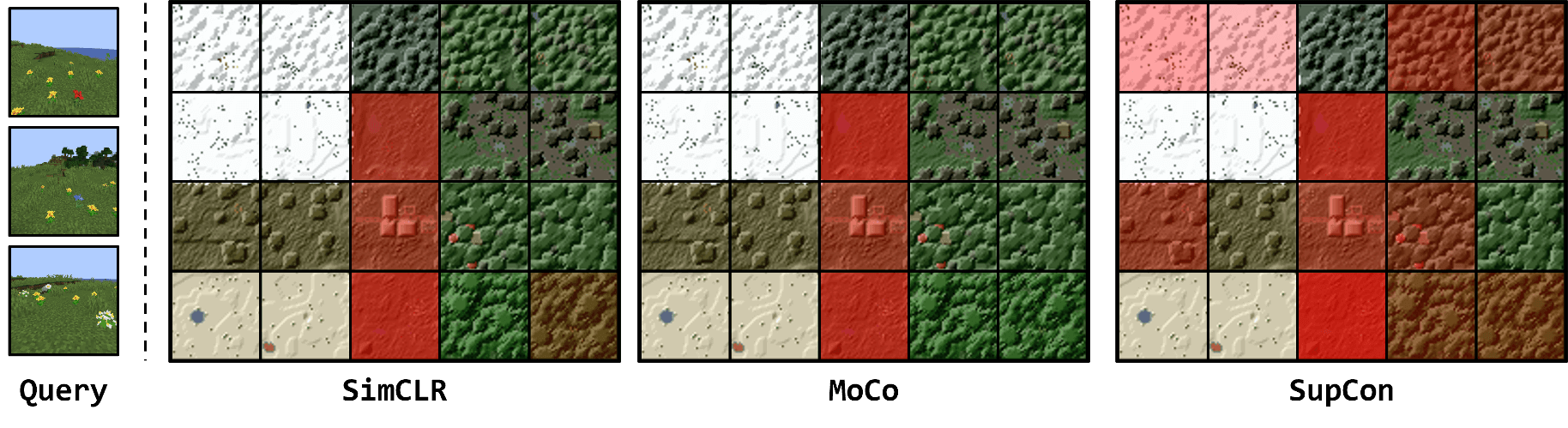}
        \vskip -0.5em
        \caption{\texttt{Flower} in \texttt{Plains}}
    \end{subfigure}
    \vskip 0.5em
    \begin{subfigure}{0.9\textwidth}
        \includegraphics[width=\textwidth]{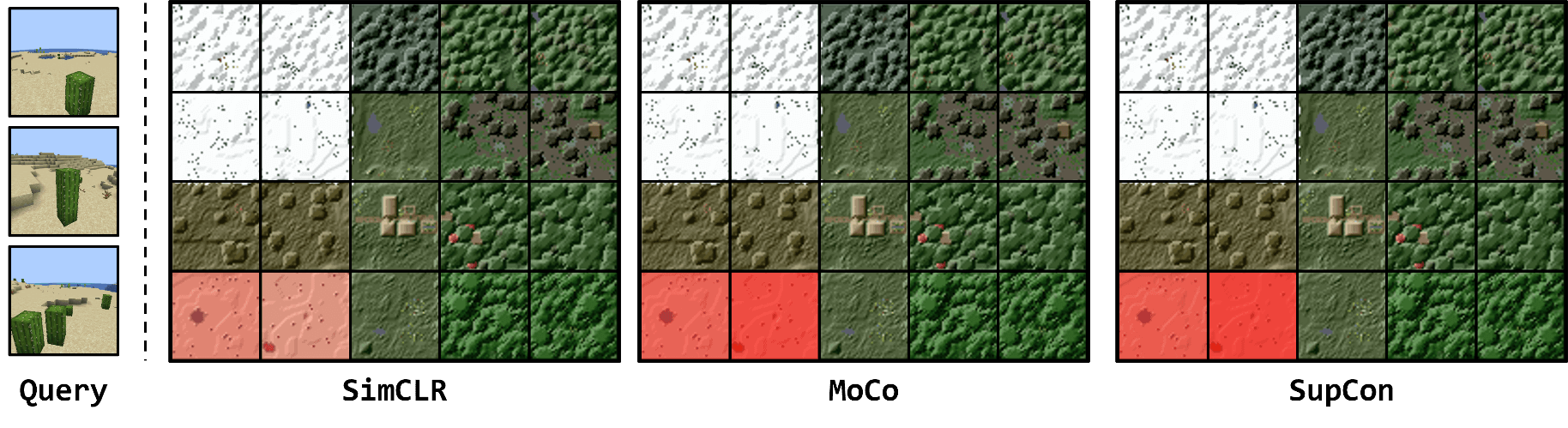}
        \vskip -0.5em
        \caption{\texttt{Cactus} in \texttt{Desert}}
    \end{subfigure}
    \vskip 0.5em
    \begin{subfigure}{0.9\textwidth}
        \includegraphics[width=\textwidth]{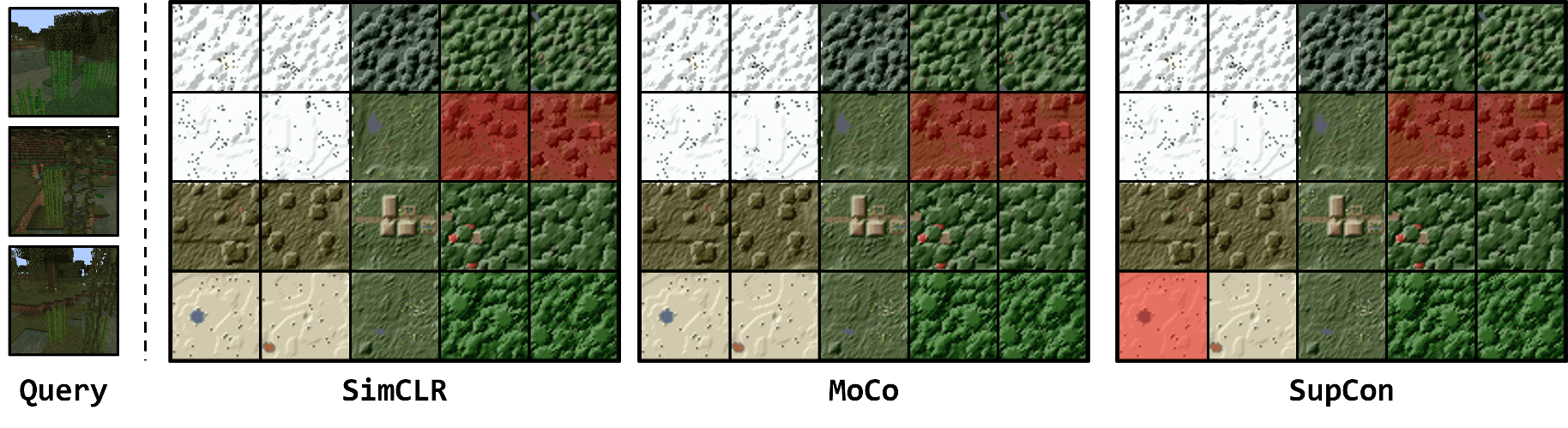}
        \vskip -0.5em
        \caption{\texttt{Sugar\_cane} in \texttt{Swamp}}
    \end{subfigure}
    \caption{Visualization of results with object-environment retrieval task in Miniature environment.}
    \label{res-minecraft-obj-retrieval-apdx}
\end{figure}

The verification of the object-environment retrieval task is performed through the following process. First, observations of objects within each grid of the Miniature environment are obtained. Then, using 5 observations of the objects as a query, the object occurrence is estimated for each model, followed by the visualization of heatmaps of the estimated probabilities of each models. \cref{res-minecraft-obj-retrieval-apdx} shows additional results with different queries.

\subsubsection{Difference between metric learning and SSL in OBSER}
\begin{figure}[ht]
    \centering
    \includegraphics[width=\textwidth]{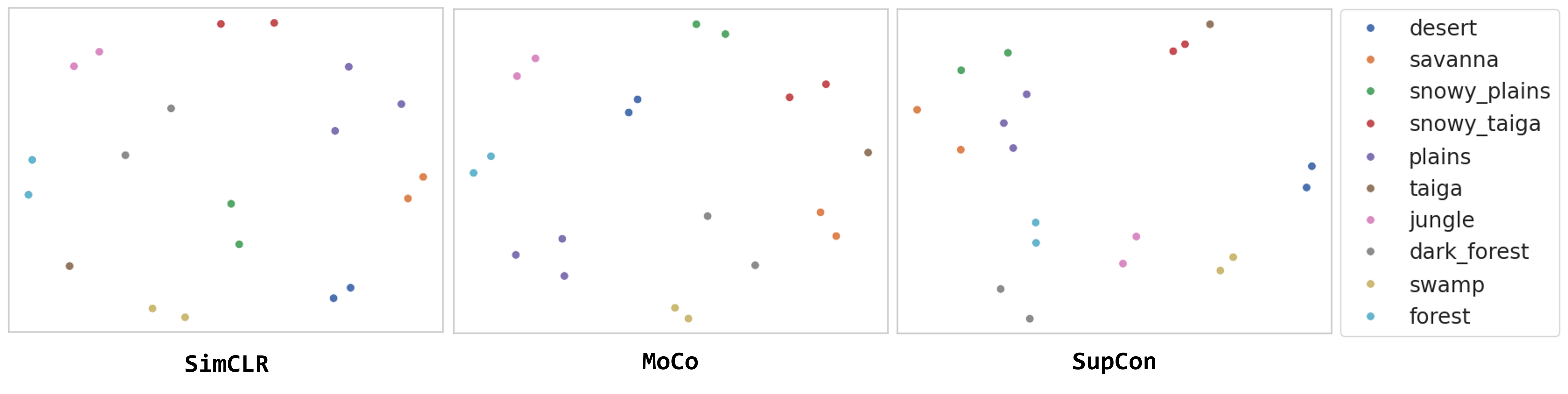}
    \vskip -1em
    \caption{MDS visualization of biomes in Miniature environment. Jensen-Shannon divergence (JSD) is used as the measure between sub-environments. In SSL models, metrics are similarly distant across biomes, while in SupCon, distinct metrics reflect the latent class distribution of objects.}
    \label{minecraft-res-mds}
    \vskip -0.5em
\end{figure}

We visualize the pair-wise environmental relationship with the Jensen-Shannon divergence. \cref{minecraft-res-mds} shows the MDS visualization~\citep{buja2008data} of biomes in the Miniature environment with each model. The SSL models tend to keep the metrics of sub-environments farther apart, as they consider ambient biases beyond object distribution within each sub-environment. In contrast, the metric learning model better captures differences in object distribution but struggles to incorporate less task-relevant information.

We have observed that both SSL and metric learning models demonstrate the capability of making sufficiently accurate inferences for each recognition task in OBSER. We have also observed distinct differences between these paradigms. SSL models tend to integrate environmental information that is not directly relevant to the task, while metric learning models focus on task-specific information for more accurate inference. We claim that SSL and metric learning models each offer unique advantages in terms of generalization and accuracy. Therefore, applying the appropriate model based on the given situation of the problem is crucial for achieving effective sub-environment recognition in agents.

\section{Chained Inference of OBSER Framework in Photorealistic Environment (Replica)}
\label{apdx-replica}

\subsection{Replica environment}
\label{apdx-replica-env}
We conduct our experiments using the Replica environment, an indoor space comprised of a high-resolution 3D mesh that serves as an excellent testbed for validating the proposed framework. In our experiment, we hypothesized that each model would efficiently retrieve the appropriate object from given queries, provided it successfully executes three consecutive step inferences.

\begin{figure}[t]
    \centering
    \includegraphics[width=0.8\textwidth]{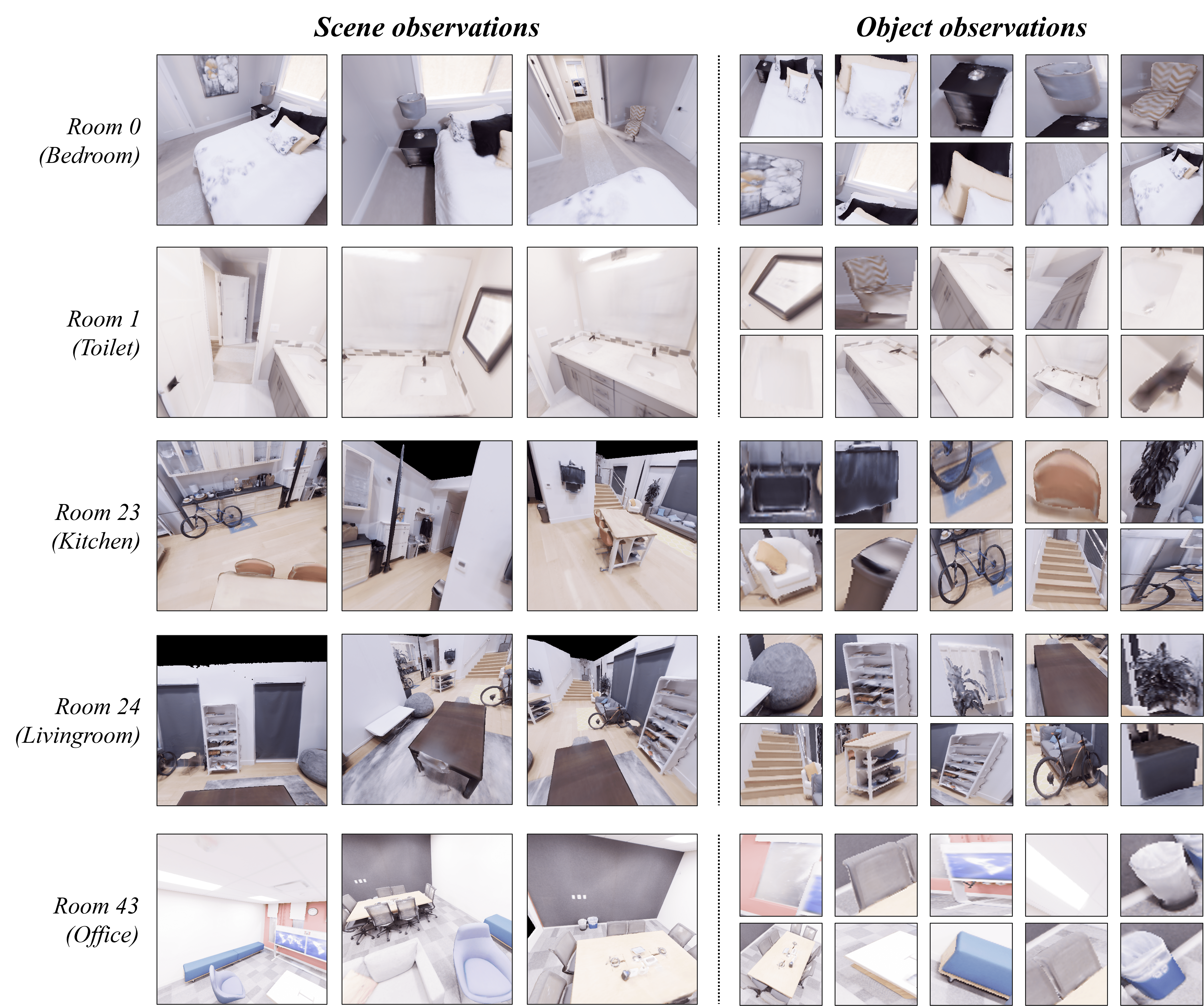}
    \caption{Example scene obsevations and object observations in each room.}
    \label{apdx-replica-scene}
\end{figure}

\begin{figure}[t]
    \centering
    \includegraphics[width=0.8\textwidth]{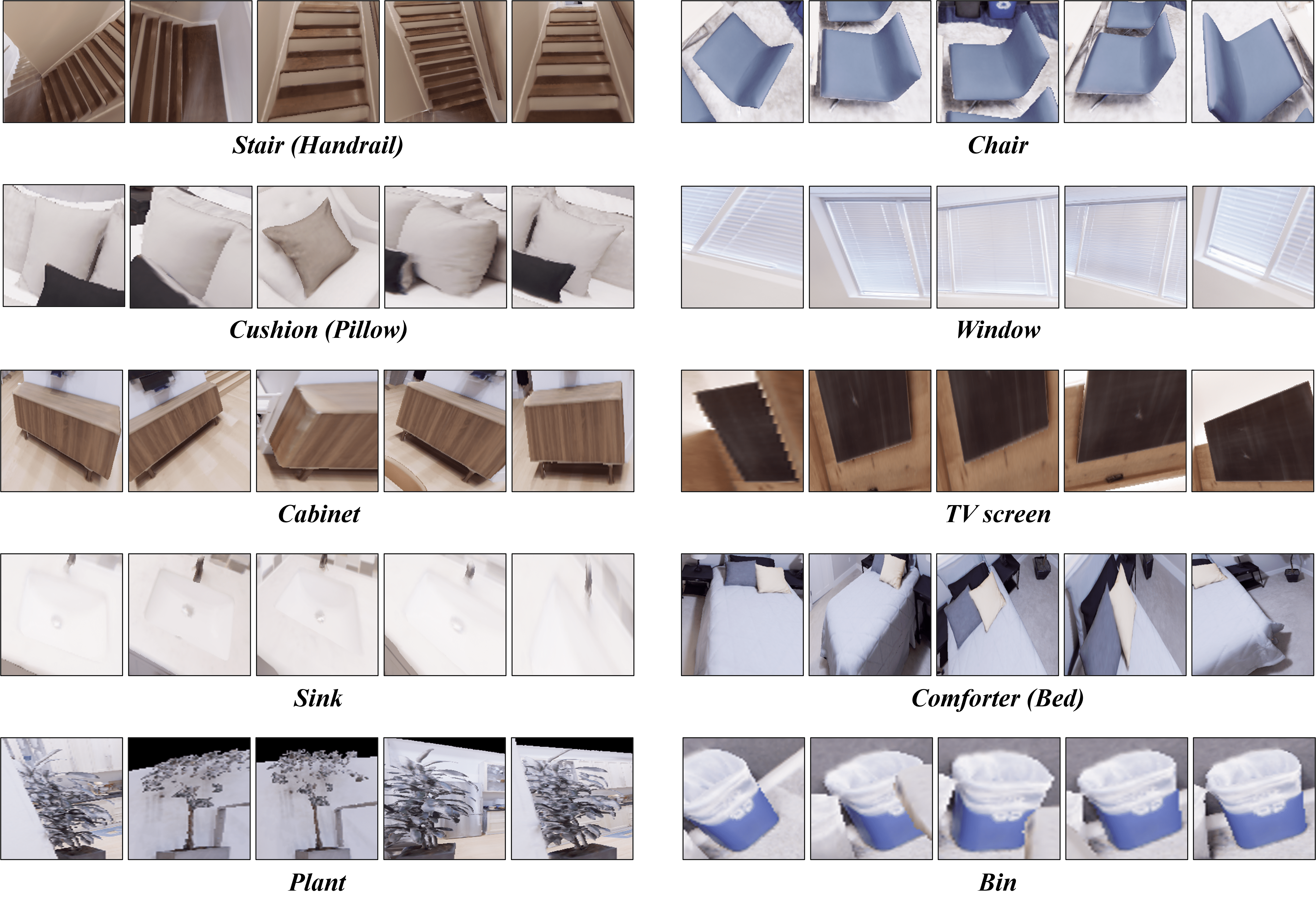}
    \caption{Queried obsevations used for the experiments. For some ambiguous queries, multiple classes were accepted as correct (indicated in parentheses).}
    \label{apdx-replica-query}
\end{figure}

We first have collected 960 random scenes from 48 rooms (20 scenes per room) and extracted object observations from each scene. These observations are used to construct empirical distributions for both episodic memory and the environment. The experiments are conducted under two conditions: a seen condition, where $\gM = \gE$, and an unseen condition, where $(\bigcup_e\hat{R}_e) \cap (\bigcup_m\hat{R}_m) = \emptyset$. Specifically, in the unseen condition, the episodic memory is limited to the rooms of \texttt{apartment\_0} (comprising 13 rooms). Using 10 different objects as queries, we compute the inference accuracy for each model. \cref{apdx-replica-scene} and \cref{apdx-replica-query} shows the example observations and queried observations each.

\paragraph{Extract object observations}{We have followed the following process to extract object observations from each scene observation. First, we extract a mesh-wise semantic segmentation for each scene. Next, we remove objects that were too small from the segmentation. In cases where label information is accessible (GT), we remove walls, ceilings, and floors. When label information is not accessible (SAM), we remove objects that contain the edges of the observations. This process allows us to extract relatively acceptable object observations.
}

\subsection{Problem definition}
In this section, we discuss the application of OBSER framework for navigation tasks. Suppose that a navigation task is defined as locating to the position $p_q$ of a given object $x_q$ in a given environment $\gE:=\{(\mu_e,R_e)\}_{e=1}^{E}$. Then an agent should infer i) the most probable sub-environment in episodic memory, ii) locate the most similar sub-environment with given memory and iii) find the object in such sub-environment.

Let an episodic memory $\gM:=\{(\hat{\mu}_m,\hat{R}_m)\}_{m=1}^{M}$ be a set of observations $\hat{\mu}_m:=\{x_{mo}\}_{o=1}^{O}$ and its locations $\hat{R}_{m}:=\{p_{mo}\}_{o=1}^{O}$. Depending on assumptions of the tasks, the location may be unknown or useless (unseen). With a given query $x_q$, the most probable sub-environment in episodic memory can be inferred with \textbf{Object Occurrence (object-environment)}:
\begin{equation*}
i) \quad m^*=\argmax_{m\in\{1,\cdots,M\}}\Phi_f(x_q;\hat\mu_m)
\end{equation*}
With a reachable region $\gN_{R}(R;\gE):=\{e|R_e\subseteq\text{reachable}(R,\gE),s\in\{1,\cdots,S\}\}$ with the region that the agent is located, an agent can retrieve the most similar sub-environment which minimizes \textbf{the KL divergence (environment-environment)} with given memory $(\hat{\mu}_{m^*},\cdot)$.
\begin{equation*}
ii) \quad e^*=\argmin_{e\in\gN_{R}(\hat R_{m^*};\gE)}\widehat{\KL}_f(\hat{\mu}_{m^*}||\mu_{e})
\end{equation*}
After the agent reaches to $R_{e^*}$, it explores the region $R_{e^*}$ to find a target position which has \textbf{the same object with given query $x_q$ (object-object)}:
\begin{equation*}
iii) \quad p^*=\argmax_{p\in R_{e^*}} \phi_f(x_q,x_p)
\end{equation*}
with observation $x_{p^*}$ in position $p^*$.

\paragraph{Baselines with CLIP}{To compare the performance of the OBSER framework, we utilize the CLIP model, which is frequently used in scene-based recognition, as the baseline. Since cosine similarity is the measurement commonly used in retrieval tasks with the CLIP model, we adapt it as the criterion for inference.

For the vision-only baseline (CLIP V), we use the average cosine similarity of scenes in each sub-environment based on the mean vector of the query. In the case of environment-environment recognition, we use the average cosine similarity across all scene pairs as the criterion. For the baseline that uses both vision and language (CLIP V$+$L), we define words for each query and room (e.g., "a \textit{sink}" in "the \textit{toilet}") and add the cosine similarity of the word embeddings for retrieval.
}

\clearpage

\subsection{Why scene-based method underperforms than OBSER?}
\label{apdx-sec-replica-clip}
\begin{figure}[ht]
\centering
    \begin{subfigure}{0.48\textwidth}
        \includegraphics[width=\textwidth]{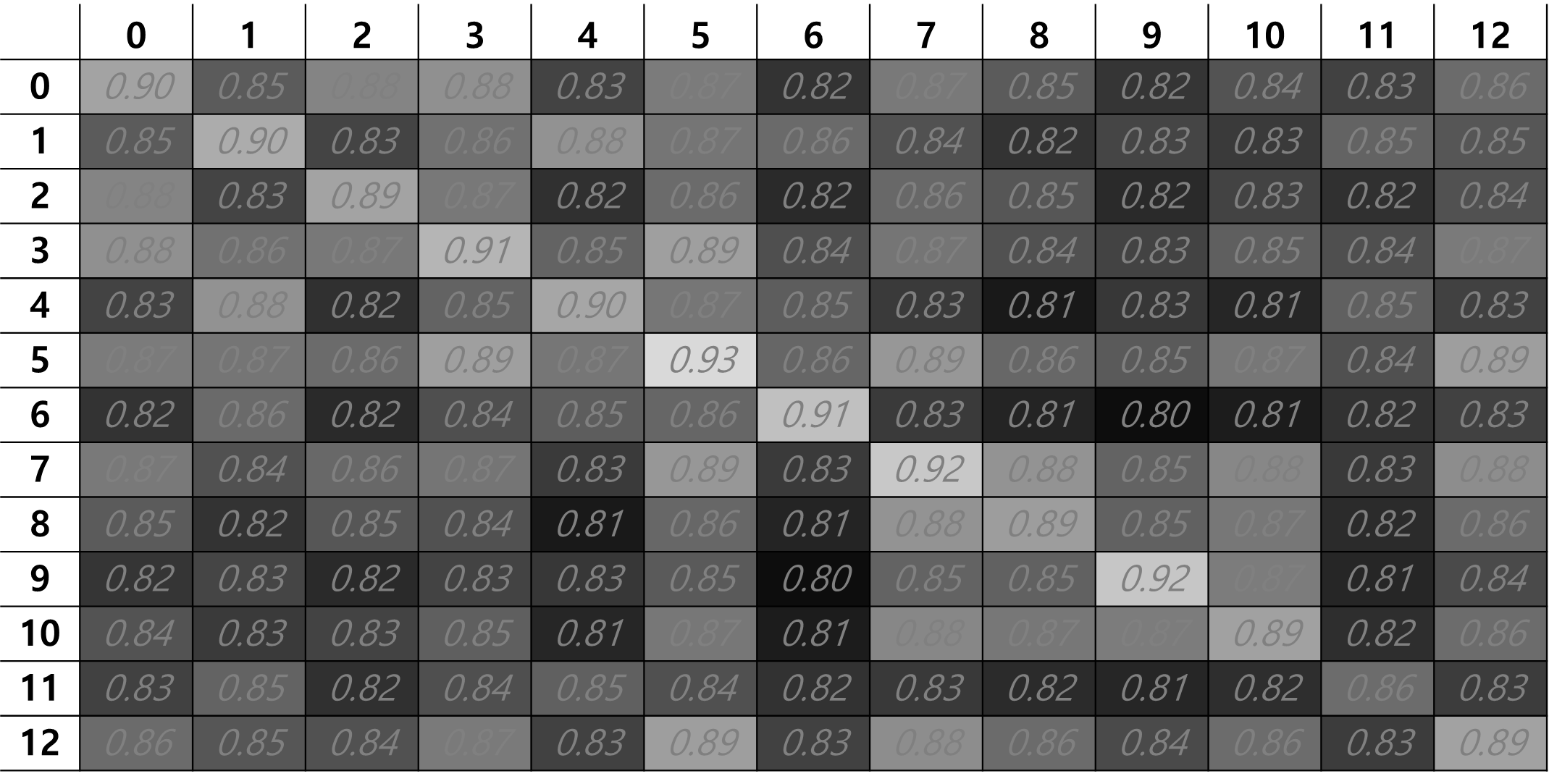}
        \caption{Scene-based (CLIP)}
    \end{subfigure}%
    \hskip 1em
    \begin{subfigure}{0.48\textwidth}
        \includegraphics[width=\textwidth]{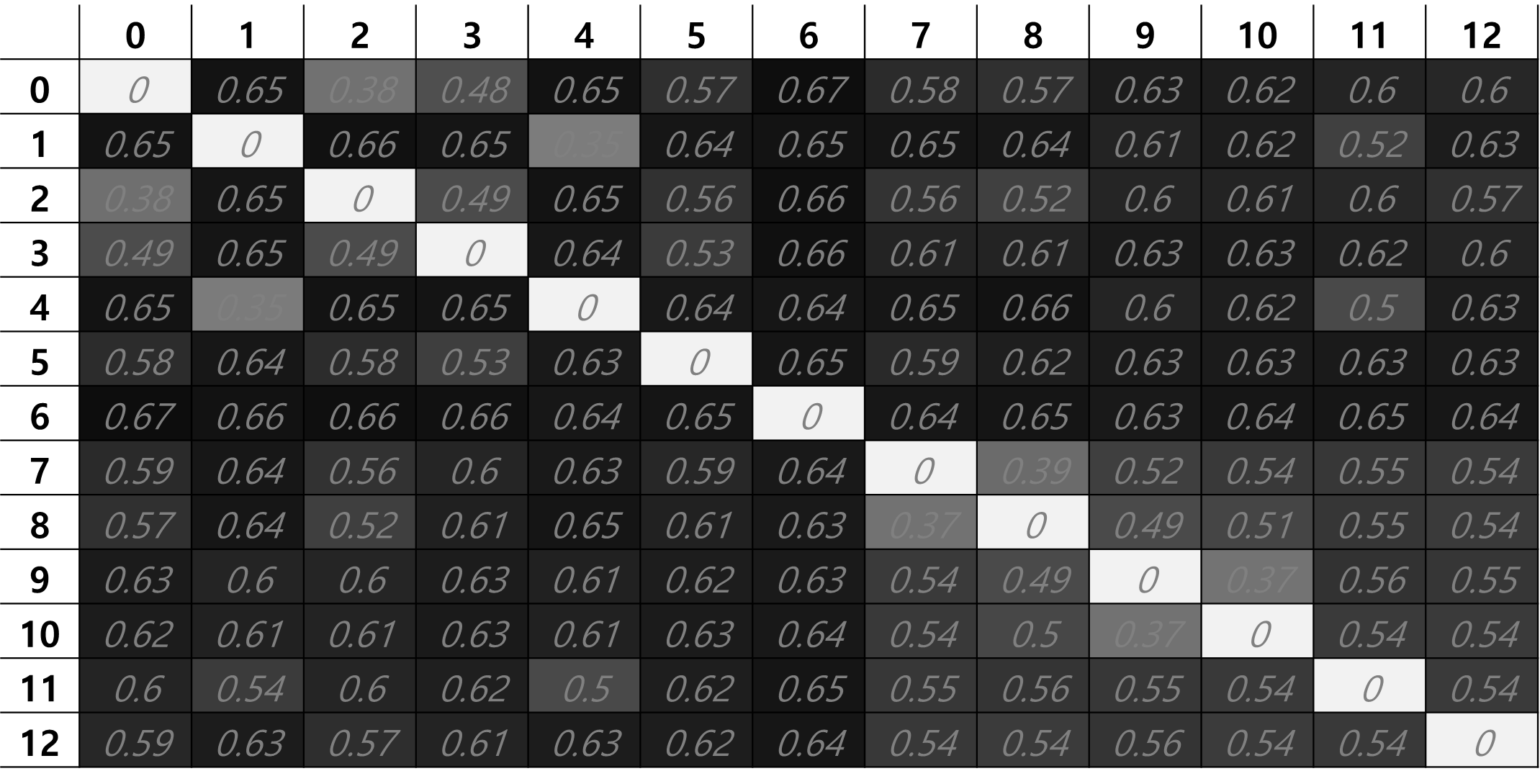}
        \caption{Object-based (DINO-v2)}
    \end{subfigure}
    \caption{Visualization of the mutual environment relationships among the rooms in \texttt{apartment\_0}. For CLIP, average cosine similarity is used, while for DINO-v2, Jensen-Shannon divergence is applied. Darker colors indicate more dissimilar rooms, whereas brighter colors indicate more similar rooms.}
    \label{apdx-replica-clip-envenv}
\end{figure}

In the chained retrieval task, we observe that the scene-based method leveraging the CLIP model underperforms relative to the OBSER framework. We examine the accuracy of environment-environment and object-environment relationships for each approach to analyze the cause. Figure \ref{apdx-replica-clip-envenv} illustrates the relationships among rooms in the \texttt{apartment\_0} environment. The CLIP-based method exhibits high average similarity between scenes, which hinders accurate discrimination among rooms, whereas the OBSER framework classifies similar rooms according to each room’s characteristics.

\begin{figure}[ht]
    \centering
    \includegraphics[width=0.8\textwidth]{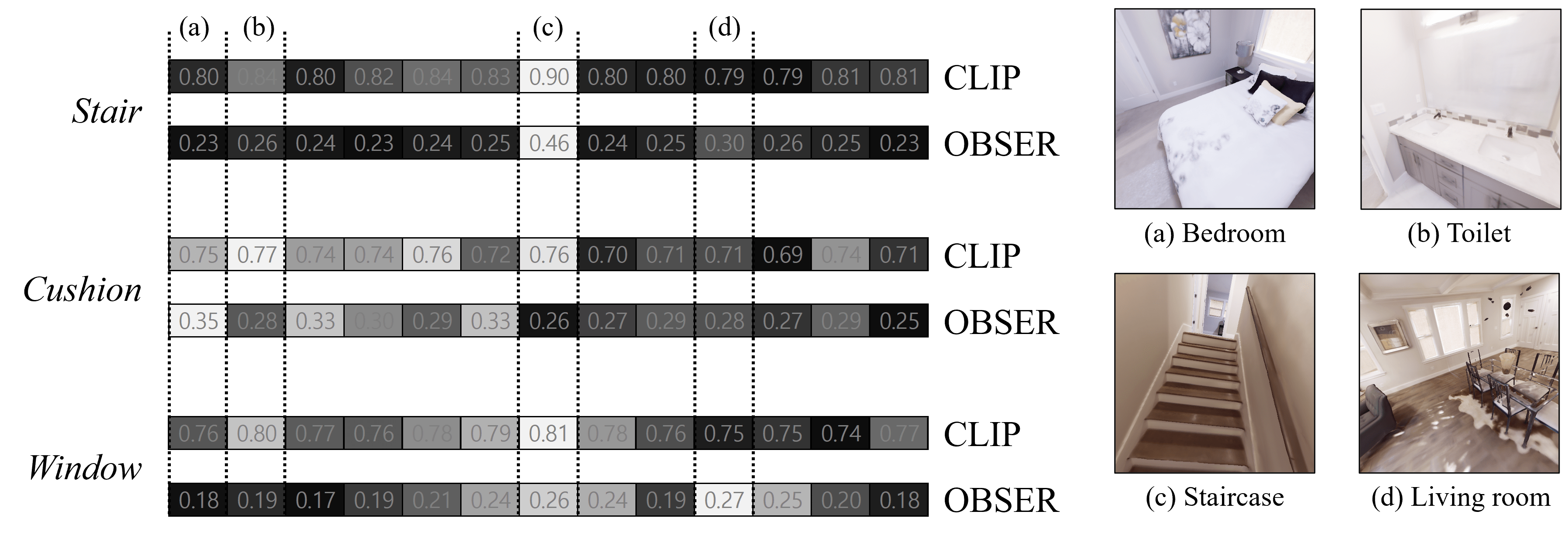}
    \caption{Visualization of the object-environment relationships between a given query and the rooms in \texttt{apartment\_0}. Darker colors indicate lower relevance, while brighter colors indicate higher relevance.}
    \label{apdx-replica-clip-objenv}
\end{figure}

Figure \ref{apdx-replica-clip-objenv} illustrates the relationships between a given query object and the rooms within \texttt{apartment\_0}. Each query utilizes the observations provided in Figure \ref{apdx-replica-query}. With CLIP, the inference relies on the query image’s visual features—such as color—instead of the environmental context, resulting in task failures (e.g., cushion, window).

In conclusion, due to the intrinsic nature of the CLIP model, representations are formed based on the language associated with the observations, which leads to inference challenges in the absence of human supervision such as instructions. In contrast, the OBSER framework approximates spatial representations as a distribution rather than a single point in latent space, allowing each measure to be computed accurately and generalizing effectively even without supervision.

\newpage

\subsection{Discussion A. building episodic memory}
\label{apdx-sec-replica-disc-a}

Applying the OBSER framework to an agent requires robust design and management of episodic memory. In our experiments, we focus on both constructing and updating episodic memory, and this section discusses strategies for achieving that.

\begin{figure}[ht]
    \centering
    \includegraphics[width=\textwidth]{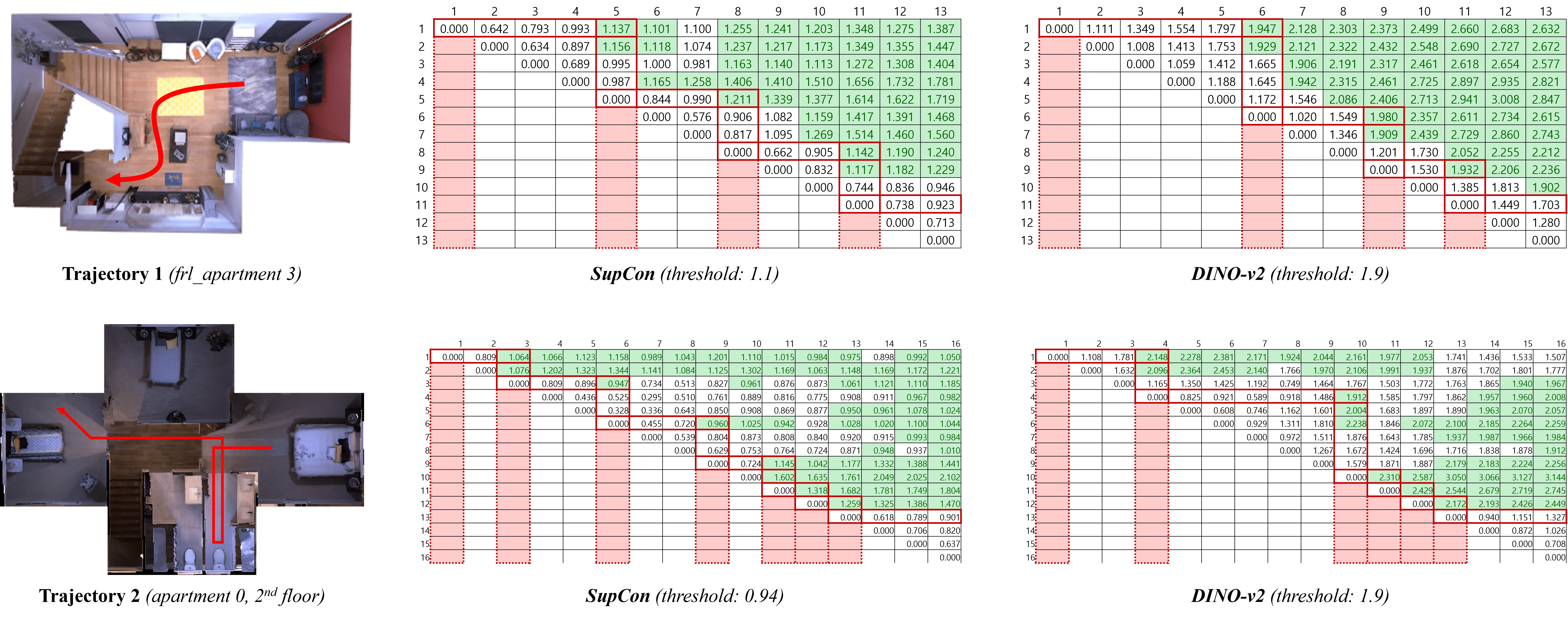}
    \caption{Visualization of the pairwise KL divergence between two waypoints in different trajectories. By using the circumstance at a specific moment as a pivot and updating the pivot when the KL divergence exceeds a threshold, we can distinguish sub-environments and form episodic memory.}
    \label{apdx-replica-traj}
\end{figure}

To add a new sub-environment to memory, we leverage OBSER’s environment-environment relationship. Figure \ref{apdx-replica-traj} shows the environmental differences measured between an arbitrary pivot waypoint and the waypoints along each trajectory. For this purpose, we capture the circumstance at each waypoint using 12 scene observations (rotating 30 degrees) and extract object observations from them to form a distribution. Consequently, by setting an appropriate threshold, we can distinguish similar waypoints in a continuous trajectory, thereby forming distinct episodic memories.

\newpage
\subsection{Qualitative result of Discussion B. recognizing objects in a fully unsupervised manner}
\label{apdx-sec-replica-disc-b}

\begin{figure}[ht]
\centering
    \begin{subfigure}{0.9\textwidth}
        \includegraphics[width=\textwidth]{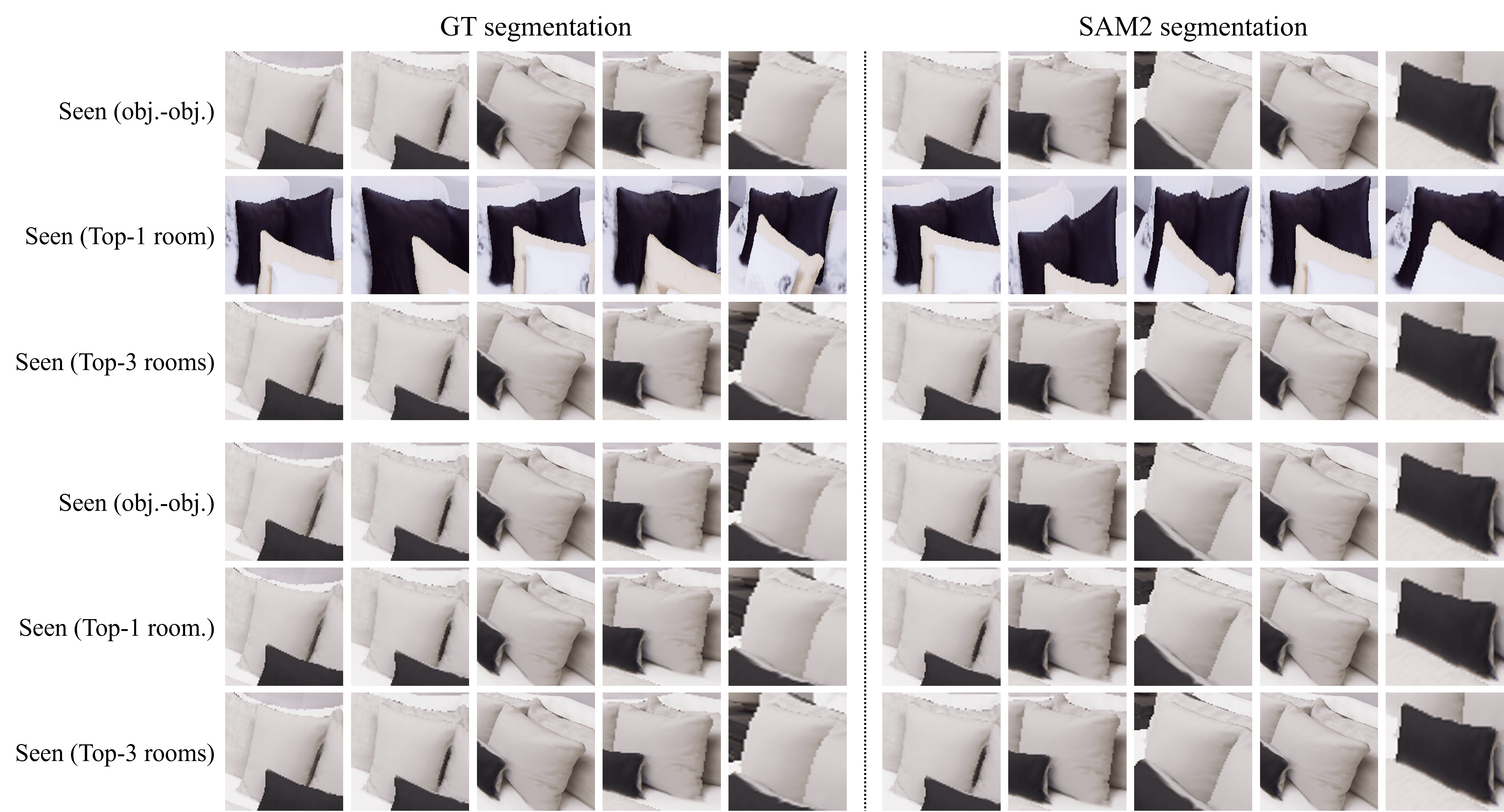}
        \caption{Query: \textit{Cushion}}
    \end{subfigure}
    \vskip 0.5em
    \begin{subfigure}{0.9\textwidth}
        \includegraphics[width=\textwidth]{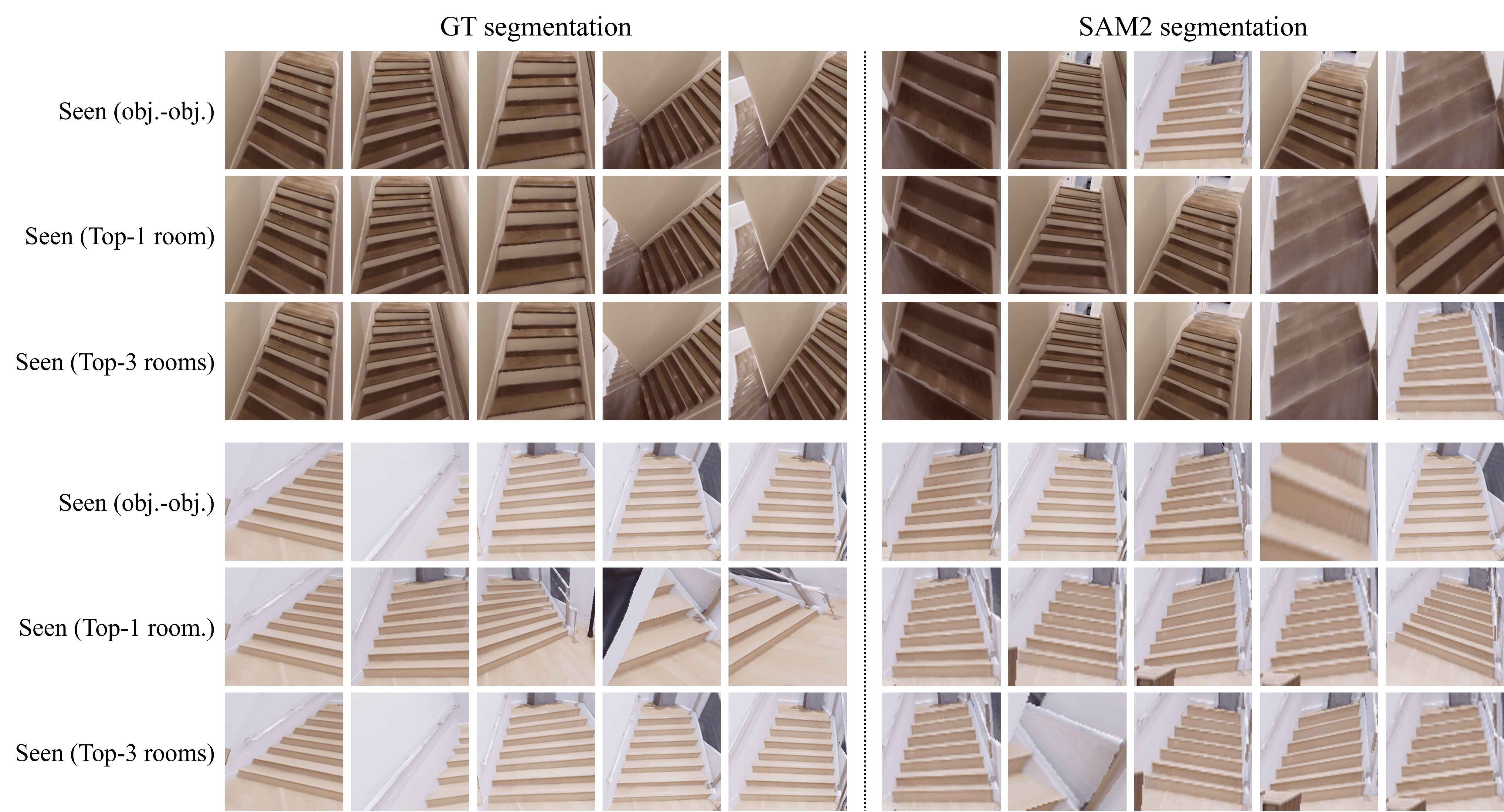}
        \caption{Query: \textit{Stair}}
    \end{subfigure}
    \caption{Visualization of the chained inference results using the object distribution extracted from both ground-truth segmentation and segmentation generated by the SAM2 model (success). Queries are given as object observations, as shown in Figure \ref{apdx-replica-query}. We utilize DINO-v2-B to show the results.}
    \label{apdx-replica-sam-success}
\end{figure}

\begin{figure}[ht]
\centering
    \begin{subfigure}{0.9\textwidth}
        \includegraphics[width=\textwidth]{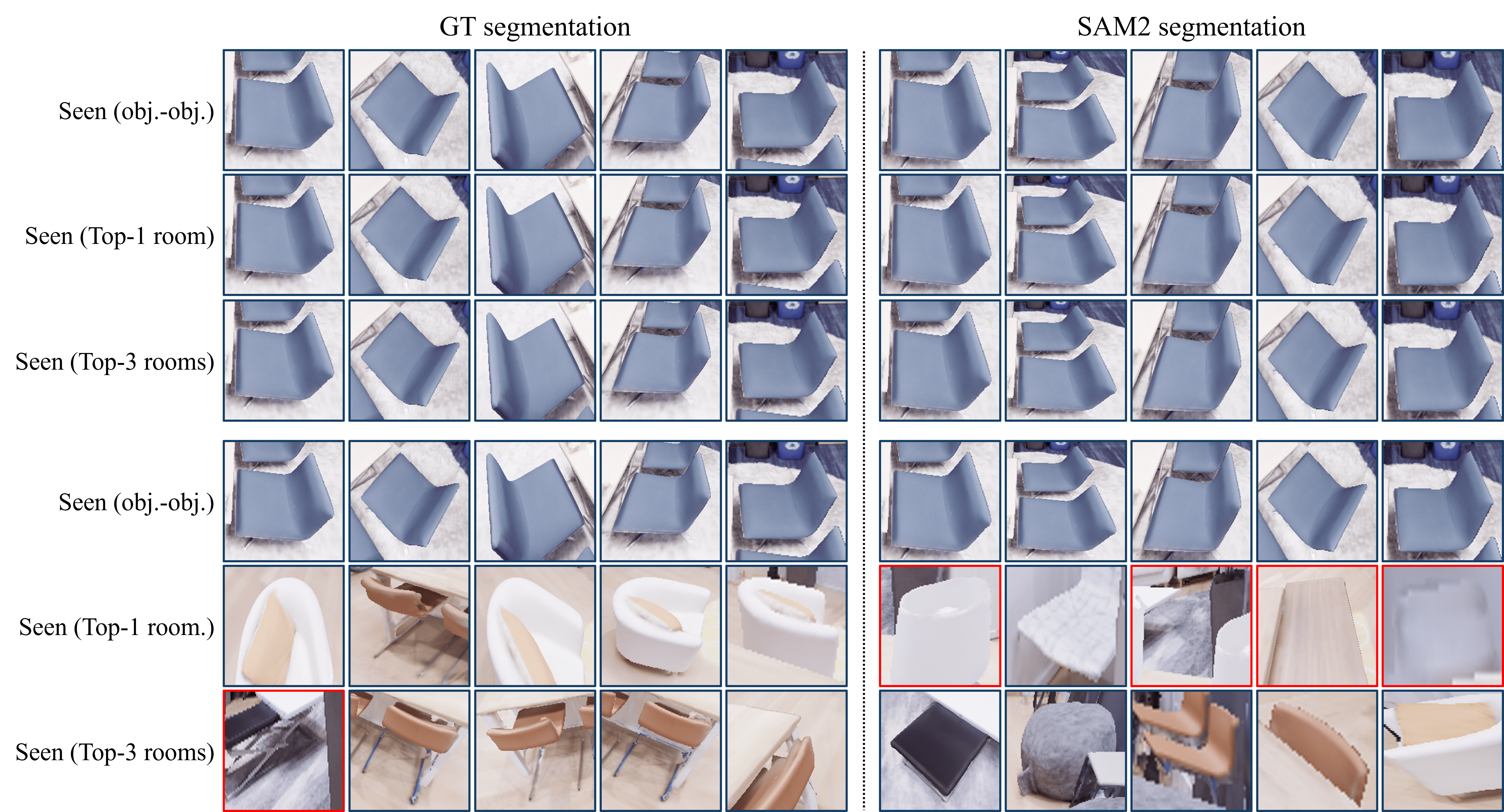}
        \caption{Query: \textit{Chair}}
    \end{subfigure}
    \vskip 0.5em
    \begin{subfigure}{0.9\textwidth}
        \includegraphics[width=\textwidth]{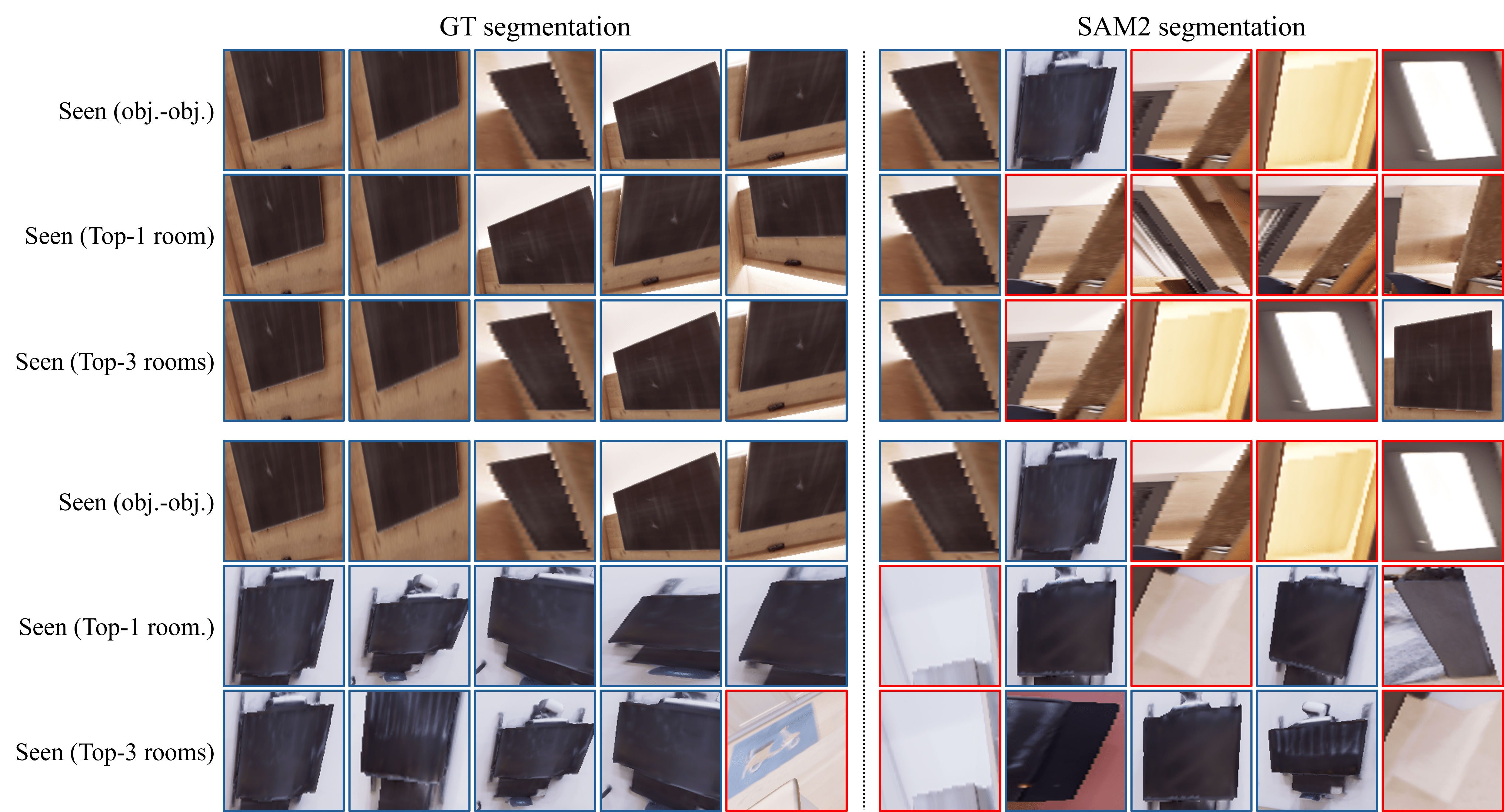}
        \caption{Query: \textit{TV screen}}
    \end{subfigure}
    \caption{Visualization of the chained inference results using the object distribution extracted from both ground-truth segmentation and segmentation generated by the SAM2 model (failed). Although SAM segmentation sometimes includes non-object observations that can negatively affect environmental recognition, our OBSER framework still demonstrates sufficient inference performance on the given queries.}
    \label{apdx-replica-sam-fail}
\end{figure}

\end{document}